\documentclass[runningheads]{llncs}

\usepackage{eccv}

\usepackage{eccvabbrv}

\usepackage{graphicx}
\usepackage{booktabs}

\usepackage[accsupp]{axessibility}  

\usepackage{hyperref}

\usepackage{orcidlink}

\usepackage{graphics,graphicx,caption,float,color}
\usepackage{booktabs} 
\usepackage{adjustbox}
\usepackage{multirow}
\usepackage{multicol}
\usepackage{amsmath}
\usepackage{xcolor}
\usepackage{ulem}
\usepackage{amsfonts} 
\usepackage{colortbl}
\usepackage{makecell} 
\usepackage{amssymb}
\usepackage{mathtools}
\usepackage{enumitem}

\usepackage{stfloats}
\usepackage{esint} 
\usepackage{cuted}

\begin{document}

\title{PredBench: Benchmarking Spatio-Temporal Prediction across Diverse Disciplines} 

\titlerunning{PredBench}

\author{
ZiDong Wang\inst{1,2,*}\orcidlink{0009-0003-8462-6819} \and
Zeyu Lu\inst{1,3,*}\orcidlink{0000-0003-0494-911X} \and
Di Huang\inst{1,4}\orcidlink{0009-0009-8712-8747} \and
Tong He\inst{1}\orcidlink{0000-0003-2772-9320} \and
Xihui Liu\inst{1,5}\orcidlink{0000-0003-1831-9952} \and \\
Wanli Ouyang\inst{1,2}\orcidlink{0000-0002-9163-2761} \and
Lei Bai\inst{1,\dag}\orcidlink{0000-0003-3378-7201}
}

\authorrunning{Z.~Wang et al.}

\institute{
Shanghai AI Laboratory \and The Chinese University of Hong Kong \and Shanghai Jiao Tong University \and Sydney University \and The University of Hong Kong \\
(*)equal contribution; (\dag) corresponding author \\
\email{\{wangzidong,luzeyu,huangdi,hetong,liuxihui,ouyangwanli,bailei\}@pjlab.org.cn}
}

\maketitle

\begin{abstract}
  In this paper, we introduce PredBench, a benchmark tailored for the holistic evaluation of spatio-temporal prediction networks. Despite significant progress in this field, there remains a lack of a standardized framework for a detailed and comparative analysis of various prediction network architectures. PredBench addresses this gap by conducting \textbf{large-scale experiments}, upholding \textbf{standardized and appropriate experimental settings}, and implementing \textbf{multi-dimensional evaluations}. This benchmark integrates 12 widely adopted methods with 15 diverse datasets across multiple application domains, offering extensive evaluation of contemporary spatio-temporal prediction networks. 
    Through meticulous calibration of prediction settings across various applications, PredBench ensures evaluations relevant to their intended use and enables fair comparisons.
    Moreover, its multi-dimensional evaluation framework broadens the analysis with a comprehensive set of metrics, providing deep insights into the capabilities of models. 
    The findings from our research offer strategic directions for future developments in the field. 
    Our codebase is available at \url{https://github.com/OpenEarthLab/PredBench}.
\end{abstract}

\begin{figure}[!htbp]
  \vspace{-0.0cm}
  \centering
  \includegraphics[width=1.0\textwidth]{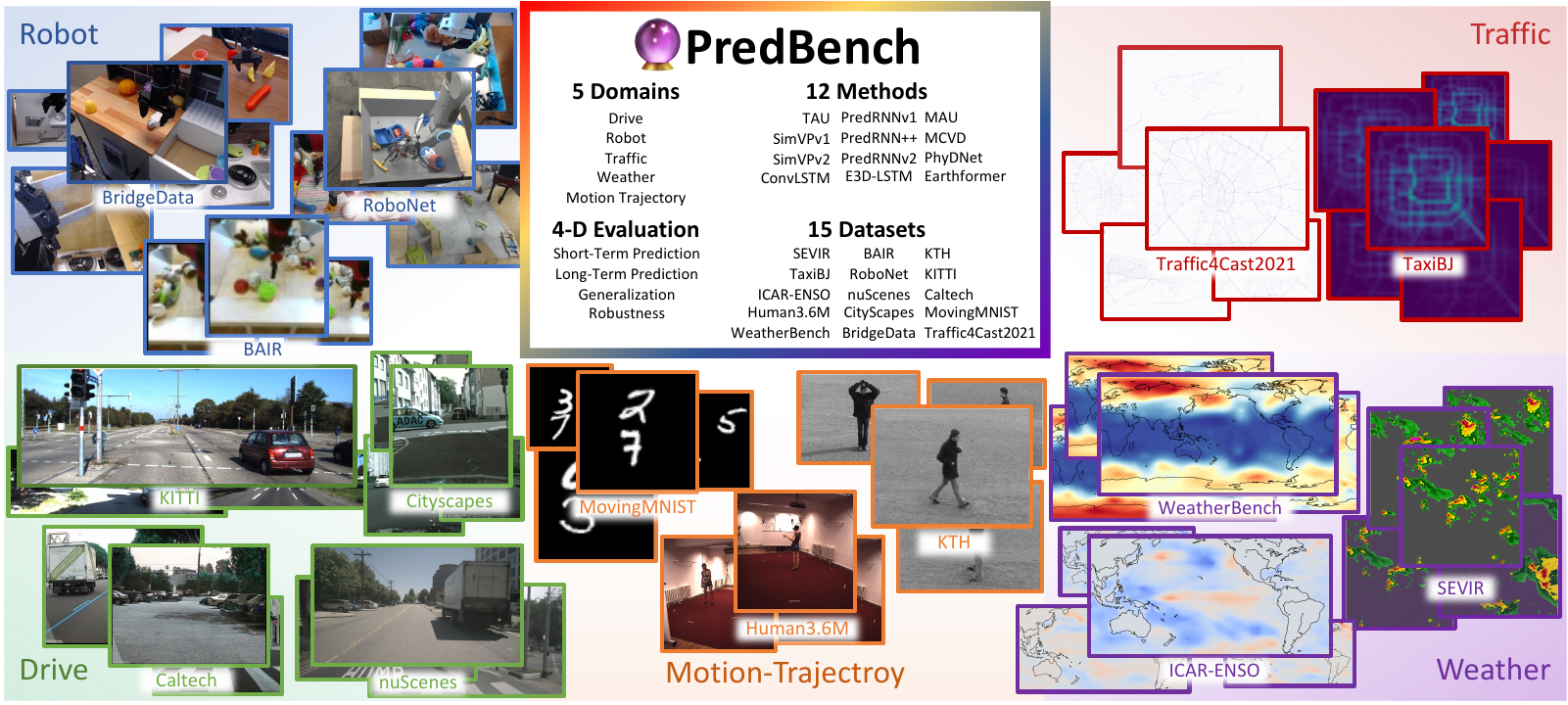}
  \vspace{-0.4cm}
  \caption{
  Overview of our \textbf{spatio-temporal Prediction Benchmark (PredBench)}. 
  It conducts a thorough 4-dimensional evaluation of 12 prevalent spatio-temporal prediction methods, spanning 5 distinct domains and covering 15 diverse datasets.
  }
  \label{fig:teaser}
  \vspace{-0.6cm}
\end{figure}

\section{Introduction}
\label{sec:intro}
Spatio-Temporal Prediction (STP) represents a cornerstone of research in computer vision and artificial intelligence. 
It leverages historical data to forecast future events, with far-reaching implications for diverse fields such as meteorology~\cite{ravuri2021skiful,bi2023pangu,lam2022graphcast,chen2023fengwu}, robotics~\cite{finn2016unsupervised,du2023learning}, and autonomous vehicles~\cite{hu2023gaia}.
Despite the proliferation of methods in STP, a comprehensive understanding of network performance across different disciplines and applications remains elusive.

\begin{figure}[tb]
  \centering
  \includegraphics[width=1.0\textwidth]{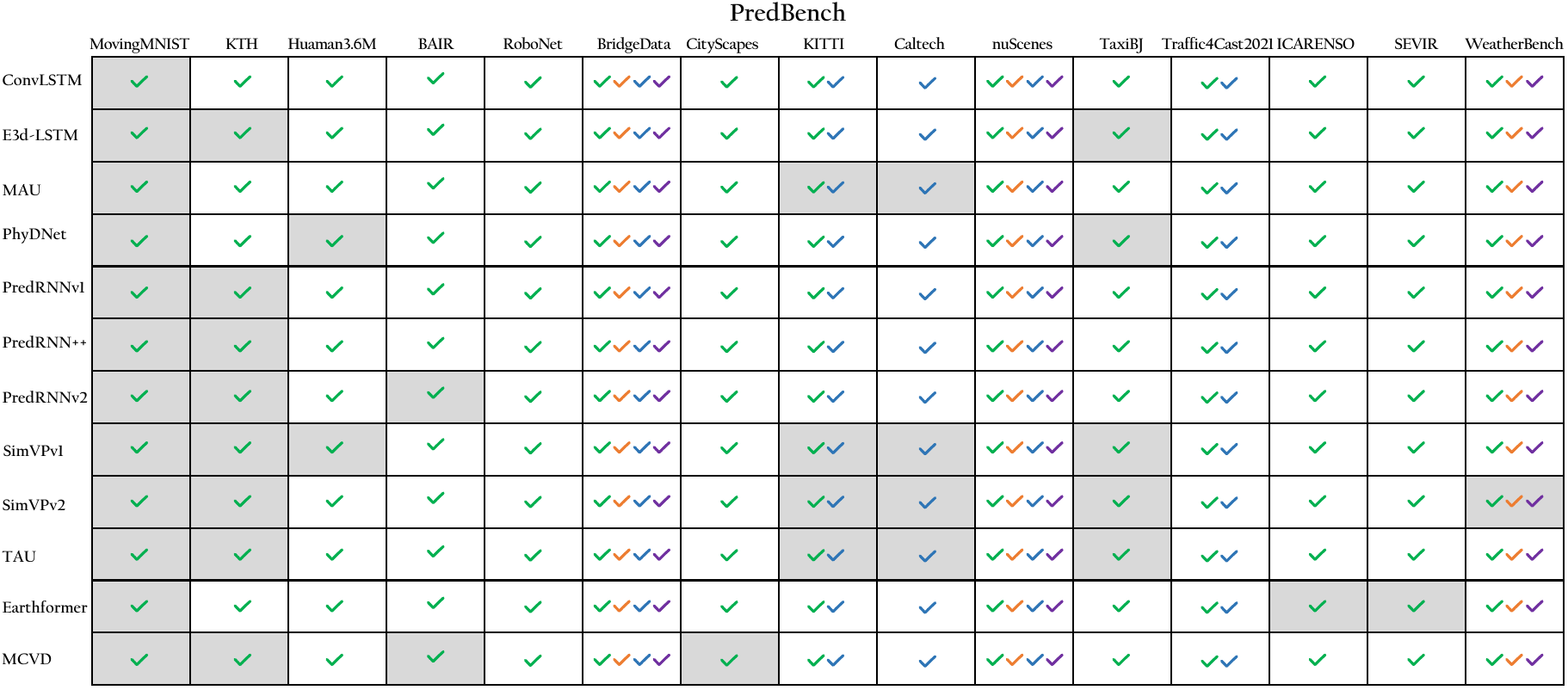}
  \vspace{-0.6cm}
  \caption{
      \textbf{We support 12 methods and 15 datasets} in our PredBench. The \textcolor{Gray}{gray cells} represent the settings in which previous methods have been conducted. We \textbf{fill the remaining blank cells} by conducting large-scale experiments and thorough evaluation. The \textcolor{Green}{green ticks} indicate that short-term prediction experiments are conducted, while \textcolor{orange}{orange ticks} signify the implementation of long-term prediction experiments. The \textcolor{RoyalBlue}{blue ticks} represent the execution of generalization experiments, and \textcolor{blue!50!purple}{purple ticks} denote experiments in temporal resolution robustness. 
  }
  \vspace{-0.5cm}
  \label{fig:supported}
\end{figure}

The pursuit of STP introduces several challenges that complicate the creation of a holistic benchmark.
Firstly, the universality of STP across numerous applications and disciplines necessitates a comprehensive evaluation encompassing a wide array of datasets.
As shown in \cref{fig:supported}, traditional STP~\cite{finn2016unsupervised,babaeizadeh2018sv2p,wang2017predrnn,shi2015convlstm,srivastava2015movingmnist,gupta2023maskvit,babaeizadeh2021fitvid} studies often assess models on limited datasets, thus failing to present the performance of the model in varied scenarios. 
Secondly, fair and meaningful comparison requires the prediction settings to maintain consistency across different networks.
Historically, there has been a setting disparity of different networks within the same dataset, leading to results that are not directly comparable. For example, the MCVD~\cite{voleti2022mcvd} model might input 1 or 2 frames and forecast the following 5 frames during training, while PredRNNv2~\cite{wang2021predrnnv2} might use 2 frames to predict the next 10 frames on \textit{BAIR}~\cite{ebert2017bair} dataset. 
Thirdly, a thorough comparison across various STP models must encompass multiple dimensions and metrics to assess the full spectrum of network performance, while previous methods often evaluate networks with limited aspects and metrics. 


This paper presents PredBench, a comprehensive framework devised for the holistic evaluation of STP networks.
As shown in Fig.~\ref{fig:teaser}, PredBench stands as the most exhaustive benchmark to date, integrating 12 established STP methods~\cite{shi2015convlstm,wang2017predrnn,wang2018predrnn++,wang2021predrnnv2,wang2019e3dlstm,gao2022simvp,tan2022simvpv2,chang2021mau,tan2023tau,voleti2022mcvd,gao2022earthformer} and 15 diverse datasets~\cite{geiger2013kitti,caesar2020nuscenes,dollar2009caltech,cordts2016cityscapes,srivastava2015movingmnist,schuldt2004kth,ionescu2013human,Traffic4Cast2021,ICARENSO,walke2023bridgedata,dasari2019robonet,ebert2017bair,garg2022weatherbench,veillette2020sevir,zhang2018taxibj} from a range of applications and disciplines.
It presents a standardized experimental protocol to facilitate fair and meaningful comparisons across diverse STP methods and datasets. Additionally, PredBench introduces four evaluation dimensions, thoroughly assessing the short-term prediction abilities, long-term prediction abilities,  generalization abilities, and temporal robustness of the model  across domains, thus addressing gaps in current evaluation practices. Through large-scale experimentation, we have derived several significant findings.
In conclusion, our contributions can be summarized as follows:
\begin{itemize}
\item The proposal and development of PredBench, the most comprehensive evaluation framework for STP networks to date, which includes 12 methods and 15 datasets spanning multiple applications and disciplines.

\item Implementation of standardized prediction settings and novel evaluation dimensions, enhancing fairness and depth in model comparisons.

\item Unearthing key insights that offer strategic direction for future STP research.

\item Development of an open and unified codebase that will significantly promote STP research and development.
\end{itemize}

\section{Related Work}
\label{sec:related_work}

The spatio-temporal prediction has been extensively studied previously, where prevalent models can be categorized into recurrent and non-recurrent methods. 

\noindent{\textbf{Recurrent Methods.}}
ConvLSTM~\cite{shi2015convlstm} is the seminal work for recurrent methods, which uses convolutions to replace the matrix multiplication of the original LSTM~\cite{hochreiter1997lstm}. Since the introduction of ConvLSTM, a series of recurrent methods have emerged, focusing on further advancements and refinements. E3D-LSTM~\cite{wang2019e3dlstm} integrates 3D convolutions into RNNs~\cite{rumelhart1986rnn} to capture better short-term and long-term features. MAU~\cite{chang2021mau} proposes a motion-aware unit that combines an attention module and a fusion module to capture reliable inter-frame motion information. PhyDNet~\cite{Guen2020phydnet} proposes a two-branch architecture to explicitly disentangle physical dynamics from residual information. PredRNNv1~\cite{wang2017predrnn} designs a spatio-temporal LSTM unit that extracts and memorizes spatial and temporal representations simultaneously, as well as proposes a new zigzag architecture that conveys memory both vertically across layers and horizontally over states. PredRNN++~\cite{wang2018predrnn++} proposes a gradient highway unit and a causal LSTM unit to capture the short-term and the long-term video dependencies adaptively. PredRNNv2~\cite{wang2021predrnnv2} extends PredRNNv1~\cite{wang2017predrnn} by introducing a decoupling loss and a reverse scheduled sampling training strategy to boost prediction performance.

\noindent{\textbf{Non-recurrent Methods.}}
In recent years, Non-recurrent methods have gained widespread attention and demonstrated impressive performance in various domains. SimVPv1~\cite{gao2022simvp} introduces a simple encoder-translator-decoder framework for video prediction that is built purely on CNN~\cite{lecun1989cnn}, where the translator employs several Inception modules~\cite{Szegedy2015inception} to learn temporal evolution. SimVPv2~\cite{tan2022simvpv2} extends SimVPv1 by introducing a gated spatio-temporal attention module as the translator, while TAU~\cite{tan2023tau} proposes a temporal attention unit as the translator and a differential divergence regularization to capture inter-frame dynamical information. Earthformer~\cite{gao2022earthformer} is a space-time transformer proposed especially for earth system forecasting, which proposes a cuboid-attention module for generic and efficient prediction. MCVD~\cite{voleti2022mcvd} is a general framework for video prediction, generation, and interpolation, which uses a probabilistic conditional score-based denoising diffusion model~\cite{song2020score} to generate future video. 

While OpenSTL~\cite{tan2023openstl} presents an STP benchmark, its scope is limited to small datasets and lacks comprehensive analysis. PredBench significantly extends this effort by conducting exhaustive experiments and providing in-depth evaluations across expansive real-world datasets, leading to several insightful discoveries.

\section{PredBench}
\label{sec:benchmark}

\subsection{Supported Methods and Datasets}
\label{subsec:bmk_methods}
\noindent\textbf{Methods.}
PredBench accommodates diverse STP approaches, which can be categorized into recurrent-based and non-recurrent methodologies. For recurrent paradigm, we include ConvLSTM~\cite{shi2015convlstm}, E3D-LSTM~\cite{wang2019e3dlstm}, MAU~\cite{chang2021mau}, PhyDNet~\cite{Guen2020phydnet}, PredRNNv1~\cite{wang2017predrnn}, PredRNN++~\cite{wang2018predrnn++}, and PredRNNv2~\cite{wang2021predrnnv2}. The non-recurrent methods include SimVP1~\cite{gao2022simvp}, SimVP2~\cite{tan2022simvpv2}, TAU~\cite{tan2023tau}, Earthformer~\cite{gao2022earthformer}, and MCVD~\cite{voleti2022mcvd}, showing a spectrum of current methods.

\noindent\textbf{Datasets.}
PredBench spans 15 datasets to evaluate various STP scenarios. For motion trajectory prediction, \textit{Moving-MNIST}~\cite{srivastava2015movingmnist}, \textit{KTH}~\cite{schuldt2004kth}, and \textit{Human3.6M}~\cite{ionescu2013human} are incorporated. Robot action prediction is assessed through \textit{BridgeData}~\cite{walke2023bridgedata}, \textit{RoboNet}~\cite{dasari2019robonet} and \textit{BAIR}~\cite{ebert2017bair}, while driving scene prediction leverages \textit{CityScapes}~\cite{cordts2016cityscapes}, \textit{KITTI}~\cite{geiger2013kitti}, \textit{nuScenes}~\cite{caesar2020nuscenes}, and \textit{Caltech}~\cite{dollar2009caltech}. In the area of traffic flow prediction, \textit{TaxiBJ}~\cite{zhang2018taxibj} and \textit{Traffic4Cast2021}~\cite{Traffic4Cast2021} are utilized. Weather forecasting is evaluated using \textit{ICAR-ENSO}~\cite{ICARENSO}, \textit{SEVIR}~\cite{veillette2020sevir}, and \textit{Weatherbench}~\cite{garg2022weatherbench}, each contributing to a holistic assessment across the STP spectrum.

\subsection{Evaluation Metrics}
\label{subsec:bmk_metrics}
The benchmark employs tailored metrics for distinct tasks:

\noindent\textbf{Error Metrics:}
We adopt Mean Absolute Error (MAE) and Root Mean Squared Error (RMSE) as fundamental metrics to assess the discrepancy between predicted and target sequences. Additionally, Weighted Mean Absolute Percentage Error (WMAPE) is utilized specifically in traffic flow prediction, considering its relevance and effectiveness in this domain.

\noindent\textbf{Similarity Metrics:}
To gauge the resemblance between prediction and ground-truth, we use Structural Similarity Index Measure (SSIM)~\cite{wang2004ssim} and Peak Signal-to-Noise Ratio (PSNR), which provide image quality assessment.

\noindent\textbf{Perception Metrics:}
Learned Perceptual Image Patch Similarity (LPIPS)~\cite{zhang2018lpips} and Fréchet Video Distance (FVD)~\cite{unterthiner2018fvd} are employed to assess perceptual similarity in line with the human visual system. LPIPS offers a perceptually aligned comparison for individual image frames, while FVD evaluates the temporal coherence, overall quality and diversity of videos.

\noindent\textbf{Weather Metrics:}
To align with GrpahCast~\cite{lam2022graphcast}, Weighted Root Mean Squared Error (WRMSE) and Anomaly Correlation Coefficient (ACC) are used for \textit{WeatherBench}~\cite{garg2022weatherbench}. Following Earthformer~\cite{gao2022earthformer}, Critical Success Index (CSI) is applied to \textit{SEVIR}~\cite{veillette2020sevir}, and three-month-moving-averaged Nino3.4 index ($C^{Nino3.4}$) is selected for \textit{ICAR-ENSO}~\cite{ICARENSO}.

\vspace{-0.1cm}
\subsection{Standardized Experimental Protocol}
\label{subsec:bmk_std}

\begin{table}[tb]
\caption{
    \textbf{The detailed dataset statics} of the supported tasks in PredBench. $N_{train}$, $N_{val}$, $N_{test}$ are the number of sequences for training, validation and testing data respectively. The model predicts $L_{s}$ frames conditioned on $L_{in}$ frames. In certain datasets, the output of the model is extrapolated to $L_l$ frames.
    \dag: Caltech~\cite{dollar2009caltech} is only used to assess the generalization ability of models. *: ``on the fly" refers to dynamically generating training data by randomly selecting digits, their locations and directions (See details in Appendix A).
} 
\label{tab:bmk_tasks}
\renewcommand\arraystretch{1.4}
\setlength\tabcolsep{3.2pt}
\center
\vspace{-0.6cm}
\begin{adjustbox}{max width=.8\columnwidth}
\begin{tabular}{lccccccccc}
\toprule[1.3pt]
  Dataset & $N_{train}$ & $N_{val}$ & $N_{test}$ & Channel & Height & Weight & $L_{in}$ & $L_{s}$ & $L_l$ \\
  
\Xhline{0.5pt}
\multicolumn{9}{l}{\textit{Motion Trajectory Prediction}} \\
    Moving-MNIST* & on the fly & 10,000 & 10,000 & 1 & 64 & 64 & 10 & 10 & - \\
    KTH & 7,482 & 1,628 & 4,047 & 1 & 128 & 128 & 10 & 10 & - \\
    Human3.6M & 66,063 & 7,341 & 8,582 & 3 & 256 & 256 & 4 & 4 & - \\
    
\Xhline{0.5pt}
\multicolumn{9}{l}{\textit{Robot Action Planning}} \\
    BAIR & 38,937 & 4,327 & 256 & 3 & 64 & 64 & 2 & 10 & - \\
    RoboNet & 145,944 & 16,218 & 256 & 3 & 120 & 160 & 2 & 10 & - \\
    BridgeData & 31,767 & 3,970 & 3,971 & 3 & 120 & 160 & 2 & 10 & 30\\
    
\Xhline{0.5pt}
\multicolumn{9}{l}{\textit{Driving Scene Prediction}} \\
    CityScapes & 8,925 & 1,500 & 1,525 & 3 & 128 & 128 & 2 & 5 & - \\
    KITTI & 9,209 & 2,224 & 2,198 & 3 & 128 & 160 & 10 & 10 & - \\
    nuScenes & 31,269 & 4,658 & 4,518 & 3 & 128 & 160 & 10 & 10 & 30 \\
    Caltech$^{\dag}$ & N.A. & N.A. & 1,980 & 3 & 128 & 160 & 10 & 10 & - \\
    
\Xhline{0.5pt}
\multicolumn{9}{l}{\textit{Traffic Flow Prediction}} \\
    TaxiBJ & 19,961 & 500 & 500 & 2 & 32 & 32 & 4 & 4 & - \\
    Traffic4Cast2021 & 35,840 & 4,480 & 4,508 & 8 & 128 & 112 & 9 & 3 & - \\
  
\Xhline{0.5pt}
\multicolumn{9}{l}{\textit{Weather Forecasting}} \\
    ICAR-ENSO & 5,205 & 334 & 1,667 & 1 & 24 & 48 & 12 & 14 & - \\
    SEVIR & 35,718 & 9,060 & 12,159 & 1 & 384 & 384 & 13 & 12 & - \\
    WeatherBench & 53,944 & 2,922 & 5,828 & 69 & 128 & 256 & 2 & 1 & 20 \\
  
\bottomrule[1.3pt]
\end{tabular}
\end{adjustbox}
\vspace{-0.0cm}
\end{table}

In our PredBench, the experimental protocol has been meticulously standardized across various prediction tasks to ensure comparability and replicability. The detailed dataset statics and experiment settings are presented in \cref{tab:bmk_tasks}.

\noindent\textbf{Motion Trajectory Prediction:}
For \textit{Moving-MNIST}~\cite{srivastava2015movingmnist}, we adhere to conventional methods to generate training data dynamically and designate $10K$ sequences for testing. To bridge the validation set gap, we pre-generate $10K$ additional sequences. In the case of \textit{KTH}~\cite{schuldt2004kth}, where PredRNN~\cite{wang2017predrnn} lacks a validation set and uses persons 17-25 for testing, we allocate persons 1-14 for training and 15-16 for validation. Additionally, we standardize the input-output setting to match PredRNN, using 10 frames for the next 10 frames of prediction to ensure experimental consistency. \textit{Human3.6M}~\cite{ionescu2013human} evaluation, previously devoid of a validation set, now sees a division where 66,063 videos form the training set, and 7,341 serve as validation from the original 73,404 training videos.

\noindent\textbf{Robot Action Prediction:}
\textit{RoboNet}~\cite{dasari2019robonet} follows the precedent setting of~\cite{wu2021hier}, selecting 256 videos for testing and using 2 frames to predict the next 10 frames. We split the remaining data in a 9:1 ratio for training and validation to complete the experimental cycle. This experimental consistency extends to \textit{BAIR}~\cite{ebert2017bair} and \textit{BridgeData}~\cite{walke2023bridgedata}, with the latter partitioned into training, validation, and testing sets in an 8:1:1 ratio, all maintaining the 2-input to 10-output frame protocol.

\noindent\textbf{Driving Scene Prediction:}
For \textit{CityScapes}~\cite{cordts2016cityscapes}, we adopt the dataset splits of MCVD~\cite{voleti2022mcvd} but use an additional validation set for model selection, which was previously neglected. \textit{KITTI}~\cite{geiger2013kitti} and \textit{nuScenes}~\cite{caesar2020nuscenes} are segmented into training, validation, and test sets in a 9:2:2 and 8:1:1 ratio, respectively. We adjust the training protocol to predict 10 frames, departing from the coarse practices of SimVP~\cite{gao2022simvp, tan2022simvpv2} and MAU~\cite{chang2021mau}. Following previous settings~\cite{chang2021mau, gao2022simvp, tan2022simvpv2, tan2023tau}, \textit{Caltech}~\cite{dollar2009caltech} is used solely for testing to evaluate the generalization ability of models.

\noindent\textbf{Traffic Flow Prediction:}
Following PhyDNet~\cite{Guen2020phydnet}, we utilize the same testing set and randomly select 500 sequences from the training set as the validation set on \textit{TaxiBJ}~\cite{zhang2018taxibj}. For \textit{Traffic4Cast2021}~\cite{Traffic4Cast2021}, we reserve the Moscow city data for generalization evaluation and adhere to the 8:1:1 training-validation-test split. 

\noindent\textbf{Weather Forecasting:}
We harmonize our evaluation with Earthformer~\cite{gao2022earthformer} for \textit{ICAR-ENSO}~\cite{ICARENSO} and \textit{SEVIR}~\cite{veillette2020sevir}, forecasting SST anomalies and VIL, respectively, with defined context frames. For \textit{WeatherBench}~\cite{garg2022weatherbench}, we follow the previous setup~\cite{lin2022conditionalforecast,garg2022weatherbench,Rasp2023weatherbench2,lam2022graphcast,bi2023pangu,chen2023fengwu} predicting 1 frame conditioned on 2 frames with the frame interval of 6 hour and use totally 69 variables for evaluation, instead of using only 1 or 4 variables and the frame interval of 1 hour in SimVPv2~\cite{tan2022simvpv2}, training on data from 2010-2015, validating on 2016, and testing on 2017-2018.

\subsection{Multi-dimensional Evaluations}
PredBench utilizes a multi-dimensional evaluation framework that ensures thorough and detailed assessments of various spatio-temporal prediction models, providing an in-depth and exhaustive analysis of their capabilities.


\noindent{\textbf{Short-Term Prediction:}}
The short-term prediction task in PredBench focuses on forecasting imminent future states given historical data. 
A spatial state at any given time is represented as $\boldsymbol{x} \in \mathbb{R}^{C\times H \times W}$, with $C$, $H$, and $W$ denoting the channel, height, and width, respectively. The historical sequence up to time $L_{in}$ is denoted as $\mathcal{X}^{1,L_{in}} = \{\boldsymbol{x}_{1},\cdots, \boldsymbol{x}_{L_{in}}\} \in \mathbb{R}^{L_{in}\times C \times H \times W}$. From this sequence, the model is tasked with predicting the subsequent $L_s$ states, forming the predicted sequence $\hat{\mathcal{X}}^{L_{in}+1, L_{in}+L_{s}} = \{\boldsymbol{x}_{L_{in}+1},\cdots, \boldsymbol{x}_{ L_{in}+L_{s}}\} \in \mathbb{R}^{L_{s}\times C \times H \times W}$. The learning objective is to minimize the disparity between the predicted future states $\hat{\mathcal{X}}^{L{in}+1, L_{in}+L_s}$ and the actual future states $\mathcal{X}^{L_{in}+1, L_{in}+L_s}$.
During training, the model optimizes directly over $L_s$ frames, which is typically less than 15 frames in practice to ensure computational efficiency and maintain predictive accuracy. The efficacy of the model in short-term prediction is assessed on its $L_s$-frame output. We benchmark across multiple scenarios, using 14 datasets (except \textit{Caltech}~\cite{dollar2009caltech}) to evaluate short-term prediction performance comprehensively.

\noindent{\textbf{Long-Term Prediction via Extrapolation:}}
Long-term prediction ability is essential for the utility of spatio-temporal models, yet directly generating long sequences during training is hindered by prohibitive computation. Our PredBench addresses this through an extrapolation approach, where a model iteratively uses its predictions as inputs to generate further into the future. Specifically, models trained on $L_{s}$-length output sequences are tasked with predicting up to $L_{l}$ frames. This work evaluates long-term prediction on \textit{BridgeData}~\cite{walke2023bridgedata} and \textit{nuScenes}~\cite{caesar2020nuscenes} by extrapolating predictions to three times $L_s$ frames, and on \textit{WeatherBench}~\cite{garg2022weatherbench}, we extend this to a full 5-day forecast~\cite{garg2022weatherbench,lam2022graphcast}.

\noindent\textbf{Generalization Across Datasets and Scenes:}
Generalization remains a pivotal yet underexplored facet of STP research. Contrary to previous studies focusing solely on \textit{Caltech}~\cite{dollar2009caltech}, we investigate generalization across diverse datasets and scenarios. For robot action prediction, three subsets of \textit{BridgeData}~\cite{walke2023bridgedata} are segmented to evaluate model performance across new tasks and scenes. In driving scene prediction, we assess the adaptability of models trained on \textit{KITTI}~\cite{geiger2013kitti} and \textit{nuScenes}~\cite{caesar2020nuscenes} by testing on \textit{Caltech}, and reciprocally test \textit{nuScenes}-trained models on \textit{KITTI}. Traffic flow prediction challenges models to apply learned patterns from nine cities to an unseen city, Moscow, in \textit{Traffic4Cast2021}~\cite{Traffic4Cast2021}.

\noindent{\textbf{Robustness of Temporal Resolution:}}
The ability of spatio-temporal predictive models to preserve accuracy amidst changes in temporal resolution is vital. 
For instance, a weather forecasting model trained on six-hour data is also expected to perform well on data sampled every twelve hours. This type of robustness, however, is rarely assessed within the spatio-temporal prediction domain. We address this by incorporating evaluations under varying temporal resolutions, thus probing the ability of models to adapt to changes in frame rates. We formalize this by denoting the frame interval as $\Delta t$ and composing the historical sequence as $\mathcal{X}^{1,(L_{in}-1)\Delta t+1}=\{\boldsymbol{x}_{1},\boldsymbol{x}_{1+\Delta t},\cdots, \boldsymbol{x}_{(L_{in}-1)\Delta t+1}\}$. The model then predicts a future sequence with the same interval, $\hat{\mathcal{X}}^{L_{in}\Delta t+1, (L_{in}+L_{s}-1)\Delta t+1} = \{\hat{\boldsymbol{x}}_{L_{in}\Delta t+1},\hat{\boldsymbol{x}}_{(L_{in}+1)\Delta t+1},\cdots, \hat{\boldsymbol{x}}_{(L_{in}+L_{s}-1)\Delta t+1}\}$. We assess this temporal robustness on \textit{BridgeData}~\cite{walke2023bridgedata}, \textit{nuScenes}~\cite{caesar2020nuscenes}, and \textit{WeatherBench}~\cite{garg2022weatherbench}, evaluating frame intervals of 1, 2, and 3 times of the training frame interval to examine the adaptability of models to temporal resolution variations.


\section{Experiments}
\label{sec:experiments}


In pursuit of a fair comparison, we maintain the dataset setting in \cref{tab:bmk_tasks} and carefully tune the hyper-parameters for each model. See the details of the model size and experiment configuration in the appendix D (supplementary material).


\subsection{Short-Term Prediction Analysis}
\label{subsec:exp_stp}

\begin{table}[t]
\renewcommand\arraystretch{1.4}
\setlength\tabcolsep{1pt}
\center
\vspace{-0.02cm}
\caption{The \textbf{short-term prediction evaluation} of models on motion trajectory prediction. For each metric, the method with \textbf{the best} performance is highlighted in bold font, while the \uline{the second-best} performance method is indicated by underlining.}
\label{tab:exp_std_m_traj}
\vspace{-0.2cm}
\begin{adjustbox}{max width=0.8\columnwidth}
\begin{tabular}{l|cccc|cccc|cccc}
\toprule[1.3pt]
  \multirow{2}*{Method}
  &\multicolumn{4}{c|}{Moving MNIST}
  &\multicolumn{4}{c|}{KTH}
  &\multicolumn{4}{c}{Human3.6M}
  \\
  
  & \textbf{SSIM$\uparrow$} & \textbf{PSNR$\uparrow$} & \textbf{LPIPS$\downarrow$} & \textbf{FVD$\downarrow$} 
  & \textbf{SSIM$\uparrow$} & \textbf{PSNR$\uparrow$} & \textbf{LPIPS$\downarrow$} & \textbf{FVD$\downarrow$} 
  & \textbf{SSIM$\uparrow$} & \textbf{PSNR$\uparrow$} & \textbf{LPIPS$\downarrow$} & \textbf{FVD$\downarrow$} 
  \\
  
\Xhline{0.5pt}
    ConvLSTM
    & 0.9290 & 22.12 & 0.0734 & 67
    & 0.9256 & 29.14 & 0.1813 & 384
    & 0.9808 & 33.28 & 0.0416 & 187 
    \\
    
    E3D-LSTM
    & 0.9458 & 22.60 & \textbf{0.0396} & \uline{10}
    & 0.8423 & 24.19 & 0.3877 & 1525
    & 0.9822 & 32.98 & 0.0350 & 109
    \\

    MAU
    & 0.9397 & 22.76 & 0.0597 & 34
    & 0.9270 & 29.21 & 0.1769 & 288
    & 0.9810 & 33.34 & 0.0410 & 246
    \\
    
    PhyDNet
    & 0.9444 & 23.18 & \uline{0.0406} & 15
    & 0.8939 & 26.47 & 0.1926 & 402 
    & 0.9806 & 33.05 & 0.0365 & \uline{97}
    \\

    PredRNNv1
    & \uline{0.9452} & 23.18 & 0.0537 & 30
    & 0.9320 & 29.85 & 0.1765 & 330
    & 0.9824 & 33.84 & 0.0380 & 178
    \\

    PredRNN++
    & \textbf{0.9504} & \textbf{23.62} & 0.0477 & 27
    & \textbf{0.9375} & \textbf{30.22} & \uline{0.1379} & \uline{221}
    & \textbf{0.9837} & \textbf{34.11} & \uline{0.0341} & 110
    \\

    PredRNNv2
    & 0.9425 & 23.19 & 0.0520 & 28
    & \uline{0.9353} & \uline{29.90} & 0.1469 & 249
    & 0.9831 & 33.98 & 0.0389 & 167
    \\

    SimVPv1
    & 0.9268 & 21.83 & 0.0805 & 47 
    & 0.9277 & 28.80 & 0.1826 & 404 
    & 0.9823 & 33.74 & 0.0390 & 164
    \\

    SimVPv2
    & 0.9404 & 22.78 & 0.0610 & 33
    & 0.9352 & 29.13 & 0.1432 & 246
    & 0.9831 & \uline{34.01} & 0.0374 & 129
    \\

    TAU
    & 0.9443 & 23.11 & 0.0558 & 30
    & 0.9342 & 28.07 & 0.1477 & 261
    & \uline{0.9833} & 33.99 & 0.0356 & 98 
    \\

    Earthformer
    & 0.9429 & \uline{23.24} & 0.0467 & 26
    & 0.9331 & 28.99 & 0.1581 & 261
    & 0.9831 & 33.91 & 0.0394 & 167
    \\

    MCVD
    & 0.6312 & 19.12 & 0.0433 & \textbf{3}
    & 0.9304 & 28.26 & \textbf{0.0804} & \textbf{97}
    & 0.9410 & 26.33 & \textbf{0.0280} & \textbf{45}
    \\
    
\bottomrule[1.3pt]
\end{tabular}
\end{adjustbox}
\vspace{-0.0cm}
\end{table}

\begin{table}[t]
\renewcommand\arraystretch{1.4}
\setlength\tabcolsep{1pt}
\center
\vspace{-0.02cm}
\caption{The \textbf{short-term prediction evaluation} results on the robot action task. 
}
\label{tab:exp_std_robot}
\vspace{-0.3cm}
\begin{adjustbox}{max width=.8\columnwidth}
\begin{tabular}{l|cccc|cccc|cccc}
\toprule[1.3pt]
  \multirow{2}*{Method}
  &\multicolumn{4}{c|}{BAIR}
  &\multicolumn{4}{c|}{RoboNet}
  &\multicolumn{4}{c}{BridgeData}
  \\
  
  & \textbf{SSIM$\uparrow$} & \textbf{PSNR$\uparrow$} & \textbf{LPIPS$\downarrow$} & \textbf{FVD$\downarrow$} 
  & \textbf{SSIM$\uparrow$} & \textbf{PSNR$\uparrow$} & \textbf{LPIPS$\downarrow$} & \textbf{FVD$\downarrow$} 
  & \textbf{SSIM$\uparrow$} & \textbf{PSNR$\uparrow$} & \textbf{LPIPS$\downarrow$} & \textbf{FVD$\downarrow$} 
  \\
  
\Xhline{0.5pt}
    ConvLSTM
    & 0.8723 & 20.86 & 0.0874 & 723
    & 0.8362 & 22.15 & 0.1682 & 781 
    & 0.8323 & 21.36 & 0.1471 & 538 
    \\
    
    E3D-LSTM
    & 0.8724 & 20.63 & \uline{0.0769} & 722
    & 0.8265 & 21.82 & 0.1613 & 835
    & 0.7848 & 20.36 & 0.2338 & 799
    \\

    MAU
    & 0.8728 & 20.87 & 0.0853 & 746
    & 0.8436 & 22.40 & 0.1630 & 756
    & 0.8213 & 21.14 & 0.1436 & 548
    \\
    
    PhyDNet
    & 0.8509 & 20.08 & 0.0803 & 738 
    & 0.7967 & 20.92 & 0.1898 & 931
    & 0.7623 & 19.61 & 0.2055 & 963
    \\

    PredRNNv1
    & \uline{0.8767} & \uline{21.04} & 0.0849 & 701
    & 0.8497 & 22.63 & \uline{0.1587} & 727
    & 0.8528 & 22.19 & 0.1404 & 434
    \\

    PredRNN++
    & \textbf{0.8782} & \textbf{21.10} & 0.0838 & \uline{691}
    & 0.8490 & 22.66 & 0.1622 & 728
    & 0.8559 & 22.34 & 0.1402 & 415
    \\

    PredRNNv2
    & 0.8748 & 20.97 & 0.0849 & 719
    & 0.8472 & 22.52 & 0.1624 & 747
    & 0.8500 & 22.01 & 0.1428 & 436
    \\

    SimVPv1
    & 0.8733 & 20.81 & 0.0880 & 720
    & 0.8540 & 22.73 & 0.1626 & 724
    & 0.8626 & 22.60 & 0.1430 & 399
    \\

    SimVPv2
    & 0.8710 & 20.69 & 0.0898 & 762
    & \uline{0.8558} & \uline{22.78} & 0.1606 & 718
    & \uline{0.8652} & \uline{22.62} & 0.1397 & 379
    \\

    TAU
    & 0.8735 & 20.77 & 0.0885 & 732
    & \textbf{0.8567} & \textbf{22.82} & 0.1591 & \uline{720}
    & \textbf{0.8671} & \textbf{22.79} & 0.1402 & \textbf{370}
    \\

    Earthformer
    & 0.8736 & 20.84 & 0.0854 & 761
    & 0.8504 & 22.46 & 0.1640 & 728
    & 0.8618 & 22.49 & \textbf{0.1388} & \uline{372}
    \\

    MCVD
    & 0.8414 & 18.76 & \textbf{0.0640} & \textbf{113}
    & 0.7767 & 18.28 & \textbf{0.1462} & \textbf{288}
    & 0.7866 & 17.02 & \uline{0.1393} & 527
    \\
    
\bottomrule[1.3pt]
\end{tabular}
\end{adjustbox}
\vspace{-0.5cm}
\end{table}

\begin{figure}[tb]
  \centering
  \includegraphics[width=1.0\textwidth]{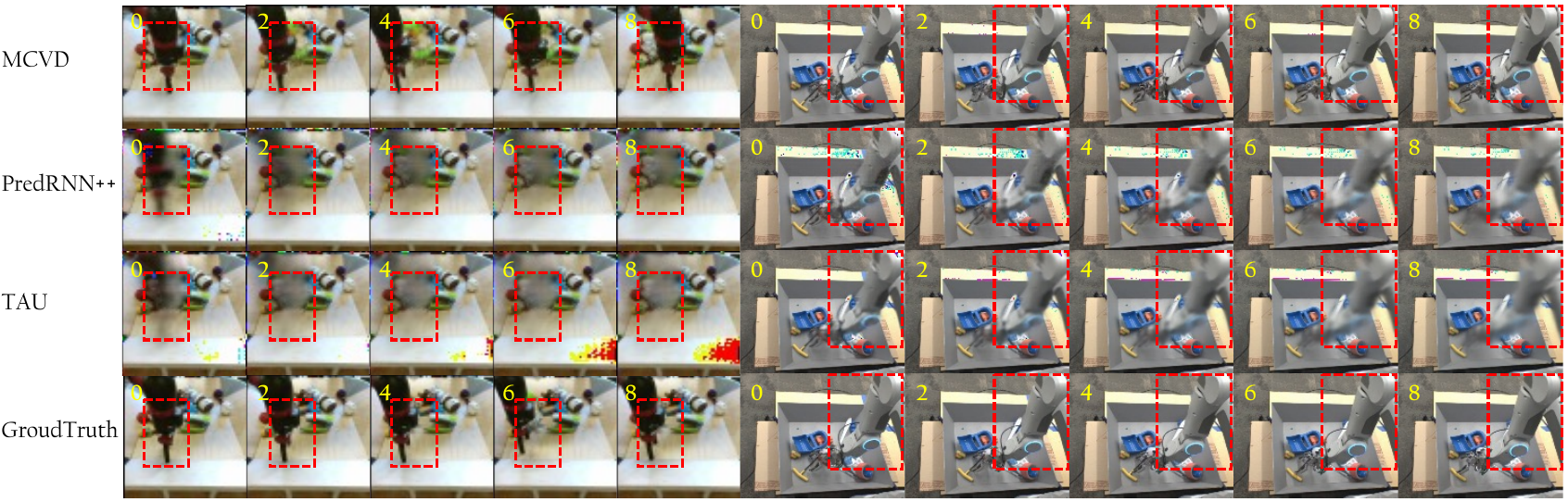}
  \vspace{-0.55cm}
  \caption{
      The \textbf{visualization results} of MCVD, PredRNN++, TAU and groudtruth on \textit{BAIR} (left) and \textit{RoboNet} (right). The \textcolor{yellow}{yellow} numbers represent frame indices. Areas where TAU and PredRNN++ exhibit ghosting are highlighted with \textcolor{red}{red} boxes. It can be observed that the output of MCVD is notably clear.
  }
  \label{fig:vs}
  \vspace{-0.2cm}
\end{figure}

\begin{table}[t]
\renewcommand\arraystretch{1.4}
\setlength\tabcolsep{1pt}
\center
\vspace{-0.02cm}
\caption{The \textbf{short-term prediction evaluation} results on the driving scenes. 
}
\vspace{-0.4cm}
\label{tab:exp_std_driving}
\begin{adjustbox}{max width=.8\columnwidth}
\begin{tabular}{l|cccc|cccc|cccc}
\toprule[1.3pt]
  \multirow{2}*{Method}
  &\multicolumn{4}{c|}{CityScapes}
  &\multicolumn{4}{c|}{KITTI}
  &\multicolumn{4}{c}{nuScenes}
  \\
  
  & \textbf{SSIM$\uparrow$} & \textbf{PSNR$\uparrow$} & \textbf{LPIPS$\downarrow$} & \textbf{FVD$\downarrow$} 
  & \textbf{SSIM$\uparrow$} & \textbf{PSNR$\uparrow$} & \textbf{LPIPS$\downarrow$} & \textbf{FVD$\downarrow$} 
  & \textbf{SSIM$\uparrow$} & \textbf{PSNR$\uparrow$} & \textbf{LPIPS$\downarrow$} & \textbf{FVD$\downarrow$} 
  \\
  
\Xhline{0.5pt}
    ConvLSTM
    & 0.8466 & 25.68 & 0.3108 & 1669
    & 0.5595 & 16.69 & 0.5622 & 1369
    & 0.7884 & 24.64 & 0.4131 & 1354
    \\
    
    E3D-LSTM
    & 0.8911 & 27.29 & 0.1507 & 641
    & 0.5337 & 15.88 & 0.5609 & 1504 
    & 0.7550 & 23.58 & 0.5621 & 2273
    \\

    MAU
    & 0.8535 & 25.85 & 0.2495 & 1469
    & 0.5794 & 16.76 & 0.4518 & 951
    & 0.7896 & 24.48 & 0.3782 & 1181
    \\
    
    PhyDNet
    & 0.8478 & 25.40 & 0.2214 & 1253
    & 0.5554 & 16.18 & \uline{0.3281} & \uline{592}
    & 0.7747 & 23.84 & 0.3149 & 893 
    \\

    PredRNNv1
    & 0.8825 & 27.09 & 0.1690 & 809 
    & 0.5816 & 17.05 & 0.5419 & 1186
    & 0.8081 & \textbf{25.27} & 0.3277 & 846
    \\

    PredRNN++
    & 0.8837 & 27.11 & 0.1545 & 758
    & 0.5184 & 16.14 & 0.6984 & 2151
    & 0.8038 & 25.10 & 0.3661 & 1015
    \\

    PredRNNv2
    & 0.8572 & 25.99 & 0.2846 & 1462
    & 0.5781 & 16.94 & 0.5559 & 1272
    & 0.7998 & 25.01 & 0.3615 & 1026
    \\

    SimVPv1
    & \uline{0.8988} & \uline{27.91} & 0.1240 & 411
    & \uline{0.5969} & \uline{17.33} & 0.5274 & 1195
    & 0.7896 & 24.42 & 0.4510 & 1394
    \\

    SimVPv2
    & 0.8572 & 25.99 & 0.2846 & 1462
    & 0.5801 & 17.09 & 0.5546 & 1300
    & \textbf{0.8120} & 24.87 & \uline{0.3073} & \uline{712}
    \\

    TAU
    & \textbf{0.9027} & \textbf{28.10} & \uline{0.1086} & \uline{367} 
    & \textbf{0.6127} & \textbf{17.59} & 0.4679 & 954
    & \uline{0.8111} & 24.98 & 0.3156 & 727
    \\

    Earthformer
    & 0.8708 & 26.49 & 0.2168 & 1149
    & 0.5859 & 17.11 & 0.5800 & 1245
    & 0.8095 & \uline{25.12} & 0.3605 & 861
    \\

    MCVD
    & 0.8165 & 19.05 & \textbf{0.0822} & \textbf{259}
    & 0.4566 & 13.62 & \textbf{0.2658} & \textbf{379}
    & 0.7197 & 20.49 & \textbf{0.1551} & \textbf{107}
    \\
    
\bottomrule[1.3pt]
\end{tabular}
\end{adjustbox}
\vspace{-0.4cm}
\end{table}

\begin{table}[t]
\renewcommand\arraystretch{1.4}
\setlength\tabcolsep{1.4pt}
\center
\vspace{-0.02cm}
\caption{The \textbf{short-term prediction evaluation} results on the traffic flow task. 
}
\label{tab:exp_std_traffic}
\vspace{-0.3cm}
\begin{adjustbox}{max width=.75\columnwidth}
\begin{tabular}{l|ccccc|ccccc}
\toprule[1.3pt]
  \multirow{2}*{Method}
  &\multicolumn{5}{c|}{TaxiBJ}
  &\multicolumn{5}{c}{Traffic4Cast2021}
  \\
  
  & \textbf{SSIM$\uparrow$} & \textbf{PSNR$\uparrow$} & \textbf{MAE$\downarrow$} & \textbf{RMSE$\downarrow$} & \textbf{WMAPE$\downarrow$}
  & \textbf{SSIM$\uparrow$} & \textbf{PSNR$\uparrow$} & \textbf{MAE$\downarrow$} & \textbf{RMSE$\downarrow$} & \textbf{WMAPE$\downarrow$}
  \\
  
\Xhline{0.5pt}
    ConvLSTM
    & \uline{0.9828} & 39.22 & 9.68 & 14.92 & 0.1305
    & 0.9265 & 30.10 & 1.367 & 8.263 & 1.211
    \\
    
    E3D-LSTM
    & 0.9803 & 39.29 & \uline{9.65} & 15.01 & 0.1375
    & 0.9222 & 30.52 & 1.694 & 8.439 & 1.883
    \\

    MAU
    & 0.9808 & 39.11 & 9.90 & 15.23 & 0.1328
    & 0.9258 & \uline{30.82} & 1.471 & 8.272 & 1.234
    \\
    
    PhyDNet
    & 0.9801 & 39.08 & 9.95 & 15.35 & 0.1399
    & 0.9247 & \textbf{30.84} & 1.369 & 8.318 & 1.247
    \\

    PredRNNv1
    & \textbf{0.9837} & \textbf{39.44} & \textbf{9.40} & \textbf{14.40} & \textbf{0.1160}
    & \uline{0.9277} & 30.17 & \textbf{1.328} & \uline{8.199} & 1.204
    \\

    PredRNN++
    & 0.9791 & 39.08 & 10.00 & 15.29 & \uline{0.1247}
    & \textbf{0.9279} & 30.19 & \uline{1.349} & \textbf{8.183} & \textbf{1.187}
    \\

    PredRNNv2
    & 0.9756 & 38.38 & 10.59 & 16.49 & 0.1323
    & 0.9264 & 30.30 & 1.370 & 8.281 & 1.255 
    \\

    SimVPv1
    & 0.9820 & 39.18 & 9.67 & 14.81 & 0.1305
    & 0.9224 & 30.67 & 1.563 & 8.304 & \uline{1.192}
    \\

    SimVPv2
    & 0.9812 & 39.21 & 9.85 & 15.00 & 0.1345
    & 0.9270 & 30.81 & 1.402 & 8.279 & 1.285
    \\

    TAU
    & 0.9813 & \uline{39.30} & 9.70 & \uline{14.77} & 0.1348
    & 0.9253 & 30.72 & 1.450 & 8.303 & 1.244
    \\

    Earthformer
    & 0.9790 & 38.85 & 10.33 & 15.70 & 0.1300
    & 0.9247 & 30.83 & 1.363 & 8.337 & 1.331
    \\

    MCVD
    & 0.9676 & 36.41 & 16.22 & 19.85 & 0.1750
    & 0.8764 & 28.19 & 2.074 & 10.89 & 2.539
    \\
    
\bottomrule[1.3pt]
\end{tabular}
\end{adjustbox}
\vspace{-0.1cm}
\end{table}

\begin{table}[t]
\renewcommand\arraystretch{1.4}
\setlength\tabcolsep{1pt}
\center
\vspace{-0.02cm}
\caption{The \textbf{short-term prediction evaluation} results on weather forecasting, with 3 representative variables out of 69 variables presented on \textit{WeatherBench}, as \cite{lam2022graphcast}.}
\vspace{-0.3cm}
\label{tab:exp_std_weather}
\begin{adjustbox}{max width=.75\columnwidth}
\begin{tabular}{l|cc|cc|cccccc}
\toprule[1.3pt]
  \multirow{3}*{Method}
  &\multicolumn{2}{c|}{\multirow{2}{*}{ICAR-ENSO}}
  &\multicolumn{2}{c|}{\multirow{2}{*}{SEVIR}}
  &\multicolumn{6}{c}{WeatherBench}
  \\ 

  & \multicolumn{2}{c|}{} & \multicolumn{2}{c|}{} 
  & \multicolumn{2}{c}{t2m} & \multicolumn{2}{c}{z500} & \multicolumn{2}{c}{t850}
  \\ 
  
  & \textbf{C$^{Nino3.4}\uparrow$} & \textbf{RMSE$\downarrow$} 
  & \textbf{CSI$\uparrow$} & \textbf{RMSE$\downarrow$} 
  & \textbf{WRMSE$\downarrow$} & \textbf{ACC$\uparrow$} & \textbf{WRMSE$\downarrow$} & \textbf{ACC$\uparrow$} & \textbf{WRMSE$\downarrow$} & \textbf{ACC$\uparrow$} 
  \\
  
\Xhline{0.5pt}
    ConvLSTM
    & \uline{0.7475} & 0.4158 
    & 0.4082 & 13.01
    & 2.0896 & 0.9827 & 91.20 & 0.9985 & \uline{1.2276} & \uline{0.9880}
    \\
    
    E3D-LSTM
    & 0.7187 & 0.4235
    & 0.3984 & 13.92
    & 1.9809 & 0.9846 & 79.17 & \uline{0.9988} & 1.3508 & 0.9854
    \\

    MAU
    & \textbf{0.7578} & \textbf{0.4105} 
    & 0.4122 & 12.89
    & 2.2038 & 0.9807 & 120.2 & 0.9974 & 1.4316 & 0.9837
    \\
    
    PhyDNet
    & 0.7334 & 0.4252 
    & 0.4281 & 13.60
    & 9.6417 & 0.6433 & 1698 & 0.5782 & 7.2720 & 0.6406
    \\

    PredRNNv1
    & 0.7391 & 0.4156 
    & 0.4288 & 12.82
    & 2.0570 & 0.9832 & 81.54 & \uline{0.9988} & 1.2327 & 0.9879
    \\

    PredRNN++
    & 0.7386 & 0.4215 
    & 0.4312 & \uline{12.64} 
    & 2.0540 & 0.9833 & \uline{79.08} & \uline{0.9988} & 1.2309 & 0.9880
    \\

    PredRNNv2
    & 0.7369 & \uline{0.4137}
    & \uline{0.4347} & \textbf{12.24}
    & 2.0760 & 0.9830 & 85.27 & 0.9987 & 1.2435 & 0.9878
    \\

    SimVPv1
    & 0.7273 & 0.4291 
    & 0.3959 & 12.66
    & 1.9151 & 0.9855 & 93.04 & 0.9984 & 1.2855 & 0.9870
    \\

    SimVPv2
    & 0.7450 & 0.4309 
    & 0.3841 & 12.83
    & 2.0309 & 0.9838 & 89.95 & 0.9985 & 1.2804 & 0.9870
    \\

    TAU
    & 0.7053 & 0.4300 
    & 0.3941 & 12.73
    & \uline{1.8240} & \uline{0.9869} & 81.78 & \uline{0.9988} & \uline{1.2116} & \uline{0.9884}
    \\

    Earthformer
    & 0.7020 & 0.4225 
    & \textbf{0.4391} & 12.85
    & \textbf{1.5875} & \textbf{0.9901} & \textbf{78.68} & \textbf{0.9989} & \textbf{1.0435} & \textbf{0.9914}
    \\

    MCVD
    & 0.6113 & \textbf{0.4105}
    & 0.0831 & 130.9
    & 6.4596 & 0.8391 & 1555 & 0.6448 & 5.2465 & 0.8071
    \\
    
\bottomrule[1.3pt]
\end{tabular}
\end{adjustbox}
\vspace{-0.3cm}
\end{table}





The results of 5 short-term prediction tasks are shown in Tabs.~\ref{tab:exp_std_m_traj}, \ref{tab:exp_std_robot}, \ref{tab:exp_std_driving}, \ref{tab:exp_std_traffic}, and \ref{tab:exp_std_weather}.

\noindent{\textbf{Observation 1}}: 
Across all datasets within the motion trajectory prediction domain and the BAIR dataset~\cite{ebert2017bair}, PredRNN++~\cite{wang2018predrnn++} achieves optimal outcomes in the SSIM and PSNR metrics. It also excels in the SSIM, RMSE, and WMAPE metrics on Traffic4Cast2021~\cite{Traffic4Cast2021}, and shows comparable results on other datasets. 

\noindent{\textbf{Finding 1}}: 
Although many years have passed since the introduction of PredRNN++, it still demonstrates remarkable performance across multiple datasets, making it suitable for a good baseline on spatio-temporal prediction.

\noindent{\textbf{Observation 2}}: 
Despite poor performance on SSIM and PSNR metrics, MCVD~\cite{voleti2022mcvd} has the best results of LPIPS and FVD on motion trajectory prediction, robot action prediction, and driving scenes prediction domains, except on \textit{BridgeData}~\cite{walke2023bridgedata}. ~\cref{fig:vs} shows that MCVD exhibits the highest visual quality, notwithstanding its lower SSIM and PSNR scores.

\noindent{\textbf{Finding 2}}:  Visualization in ~\cref{fig:vs}, coupled with additional human-based analyses (in appendix F), underscores that the LPIPS and FVD metrics are more aptly suited for tasks involving visual prediction.

\noindent{\textbf{Observation 3}}: 
For weather forecasting tasks, MAU~\cite{chang2021mau} performs the best on \textit{ICAR-ENSO}, while Earthformer~\cite{gao2022earthformer} excels on \textit{SEVIR} and \textit{WeatherBench}.

\noindent{\textbf{Finding 3}}: Given the high-resolution nature of data in SEVIR~\cite{veillette2020sevir} and the intricate meteorological information in WeatherBench~\cite{garg2022weatherbench}, spatio-temporal prediction models face considerable challenges in accurately capturing such dynamic complexity. However, Earthformer~\cite{gao2022earthformer} emerges as a standout performer on these datasets compared to CNN-based methods like SimVP~\cite{gao2022simvp,tan2022simvpv2} or RNN-based methods like PredRNN~\cite{wang2017predrnn,wang2018predrnn++,wang2021predrnnv2}. This highlights the superior capability of transformer architectures in effectively modeling the dynamic patterns inherent in meteorological data, surpassing the performance of both convolutional neural networks and recurrent neural networks.


\noindent{\textbf{Observation 4}}: 
In conclusion, for short-term prediction tasks, PredRNN++~\cite{wang2018predrnn++} demonstrates superior performance in the motion trajectory domain. MCVD~\cite{voleti2022mcvd} emerges as the leading choice involving the \textit{BAIR}~\cite{ebert2017bair}, \textit{RoboNet}~\cite{dasari2019robonet}, and driving scene domains. TAU~\cite{tan2023tau} showcases its dominance in \textit{BridgeData}. In the traffic flow domain, PredRNNv1~\cite{wang2017predrnn} and PredRNN++~\cite{wang2018predrnn++} stand out as the premier models. For weather forecasting, MAU is the most effective for \textit{ICAR-ENSO}~\cite{ICARENSO}, while Earthformer~\cite{gao2022earthformer} takes the lead in \textit{SEVIR}~\cite{veillette2020sevir} and \textit{WeatherBench}~\cite{garg2022weatherbench}. In addition, some methods may excel on specific metrics for certain datasets, e.g., PhyDNet~\cite{Guen2020phydnet} achieves the highest PSNR on \textit{Traffic4Cast2021}~\cite{Traffic4Cast2021}.

\noindent{\textbf{Finding 4}}: Previously, most studies utilize motion trajectory prediction datasets for experimentation, but we find that performance on these datasets does not reliably indicate true performance on some larger real-world STP datasets, e.g., \textit{BridgeData}, \textit{nuScenes}, \textit{SEVIR} and \textit{WeatherBench}.
Adopting a more holistic perspective, the variation in patterns observed across different datasets, in conjunction with the varied focal points of distinct evaluative metrics, there is no single method that excels in all tasks and metrics. Consequently, various methodologies demonstrate their unique strengths and advantages.

\vspace{-0.3cm}
\subsection{Long-Term Prediction Analysis}
\label{subsec:exp_ltp}
\vspace{-0.2cm}

\begin{figure}[tb]
  \centering
  \includegraphics[width=1.0\textwidth]{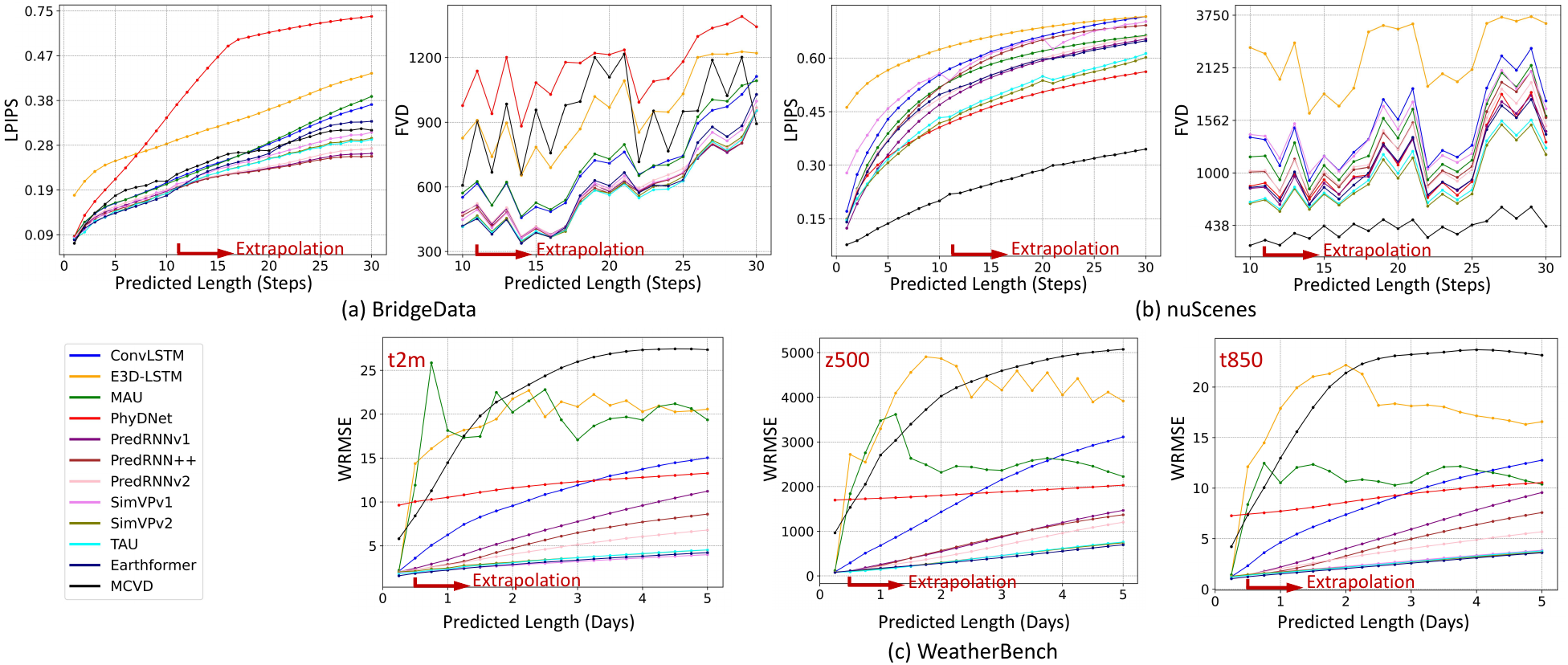}
  \vspace{-0.6cm}
  \caption{
      \textbf{Extrapolation results} on (a) \textit{BridgeData}, (b) \textit{nuScenes}, and (c) \textit{Weatherbench}. 
        Prediction results beyond the tenth frame on \textit{BridgeData} and \textit{nuScenes}, as well as weather forecasting results after the first frame, are generated by extrapolation.
  }
  \label{fig:extrapolation}
  \vspace{-0.4cm}
  
\end{figure}




The quantitative results for long-term prediction results are shown in \cref{fig:extrapolation}. 

\noindent{\textbf{Observation 5}}: 
While TAU~\cite{tan2023tau} excels in short-term prediction, PredRNN++\cite{wang2018predrnn++} outperforms in long-term performance on BridgeData\cite{walke2023bridgedata}. MCVD~\cite{voleti2022mcvd} shows remarkable results for both short and long-term prediction validation on nuScenes~\cite{caesar2020nuscenes}.


\noindent{\textbf{Finding 5}}: Models performing the best in the short-term prediction may not necessarily yield the best results in the long-term prediction evaluation.

\noindent{\textbf{Observation 6}}: 
On \textit{WeatherBench}~\cite{garg2022weatherbench}, Earthformer~\cite{gao2022earthformer} demonstrates superior performance. Surprisingly, non-recurrent methods such as Earthformer, SimVP~\cite{gao2022simvp,tan2022simvpv2}, and TAU~\cite{tan2023tau} exhibit better extrapolation performance compared to recurrent-based methods such as ConvLSTM~\cite{shi2015convlstm}, E3D-LSTM~\cite{wang2019e3dlstm}, and MAU~\cite{chang2021mau}.

\noindent{\textbf{Finding 6}}: 
The auto-regressive paradigm employed for extrapolation doesn't guarantee superior performance for recurrent methods over non-recurrent methods. we postulate that training on \textit{WeatherBench} with optimization for only one output frame might limit the extrapolation capability of RNN methods.

\subsection{Generalization Ability Analysis}

\begin{table}[t]
\renewcommand\arraystretch{1.4}
\setlength\tabcolsep{1pt}
\center
\vspace{-0.1cm}
\caption{The \textbf{generalization ability evaluation} results across new tasks and scenes on \textit{BridgeData}. Notably, the sequences containing new scenes or new tasks have not appeared in the training data. All three testing data have approximately 1,000 sequences.
}
\label{tab:exp_gen_robot}
\vspace{-0.3cm}
\begin{adjustbox}{max width=.8\columnwidth}
\begin{tabular}{l|cccc|cccc|cccc}
\toprule[1.3pt]
  \multirow{2}*{Method}
  &\multicolumn{4}{c|}{Original Scene, New Task}
  &\multicolumn{4}{c|}{New Scene, Original Task}
  &\multicolumn{4}{c}{New Scene, New Task}
  \\
  
  & \textbf{SSIM$\uparrow$} & \textbf{PSNR$\uparrow$} & \textbf{LPIPS$\downarrow$} & \textbf{FVD$\downarrow$} 
  & \textbf{SSIM$\uparrow$} & \textbf{PSNR$\uparrow$} & \textbf{LPIPS$\downarrow$} & \textbf{FVD$\downarrow$} 
  & \textbf{SSIM$\uparrow$} & \textbf{PSNR$\uparrow$} & \textbf{LPIPS$\downarrow$} & \textbf{FVD$\downarrow$} 
  \\
  
\Xhline{0.5pt}
    ConvLSTM
    & 0.8554 & 21.81 & 0.1185 & 741
    & 0.7882 & 17.90 & 0.1707 & 596
    & 0.7896 & 18.45 & 0.1731 & 700
    \\
    
    E3D-LSTM
    & 0.8045 & 20.50 & 0.2139 & 1064
    & 0.7483 & 17.27 & 0.2556 & 837
    & 0.7495 & 17.80 & 0.2523 & 976
    \\

    MAU
    & 0.8445 & 21.55 & 0.1126 & 767
    & 0.7775 & 17.80 & 0.1684 & 669
    & 0.7752 & 18.25 & 0.1748 & 742
    \\
    
    PhyDNet
    & 0.7809 & 19.93 & 0.1757 & 1422
    & 0.7122 & 17.17 & 0.2277 & 1159
    & 0.7094 & 17.41 & 0.2387 & 1357
    \\

    PredRNNv1
    & 0.8783 & \uline{22.67} & \textbf{0.1076} & 569
    & 0.8133 & \textbf{18.27} & \uline{0.1557} & 600
    & 0.8048 & \uline{18.43} & 0.1696 & 660
    \\

    PredRNN++
    & 0.8778 & 22.46 & \textbf{0.1076} & 542
    & 0.8066 & 17.86 & 0.1666 & 616
    & 0.7991 & 18.12 & 0.1817 & 681
    \\

    PredRNNv2
    & 0.8764 & \textbf{22.54} & 0.1100 & 571
    & \uline{0.8085} & \uline{18.10} & 0.1620 & 597
    & 0.8025 & 18.34 & 0.1732 & 650
    \\

    SimVPv1
    & 0.8828 & 22.48 & 0.1118 & 508
    & 0.8145 & 17.61 & 0.1616 & 557
    & \uline{0.81} & 18.31 & \textbf{0.1705} & \uline{573}
    \\

    SimVPv2
    & \textbf{0.8846} & 22.35 & 0.1093 & 494
    & 0.8120 & 17.46 & 0.1601 & 583
    & 0.8091 & 17.94 & 0.1733 & 586
    \\

    TAU
    & \uline{0.8839} & 22.27 & 0.1109 & \uline{484}
    & 0.8083 & 17.49 & 0.1646 & \uline{569}
    & 0.8116 & 18.36 & \uline{0.1722} & \textbf{571}
    \\

    Earthformer
    & 0.8821 & 22.40 & \uline{0.1080} & \textbf{483}
    & \textbf{0.8227} & 17.93 & \textbf{0.1518} & \textbf{566}
    & \textbf{0.8154} & \textbf{18.55} & 0.1725 & 592
    \\

    MCVD
    & 0.8048 & 17.07 & 0.1215 & 739
    & 0.7467 & 14.25 & 0.1740 & 701
    & 0.7379 & 14.55 & 0.1953 & 585
    \\
    
\bottomrule[1.3pt]
\end{tabular}
\end{adjustbox}
\vspace{-0.4cm}
\end{table}

\begin{table}[t]
\renewcommand\arraystretch{1.4}
\setlength\tabcolsep{1pt}
\center
\vskip 0.15in
\caption{The \textbf{generalization ability evaluation} results on driving scenes. 
The models trained on \textit{KITTI} and \textit{nuScenes} are evaluated on both the \textit{Caltech} and \textit{KITTI}.
}
\label{tab:exp_gen_driving}
\vspace{-0.3cm}
\begin{adjustbox}{max width=.8\columnwidth}
\begin{tabular}{l|cccc|cccc|cccc}
\toprule[1.3pt]
  \multirow{2}*{Method}
  &\multicolumn{4}{c|}{KITTI $\rightarrow$ Caltech}
  &\multicolumn{4}{c|}{nuScenes $\rightarrow$ Caltech}
  &\multicolumn{4}{c}{nuScenes $\rightarrow$ KITTI}
  \\
  
  & \textbf{SSIM$\uparrow$} & \textbf{PSNR$\uparrow$} & \textbf{LPIPS$\downarrow$} & \textbf{FVD$\downarrow$} 
  & \textbf{SSIM$\uparrow$} & \textbf{PSNR$\uparrow$} & \textbf{LPIPS$\downarrow$} & \textbf{FVD$\downarrow$} 
  & \textbf{SSIM$\uparrow$} & \textbf{PSNR$\uparrow$} & \textbf{LPIPS$\downarrow$} & \textbf{FVD$\downarrow$} 
  \\
  
\Xhline{0.5pt}
    ConvLSTM
    & 0.7121 & 19.54 & 0.4620 & 1183
    & 0.7588 & 21.06 & 0.3577 & 846
    & 0.5464 & 16.10 & 0.5741 & 1401
    \\
    
    E3D-LSTM
    & 0.6498 & 17.82 & 0.4951 & 1573
    & 0.6957 & 19.27 & 0.5244 & 1753
    & 0.5078 & 15.32 & 0.6589 & 2056
    \\

    MAU
    & 0.7257 & 19.55 & 0.3422 & 752
    & 0.7557 & 20.75 & 0.3267 & 839
    & 0.5398 & 15.69 & 0.5505 & 1611
    \\
    
    PhyDNet
    & 0.6905 & 18.28 & \uline{0.2390} & \textbf{419}
    & 0.7385 & 20.07 & 0.2819 & 676
    & 0.5218 & 15.18 & 0.5022 & 1369
    \\

    PredRNNv1
    & 0.7413 & 20.35 & 0.4002 & 964
    & 0.7622 & 21.23 & 0.2891 & 499
    & 0.5636 & \textbf{16.53} & 0.4775 & 1046
    \\

    PredRNN++
    & 0.6712 & 19.15 & 0.5752 & 2332
    & 0.7628 & 21.19 & 0.3179 & 577
    & 0.5573 & \uline{16.52} & 0.5164 & 1092
    \\

    PredRNNv2
    & 0.7267 & 20.02 & 0.4638 & 1148
    & 0.7450 & 20.65 & 0.3487 & 711
    & 0.5538 & 16.30 & 0.5221 & 1172
    \\

    SimVPv1
    & 0.7696 & 20.52 & 0.3320 & 689
    & 0.7748 & 20.93 & 0.3673 & 726
    & 0.5551 & 15.54 & 0.5844 & 1362
    \\

    SimVPv2
    & \uline{0.7796} & \uline{21.16} & 0.3370 & 768
    & \uline{0.7915} & \textbf{21.64} & \uline{0.2554} & 390
    & \textbf{0.5760} & 15.92 & \uline{0.4562} & \uline{903} 
    \\

    TAU
    & \textbf{0.7879} & \textbf{21.21} & 0.2763 & 455
    & \textbf{0.7937} & \uline{21.57} & 0.2626 & \uline{381}
    & \uline{0.5705} & 15.85 & 0.4781 & 931
    \\

    Earthformer
    & 0.7611 & 20.61 & 0.3714 & 726
    & 0.7731 & 21.29 & 0.3129 & 582
    & 0.5692 & 16.01 & 0.5414 & 1151
    \\

    MCVD
    & 0.6630 & 16.17 & \textbf{0.2226} & \uline{462}
    & 0.6588 & 17.73 & \textbf{0.1981} & \textbf{193}
    & 0.4479 & 13.57 & \textbf{0.3005} & \textbf{322}
    \\
    
\bottomrule[1.3pt]
\end{tabular}
\end{adjustbox}
\vspace{-0.4cm}
\end{table}

\begin{table}[t]
\renewcommand\arraystretch{1.4}
\setlength\tabcolsep{3.2pt}
\center
\vspace{-0.1cm}
\caption{
The \textbf{generalization ability evaluation} of models on \textit{Traffic4Cast2021}. We evaluate models trained on data from nine different cities for their performance on Moscow-city data (containing 2576 sequences).
} 
\label{tab:exp_gen_traffic}
\vspace{-0.3cm}
\begin{adjustbox}{max width=0.5\columnwidth}
\begin{tabular}{l|ccccc}
\toprule[1.3pt]
  \multirow{2}*{Method}
  &\multicolumn{5}{c}{Traffic4Cast2021}
  \\
  
  & \textbf{SSIM$\uparrow$} & \textbf{PSNR$\uparrow$} & \textbf{MAE$\downarrow$} & \textbf{RMSE$\downarrow$} & \textbf{WMAPE$\downarrow$}
  \\
  
\Xhline{0.5pt}
    ConvLSTM
    & 0.6722 & 23.30 & 6.886 & 17.50 & 1.535
    \\
    
    E3D-LSTM
    & 0.6369 & 22.93 & 7.704 & 18.26 & 1.720
    \\

    MAU
    & 0.6736 & 22.93 & 6.006 & 18.265 & 1.340
    \\
    
    PhyDNet
    & 0.6897 & 23.18 & \uline{5.189} & 17.75 & \uline{1.157}
    \\

    PredRNNv1
    & 0.6735 & 23.54 & 7.096 & 17.02 & 1.582
    \\

    PredRNN++
    & 0.6611 & 23.29 & 7.187 & 17.51 & 1.603
    \\

    PredRNNv2
    & 0.6757 & 23.25 & 6.652 & 17.60 & 1.486
    \\

    SimVPv1
    & 0.6638 & 23.18 & 7.072 & 17.72 & 1.578
    \\

    SimVPv2
    & 0.7113 & 23.93 & 6.765 & 16.260 & 1.51
    \\

    TAU
    & \uline{0.7386} & \uline{24.33} & 6.484 & \uline{15.52} & 1.448
    \\

    Earthformer
    & \textbf{0.8216} & \textbf{26.13} & \textbf{4.372} & \textbf{12.59} & \textbf{0.989}
    \\

    MCVD
    & 0.6769 & 22.84 & 5.528 & 18.45 & 1.234
    \\
    
\bottomrule[1.3pt]
\end{tabular}
\end{adjustbox}
\vspace{-0.1cm}
\end{table}





The generalization evaluation results are demonstrated in Tabs.~\ref{tab:exp_gen_robot}, ~\ref{tab:exp_gen_driving}, and ~\ref{tab:exp_gen_traffic}.

\noindent{\textbf{Observation 7}}: 
For robot action prediction in ~\cref{tab:exp_gen_robot}, we observe a significant decline in the performance of models when encountering previously unseen scenes, while this phenomenon does not occur in new tasks within seen scenes.

\noindent{\textbf{Finding 7}}: 
Robot action prediction involves complex backgrounds and diverse tasks. It seems that the models often capture the scene context but struggle to learn the specific task dynamics, e.g., the movement of robotic arms or objects. 

\noindent{\textbf{Observation 8}}: 
Interestingly, when comparing the generalization results in ~\cref{tab:exp_gen_driving} with the short-term prediction results in ~\cref{tab:exp_std_driving}, almost all models exhibit better performance on \textit{Caltech}~\cite{dollar2009caltech} than on their testing set of \textit{KITTI}~\cite{geiger2013kitti} or \textit{nuScenes}~\cite{caesar2020nuscenes}.

\noindent{\textbf{Finding 8}}: \textit{Caltech} dataset is relatively simplistic, evidenced by the fact that models often outperform their original testing datasets. Therefore, relying solely on the evaluation with \textit{Caltech} is inadequate to fully assess model performance. The previous experimental setting~\cite{chang2021mau,gao2022simvp,tan2022simvpv2,tan2023tau} where models are trained on \textit{KITTI} and evaluated on \textit{Caltech} is deemed unreasonable.

\noindent{\textbf{Observation 9}}: 
When evaluating models on \textit{Caltech}~\cite{dollar2009caltech}, models trained on \textit{nuScenes}~\cite{caesar2020nuscenes} universally outperform those trained on \textit{KITTI}~\cite{geiger2013kitti}. Notably, the training set of \textit{nuScenes} is three times larger than \textit{KITTI}.

\noindent{\textbf{Finding 9}}: Expanding the scale of the training dataset does improve the generalization of the model.

\noindent{\textbf{Observation 10}}: 
Earthformer~\cite{gao2022earthformer} maintains a significant lead over other methods on \textit{BridgeData}~\cite{walke2023bridgedata} and \textit{Traffic4Cast2021}~\cite{Traffic4Cast2021}, despite not having the best short-term prediction performance. 

\noindent{\textbf{Finding 10}}: It is not necessarily true that the model with the best short-term prediction capability will also have the best generalization ability.

\subsection{Robustness Analysis}
\label{subsec:exp_rbs}

\begin{table*}[t]
\renewcommand\arraystretch{1.4}
\setlength\tabcolsep{1pt}
\center
\caption{\textbf{The robustness evaluation} on \textit{BridgeData}, \textit{nuScenes}. Only results of $\Delta t=2$ and $\Delta t=3$ are presented, where the frame interval is denoted as $\Delta t$. 
}
\label{tab:exp_rbs}
\vspace{-0.3cm}
\begin{adjustbox}{max width=1\columnwidth}
\begin{tabular}{l|cccc|cccc|cccc|cccc}
\toprule[1.3pt]
  \multirow{2}*{Method}
  &\multicolumn{4}{c|}{BridgeData ($\Delta t=2$)}
  &\multicolumn{4}{c|}{BridgeData ($\Delta t=3$)}
  &\multicolumn{4}{c|}{nuScenes ($\Delta t=2$)}
  &\multicolumn{4}{c}{nuScenes ($\Delta t=3$)}
  \\
  
  & \textbf{SSIM$\uparrow$} & \textbf{PSNR$\uparrow$} & \textbf{LPIPS$\downarrow$} & \textbf{FVD$\downarrow$} 
  & \textbf{SSIM$\uparrow$} & \textbf{PSNR$\uparrow$} & \textbf{LPIPS$\downarrow$} & \textbf{FVD$\downarrow$} 
  & \textbf{SSIM$\uparrow$} & \textbf{PSNR$\uparrow$} & \textbf{LPIPS$\downarrow$} & \textbf{FVD$\downarrow$} 
  & \textbf{SSIM$\uparrow$} & \textbf{PSNR$\uparrow$} & \textbf{LPIPS$\downarrow$} & \textbf{FVD$\downarrow$} 
  \\ 
\Xhline{0.5pt}
    ConvLSTM
    & 0.7805 & 19.17 & 0.2027 & 711
    & 0.7639 & 18.01 & 0.2213 & 790
    & 0.7368 & 22.64 & 0.5043 & 1619 
    & 0.7066 & 21.37 & 0.5578 & 1819
    \\
    
    E3D-LSTM
    & 0.7412 & 18.51 & 0.2790 & 1032
    & 0.7224 & 17.53 & 0.2984 & 1124
    & 0.7175 & 22.00 & 0.5970 & 2356
    & 0.6919 & 20.85 & 0.6190 & 2393
    \\

    MAU
    & 0.7738 & 19.15 & 0.1978 & 714
    & 0.7552 & 17.98 & 0.2159 & 780
    & 0.7366 & 22.47 & 0.4603 & 1439
    & 0.7043 & 21.16 & 0.5141 & 1625
    \\
    
    PhyDNet
    & 0.7201 & 18.16 & 0.2594 & 1283
    & 0.7068 & 17.27 & 0.2758 & 1381 
    & 0.7199 & 21.95 & 0.4102 & 1172
    & 0.6892 & 20.79 & 0.4706 & 1403
    \\

    PredRNNv1
    & 0.7956 & 19.50 & 0.1927 & 553
    & 0.7785 & 18.21 & 0.2030 & 613
    & 0.7463 & \uline{22.92} & 0.4259 & 1155
    & 0.7107 & \uline{21.51} & 0.4877 & 1398
    \\

    PredRNN++
    & 0.7980 & 19.55 & 0.1914 & 538
    & 0.7807 & 18.25 & \uline{0.1977} & 590 
    & 0.7420 & 22.77 & 0.4619 & 1355
    & 0.7064 & 21.38 & 0.5207 & 1607
    \\

    PredRNNv2
    & 0.7961 & 19.52 & 0.1935 & 574
    & 0.7799 & 18.27 & 0.2037 & 645
    & 0.7448 & \textbf{22.93} & 0.4551 & 1317
    & 0.7104 & \textbf{21.55} & 0.5088 & 1535
    \\

    SimVPv1
    & 0.8024 & 19.72 & 0.1943 & 544 
    & 0.7812 & \textbf{18.32} & 0.2082 & 597
    & 0.7405 & 22.57 & 0.5246 & 1701
    & 0.7099 & 21.32 & 0.5670 & 1889
    \\

    SimVPv2
    & 0.8047 & 19.70 & \uline{0.1916} & 528
    & \uline{0.7830} & \textbf{18.32} & 0.2039 & 574
    & \uline{0.7499} & 22.61 & \uline{0.4021} & \uline{952}
    & 0.7120 & 21.20 & \uline{0.4609} & \uline{1166}
    \\

    TAU
    & \textbf{0.8056} & \textbf{19.74} & 0.1917 & 503 
    & \uline{0.7830} & \uline{18.29} & 0.2037 & 556
    & \textbf{0.7500} & 22.73 & 0.4110 & 964
    & \uline{0.7122} & 21.30 & 0.4702 & 1169
    \\

    Earthformer
    & \uline{0.8050} & \uline{19.73} & \textbf{0.1881} & \uline{498}
    & \textbf{0.7852} & \uline{18.29} & \textbf{0.1945} & \uline{523}
    & 0.7487 & 22.77 & 0.4526 & 1175
    & \textbf{0.7140} & 21.38 & 0.5044 & 1396
    \\

    MCVD
    & 0.7264 & 15.38 & 0.1944 & \textbf{342}
    & 0.7163 & 14.75 & 0.2144 & \textbf{356}
    & 0.6597 & 19.07 & \textbf{0.2078} & \textbf{115}
    & 0.6226 & 18.14 & \textbf{0.2537} & \textbf{140} 
    \\
    
\bottomrule[1.3pt]
\end{tabular}
\end{adjustbox}
\vspace{-0.4cm}
\end{table*}

\begin{table*}[t]
\renewcommand\arraystretch{1.4}
\setlength\tabcolsep{1pt}
\center
\caption{\textbf{The robustness evaluation} on \textit{WeatherBench}. Note that 6-frame interval is used in training, so the results of $\Delta t=12$ and $\Delta t=18$ are demonstrated. }
\label{tab:exp_weather_rbs}
\vspace{-0.5cm}
\begin{adjustbox}{max width=1.\columnwidth}
\begin{tabular}{l|cccccc|cccccc}
\toprule[1.3pt]
  \multirow{3}*{Method}
  &\multicolumn{6}{c|}{WeatherBench ($\Delta t =12$)}
  &\multicolumn{6}{c}{WeatherBench ($\Delta t =18$)}
  \\
  
  & \multicolumn{2}{c}{t2m} & \multicolumn{2}{c}{z500} & \multicolumn{2}{c|}{t850}
  & \multicolumn{2}{c}{t2m} & \multicolumn{2}{c}{z500} & \multicolumn{2}{c}{t850}
  \\ 

  & \textbf{WRMSE$\downarrow$} & \textbf{ACC$\uparrow$} 
  & \textbf{WRMSE$\downarrow$} & \textbf{ACC$\uparrow$} 
  & \textbf{WRMSE$\downarrow$} & \textbf{ACC$\uparrow$} 
  & \textbf{WRMSE$\downarrow$} & \textbf{ACC$\uparrow$} 
  & \textbf{WRMSE$\downarrow$} & \textbf{ACC$\uparrow$} 
  & \textbf{WRMSE$\downarrow$} & \textbf{ACC$\uparrow$} 
  \\
  
\Xhline{0.5pt}
    ConvLSTM
    & 4.0130 & 0.9368 & 212.59 & 0.9916 & 1.9073 & 0.9713 
    & 3.4176 & 0.9537 & 374.16 & 0.9739 & 2.3873 & 0.9543 
    \\
    
    E3D-LSTM
    & 3.7413 & 0.9453 & \uline{209.33} & \uline{0.9918} & 1.9389 & 0.9701 
    & \textbf{2.8937} & \textbf{0.9671} & 366.73 & 0.9747 & \uline{2.3544} & \uline{0.9551} 
    \\

    MAU
    & 3.8471 & 0.9415 & 229.25 & 0.9902 & 1.9929 & 0.9686 
    & 3.1256 & 0.9612 & 373.83 & 0.9738 & 2.4075 & 0.9534 
    \\
    
    PhyDNet
    & 9.8750 & 0.6267 & 1700.8 & 0.5774 & 7.3076 & 0.6369 
    & 9.7852 & 0.6331 & 1706.7 & 0.5745 & 7.3477 & 0.6324 
    \\

    PredRNNv1
    & 4.1033 & 0.9342 & 212.05 & 0.9917 & 1.8992 & 0.9715 
    & 3.7964 & 0.9433 & 369.69 & 0.9747 & 2.4334 & 0.9526 
    \\

    PredRNN++
    & 3.8997 & 0.9401 & 213.74 & 0.9915 & \uline{1.8639} & \uline{0.9725} 
    & 3.4307 & 0.9530 & \uline{363.82} & \uline{0.9754} & 2.3642 & \uline
    {0.9551} 
    \\

    PredRNNv2
    & 3.9258 & 0.9397 & 230.41 & 0.9901 & 1.8756 & 0.9720 
    & \uline{3.1144} & \uline{0.9616} & 386.23 & 0.9721 & 2.3735 & 0.9543 
    \\

    SimVPv1
    & \textbf{3.5724} & \textbf{0.9497} & 217.02 & 0.9913 & 1.9397 & 0.9702 
    & 3.6105 & 0.9485 & 370.19 & 0.9743 & 2.4541 & 0.9517 
    \\

    SimVPv2
    & 3.9050 & 0.9400 & 213.54 & 0.9916 & 1.9465 & 0.9701 
    & 3.6083 & 0.9487 & 374.07 & 0.9740 & 2.4319 & 0.9527 
    \\

    TAU
    & \uline{3.8040} & \uline{0.9434} & 212.77 & 0.9917 & 1.9271 & 0.9706 
    & 3.5750 & 0.9497 & 371.25 & 0.9744 & 2.4179 & 0.9529 
    \\

    Earthformer
    & 3.9257 & 0.9401 & \textbf{204.16} & \textbf{0.9923} & \textbf{1.8052} & \textbf{0.9744} 
    & 3.2683 & 0.9584 & \textbf{356.39} & \textbf{0.9764} & \textbf{2.3082} & \textbf{0.9573} 
    \\

    MCVD
    & 6.7007 & 0.8260 & 1572.2 & 0.6370 & 5.3949 & 0.7950 
    & 6.8436 & 0.8182 & 1588.7 & 0.6284 & 5.5631 & 0.7806 
    \\
    
\bottomrule[1.3pt]
\end{tabular}
\end{adjustbox}
\vspace{-0.6cm}
\end{table*}




The comparison of robustness among methods is shown in~\cref{tab:exp_rbs,tab:exp_weather_rbs}.

\noindent{\textbf{Observation 11}}: 
For \textit{nuScenes}~\cite{caesar2020nuscenes} and \textit{BridgeData}~\cite{walke2023bridgedata}, we observe a significant performance decline in most models as the frame interval increases (e.g., for TAU, the FVD is 370 when $\Delta t=1$, while its FVD is 503 when $\Delta t=2$ on \textit{BridgeData}).  However, MCVD~\cite{voleti2022mcvd} maintains a stable and superior performance with the best robustness. Interestingly, for \textit{BridgeData}, MCVD has better FVD metrics when evaluated on increased frame intervals than on the original interval. 
The same phenomenon is also observed on \textit{WeatherBench}~\cite{garg2022weatherbench}, where Earthformer~\cite{gao2022earthformer} has the best robustness, consistent with the short-term prediction.

\noindent{\textbf{Finding 11}}: Almost all models demonstrate performance decline on varied temporal frame intervals, especially on \textit{WeatherBench}. 
This paper first introduces the evaluation of temporal robustness for spatio-temporal prediction models, revealing that no single model consistently exhibits superior robustness. To enhance this capability, potential improvements could include integrating dynamic frame interval training strategies or implementing specific temporal modules.


\section{Conclusion}
\label{sec:conclusion}
In this work, we introduce PredBench, a comprehensive benchmark for evaluating spatio-temporal prediction networks. Encompassing a wide range of applications and disciplines, PredBench integrates 12 established STP methods and 15 diverse datasets to provide a thorough evaluation platform. By standardizing experimental settings, we ensure equitable comparisons across various STP networks, fostering a level field for analysis.
PredBench extends beyond conventional evaluation metrics to include multi-dimensional assessments that address both short-term and long-term predictive capabilities, as well as the generalization and temporal robustness of the models. 
Several findings gleaned from our extensive experiments yield valuable insights for the future of STP research. 


The empirical observations and methodological contributions of PredBench are intended to catalyze progress in the STP domain, inspiring new research directions and innovations. We anticipate that PredBench will serve as a valuable resource for researchers seeking to advance the state-of-the-art in spatio-temporal prediction.


\section*{Acknowledgements}
This work is supported by Shanghai Artificial Inteligence Laboratory.

%
%
\bibliographystyle{splncs04}
\bibliography{main}

\clearpage

\title{Appendices} 
\titlerunning{PredBench}
\author{}
\authorrunning{Z.~Wang et al.}
\institute{}
\maketitle

\appendix

\section*{Overview}
This supplementary document provides additional details to support our main paper, organized as follows:
\begin{itemize}
    \item ~\cref{sec:suppl_sepd} shows more details about the standard experimental protocols, which include the details of all datasets, the previous intricate experiment settings, and our meticulous calibration.
    \item ~\cref{sec:suppl_metric} summarizes the calculation approaches of all the metrics used in the main paper.
    \item ~\cref{sec:suppl_codebase} provides a comprehensive presentation of the reproduction results within our codebase for previous methods, alongside statistical measures such as standard deviation and significance tests.
    \item ~\cref{sec:suppl_implement} provides details of model size and configurations for training, including learning rate, batch size, optimizer, and so on.
    \item ~\cref{sec:suppl_qua_res} demonstrates more qualitative results of model performance on each dataset.
    \item ~\cref{sec:suppl_human} showcases the \textbf{human assessment} results on prediction quality to determine the indicator that best reflects model performance.
    \item ~\cref{sec:suppl_discuss} discusses the broader impacts and limitations of our PredBench.
\end{itemize}

\section{Standard Experimental Protocol Details}
\label{sec:suppl_sepd}

We meticulously calibrate the dataset setting and demonstrate the dataset statistics in section 3.3 of the main paper. We provide more detailed information for each dataset.



\noindent{\textbf{Motion Trajectory Prediction:}}
\begin{itemize}
    \item \textit{Moving-MNIST}~\cite{srivastava2015movingmnist} is one of the seminal datasets that has been widely utilized. This dataset contains handwritten digits sampled from the MNIST dataset, moving at a constant speed and bounded within a $64\times 64$ frame. By selecting random digits, placing each digit at random locations, and assigning random speed and direction, it is possible to generate infinite sequences. Following conventions on this dataset, we generate the training data on the fly and utilize $10K$ videos as the testing set. 
    \item \textit{KTH}~\cite{schuldt2004kth} contains 6 types of human actions, namely, walking, jogging, running, boxing, hand-waving, and hand-clapping, performed by 25 persons in 4 different scenes. Conventionally, validation is ignored in this dataset, i.e., persons 1-16 for training and persons 17-25 for testing. We utilize persons 1-14 as the training set and persons 15-16 as the validation set to fill the gap. Besides, there are some differences in the training settings between PredRNN~\cite{wang2017predrnn, wang2018predrnn++, wang2021predrnnv2} and SimVP~\cite{gao2022simvp, tan2022simvpv2}. PredRNN predicts the subsequent 10 frames during the training stage, while SimVP predicts the subsequent 20 or 40 frames. We follow the input-output setting of PredRNN for training in our experiments.
    \item \textit{Human3.6M}~\cite{ionescu2013human} represents general human actions with complex 3D articulated motions, including 3.6 million poses and corresponding images. This dataset contains diverse human actions across 15 types, i.e., discussion, eating, greeting, walking, and so on.  SimVPv1~\cite{gao2022simvp} utilizes 73,404 and 8,582 videos from \textit{Human3.6M} as the training set and test set without the validation set. We randomly select 66,063 videos from the past training set as our training set. The remaining 7,341 videos are our validation set. 
\end{itemize}

\noindent{\textbf{Robot Action Prediction:}}
\begin{itemize}
    \item \textit{RoboNet}~\cite{dasari2019robonet} is a large-scale dataset for robot action planning, including roughly 15 million video frames from 7 different robot platforms. We resize each image to $120\times 160$ due to the computational constraints. According to GHVAE~\cite{wu2021hier}, we utilize the same 256 videos as the testing set and use 2 frames as input, and predict the subsequent 10 frames during the training stage. However, the validation set was not adopted during their experiments. For experimental completeness, we split the remaining data into the training and validation set according to 9:1 splits. The input and output settings of the models trained on \textit{RoboNet} are consistent. 
    
    \item \textit{BAIR}~\cite{ebert2017bair} contains the action-conditioned videos collected by a Sawyer robotic arm pushing various objects. we follow the dataset setting of MCVD~\cite{voleti2022mcvd} and use the same 256 videos as the testing set. We split the remaining data into the training and validation set to solve the same problem of missing validation set like \textit{RoboNet}. However, there are significant differences in the training settings between PredRNNv2~\cite{wang2021predrnnv2} and MCVD. Specifically, MCVD uses 1 or 2 frames as input and predicts the subsequent 5 frames during the training stage, while PredRNN v2 uses 2 frames as input and predicts the subsequent 10 frames. To maintain consistency in training settings, we follow the input-output setting in \textit{RoboNet} with 2 frames as input and 10 frames as output.
    
    \item \textit{BridgeData}~\cite{walke2023bridgedata} is a large multi-domain and multi-task dataset, with more than 7 thousand demonstrations containing 71 tasks (e.g., close fridge) across 10 scenes (e.g., kitchen and tabletop). It is noteworthy that we first introduce this dataset into spatio-temporal prediction tasks. We divide this dataset into the training, validation, and testing sets according to 8:1:1 splits, where each image is resized to $120\times 160$. We utilize the same input-output setting in \textit{RoboNet} with 2 frames as input and 10 frames as output. Exactly, in Table 7 of the paper, $($\textit{new scene, new task}$)$ is $($\textit{sink, flip cup}$)$, $($\textit{new scene, original task}$)$ is $($\textit{sink, turn lever}$)$, and $($\textit{original scene, new task}$)$ is $($\textit{kitchen, lift bowl}$)$. 
\end{itemize}

\noindent{\textbf{Driving Scene Prediction:}}
\begin{itemize}
    \item \textit{CityScapes}~\cite{cordts2016cityscapes} is a large, diverse dataset containing stereo video sequences recorded in streets from 50 different cities. We adopt the same training, validation, and test sets as MCVD. However, MCVD directly evaluates the models in the test set without using the validation set. We choose the models for evaluation according to the performance obtained from the validation set. 
    \item \textit{KITTI}~\cite{geiger2013kitti} is a challenging real-world car-mounted camera video dataset with 5 diverse scenarios, i.e., city, residential, road, campus, and person. We discard the data of the person scenario, as it is characterized by human movement rather than driving scenes. For the data of the other four scenarios, we exclude the static videos (where frames have negligible change) and divide the remaining data into training, validation, and test sets in a 9:2:2 ratio, which differs from SimVP~\cite{gao2022simvp, tan2022simvpv2} and MAU~\cite{chang2021mau} which did not perform validation and test on \textit{KITTI}. We crop and resize each image to $128\times 160$ to fit the image size of \textit{Caltech}. We set the input and output of the model to 10 frames for training instead of only predicting 1 frame~\cite{gao2022simvp, tan2022simvpv2, chang2021mau} which can not present the full-scale performance of the model.
    \item \textit{Caltech}~\cite{dollar2009caltech}, initially proposed for pedestrian detection, has become a widely used benchmark dataset in spatio-temporal prediction. It is conventionally utilized as a testing dataset for models trained on \textit{KITTI} due to the scene similarity between these two datasets. 
    \item \textit{nuScenes}~\cite{caesar2020nuscenes} is a newly proposed driving scene dataset collected by 6 cameras, 5 radars, and 1 lidar mounted on the driving platform. We utilize the driving scene videos collected by the front camera, divide the data into training, validation, and test sets in an 8:1:1 ratio, and set the input and output of the model to 10 frames for training. Each image is cropped and resized to $128\times 160$ to fit the image size of \textit{Caltech}.
\end{itemize}

\noindent{\textbf{Traffic Flow Prediction:}}
\begin{itemize}
    \item \textit{TaxiBJ}~\cite{zhang2018taxibj} includes GPS data in Beijing containing inflow and outflow information in a 30-minute interval. 
    We randomly select 500 sequences from the training set in PhyDnet as a validation set. The remaining 19,961 sequences are our training set. Following PhyDNet~\cite{Guen2020phydnet}, we utilize 500 sequences as a test set and follow the input-output setting for training.
    \item \textit{Traffic4Cast2021}~\cite{Traffic4Cast2021} is an industrial-scale dataset capturing the traffic dynamics across 10 diverse cities in a period of 2 years. Each data contains 8 dynamic channels, encoding traffic volume and average speed per heading direction: NE, SE, SW, and NW. We center-crop the original $495\times 436$ image to $128\times 112$ due to the computational constraints. We set aside the data of Moscow city in \textit{Traffic4Cast2021} for generalization evaluation. We divide the remaining data into training, validation, and test sets in an 8:1:1 ratio. We follow the input-output setting of PredRNNv2~\cite{wang2021predrnnv2} for training in our experiments, where the model predicts 3 frames based on the 9 historical frames.
\end{itemize}

\noindent{\textbf{Weather Forecasting:}}
\begin{itemize}
    \item \textit{ICAR-ENSO}~\cite{ICARENSO} consists of historical climate observation and stimulation sea surface temperature (SST) data provided by the Institute for Climate and Application Research (ICAR). Each SST data covers the geographical region $(90^{\circ}E-330^{\circ}W, 55^{\circ}S-60^{\circ}N)$ of the Pacific Rim, with the spatial resolution of $5^{\circ}$ $(24\times 48)$ and the time interval of 1 month. It is worth noting that only the SST data across a certain area $(170^{\circ}W-120^{\circ}W, 5^{\circ}S-5^{\circ}N)$ is used to calculate $C^{Nino3.4}$. Following Earthformer~\cite{gao2022earthformer}, we use the same training, validation, and test sets for evaluation. We forecast the SST anomalies up to 14 steps given a context of 12 steps of SST anomalies observations.
    \item  \textit{SEVIR}~\cite{veillette2020sevir} a spatio-temporally aligned dataset containing over 10,000 weather events, spanning 4 hours in 5-minute steps. Images in \textit{SEVIR} are sampled and aligned to $384\times 384$ across 5 different types: three channels (C02, C09, C13) from the GOES-16 advanced baseline imager, NEXRAD Vertically Integrated Liquid (VIL) mosaics, and GOES-16 Geostationary Lightning Mapper (GLM) flashes. Following Earthformer~\cite{gao2022earthformer}, we use the same training, validation, and test sets and predict the future VIL up to 60 minutes (12 frames) given 65 minutes of context VIL (13 frames).
    \item \textit{WeatherBench}~\cite{garg2022weatherbench} is a large-scale dataset derived from ERA5 archive, which is down-sampled to $1.40625^{\circ}$ ($128\times 256$ grid points). This dataset provides a wide range of variables, including 6 surface variables and 8 atmospheric variables with 13 levels, a total of 110 $(6+8\times 13 =110)$ variables. Following the setup of previous works~\cite{lin2022conditionalforecast,garg2022weatherbench,Rasp2023weatherbench2,lam2022graphcast,bi2023pangu,chen2023fengwu} in meteorology, we use 4 surface variables, (t2m, u10, v10, tp) and 5 atmospheric variables (z, t, r, u, v), a total of 69 variables. Specifically, the atmospheric variables are geopotential (z), temperature (t), relative humidity(r), wind in longitude direction (u), and wind in latitude direction (v) at 13 levels (50, 100, 150, 200, 250, 300, 400, 500, 600, 700, 850, 925, 1000 hPa). The surface variables are 2-meter temperature (t2m), 10-meter u wind component (u10), 10-meter v wind component (v10), and total precipitation (tp). The model is trained on data from 1979-2015, validated on data from 2016, and tested on data from 2017-2018, with 2 frames as input and 1 frame as output. We present metrics on variables t2m, t850, and z500, following the conventions in meteorology.
\end{itemize}

\section{Detailed Evaluation Metrics}
\label{sec:suppl_metric}
The evaluation metrics used in our experiments are presented in main paper, we provide detailed calculations of each metric in this section.

\noindent \textbf{Error Metrics.}
We adopt Mean Absolute Error (MAE), Root Mean Squared Error (RMSE), and Weighted Mean Absolute Percentage Error to assess the pixel-level disparity between predicted and actual sequences.

Given the L-length prediction results from the T timestamp, $\hat{\mathcal{X}}^{T+1, T+L} = \{\boldsymbol{x}^{T+1},\cdots, \boldsymbol{x}^{T+L}\} \in \mathbb{R}^{L\times C \times H \times W}$ and the target $\mathcal{X}^{T+1, T+L}$, MAE, RMSE and WMAPE are defined as follows:

\vskip -0.05 in
\begin{equation}
    \begin{aligned}
        & MAE = \frac{1}{L\cdot C\cdot H\cdot W}\sum_{t=T+1}^{T+L}\sum_{c,h,w} \left\lvert {x}_{c,h,w}^t - \hat{x}_{c,h,w}^t \right\rvert , \\
        & RMSE = \frac{1}{L} \sum_{t=T+1}^{T+L} \sqrt{\frac{1}{C\cdot H\cdot W} \sum_{c,h,w}({x}_{c,h,w}^t - \hat{x}_{c,h,w}^t)^2} , \\
        & WMAPE = \frac{1}{L} \sum_{t=T+1}^{T+L} \frac{
            \sum\limits_{c,h,w} \left\lvert {x}_{c,h,w}^t - \hat{x}_{c,h,w}^t\right \rvert
        }{
            \sum\limits_{c,h,w} \left\lvert {x}_{c,h,w}^t \right \rvert
        } ,
    \end{aligned}
\end{equation}
where $C$, $H$, and $W$ represent the channel, height, and width of a single frame, as well as $t$, $c$, $h$, and $w$ denote the index for time, channel, height, and width.

\noindent \textbf{Similarity Metrics.}
We use Structural Similarity Index Measure (SSIM)~\cite{wang2004ssim} and Peak Signal-to-Noise Ratio (PSNR) to assess the image quality. Using the same notations, SSIM and PSNR are computed as follows:

\vskip -0.05 in
\begin{equation}
    \begin{aligned}
        & SSIM = \frac{1}{L} \sum_{t=T+1}^{T+L}\cdot \frac{(2\mu_{x^t}\mu_{\hat{x}^t})(2\sigma_{x^t\hat{x}^t}+c_2)}{(\mu_{x^t}^2 + \mu_{\hat{x}^t}^2 + c_1)(\sigma_{x^t}^2 + \sigma_{\hat{x}^t}^2 + c_2)}, \\
        & PSNR = \frac{1}{L} \sum_{t=T+1}^{T+L} 20\cdot\log_{10}\left( \dfrac{\max\limits_{c,h,w}x_{c,h,w}^t}{RMSE(x^t)}\right), \\
    \end{aligned}
\end{equation}
where $\mu_{x}$ and $\sigma_{x}$ denote the pixel sample mean and variance of a single frame $x$, $\sigma_{xy}$ is the covariance of two frames $x$ and $y$, $c_1$ and $c_2$ are two variables to stabilize the division with weaker denominator, and $RMSE(x)$ means the root mean squared error of a single frame $x$.

\noindent \textbf{Perception Metrics.}
Learned Perceptual Image Patch Similarity (LPIPS)~\cite{zhang2018lpips} and Fréchet Video Distance (FVD)~\cite{unterthiner2018fvd} are employed to assess perceptual similarity in line with the human visual system. We follow the official implementation~\footnote{\url{https://github.com/richzhang/PerceptualSimilarity/tree/master/lpips}} and use the extracted features to compute LPIPS. For FVD, we follow the official implementation~\footnote{\url{https://github.com/google-research/google-research/blob/master/frechet_video_distance}} and convert the official I3D~\cite{carreira2017i3d} model trained on \textit{Kinetics-400}~\cite{kay2017kinetics} to PyTorch to extract video features.

\noindent \textbf{Weather Metrics.}
Weighted Root Mean Squared Error (WRMSE) and Anomaly Correlation Coefficient (ACC) are used for \textit{WeatherBench}~\cite{garg2022weatherbench}, Critical Success Index (CSI) is applied to \textit{SEVIR}, while the three-month-moving-averaged Nino3.4 index ($C^{Nino3.4}$) is selected for \textit{ICAR-ENSO}~\cite{ICARENSO}. 

\noindent{\textit{WRMSE and ACC:}} WRMSE and ACC are computed for every \textbf{single variable} (i.e., single channel). Let $c$ denote the index of the channel for a specific variable, the WRMSE is defined as follows:

\vskip -0.05 in
\begin{equation}
    WRMSE = \frac{1}{L} \sum_{t=T+1}^{T+L} \sqrt{\frac{1}{H\cdot W} \sum_{h,w}\alpha_w \cdot ({x}_{c,h,w}^t - \hat{x}_{c,h,w}^t)^2} ,
\end{equation}
where $w$ and $h$ represent the indices for each grid along the latitude and longitude indices, $\alpha_w$ is the weight coefficient for each latitude index $w$. Denote $\phi_{w,h}$ as the latitude of point $(w,h)$, the weight coefficient $\alpha_w$ is defined as:

\vskip -0.05 in
\begin{equation}
    \alpha_w = W \cdot \frac{\cos(\phi_{w,h})}{\sum\limits_{w'=1}\cos(\alpha_{w',h})} .
\end{equation}

Given $C_{c,h,w}^t$ as the climatological mean over the day-of-year containing the validity time $t$ for a given weather variable $c$ at point $(w,h)$. The ACC is defined as:

\vskip -0.05 in
\begin{equation}
    \begin{aligned}
        & \mathtt{y}_{c,h,w}^{t} = x_{c,h,w}^t-C_{c,h,w}^t, \\
        & \hat{\mathtt{y}}_{c,h,w}^{t} = \hat{x}_{c,h,w}^t-C_{c,h,w}^t, \\
        & ACC = \frac{1}{L} \sum_{t=T+1}^{T+L} \dfrac{
            \sum\limits_{w,h} \alpha_w\cdot \mathtt{y}_{c,h,w}^{t}\cdot \hat{\mathtt{y}}_{c,h,w}^{t}
        }{
            \sum\limits_{w,h}\alpha_w (\mathtt{y}_{c,h,w}^{t})^2 \cdot \sum\limits_{w,h}\alpha_w(\hat{\mathtt{y}}_{c,h,w}^{t})^2
        }. \\
    \end{aligned}
\end{equation}

\noindent{\textit{CSI:}} Following \textit{SEVIR}~\cite{veillette2020sevir}, the predicted and target sequences are scaled to the range $0-255$ and binarized at thresholds $[16, 74, 133, 160, 181, 219]$ to calculate CSI. As shown in \cref{tab:sc_t}, the $\mathtt{Hit}$, $\mathtt{Mis}$, $\mathtt{Fas}$ and $\mathtt{Cr}$ at threshold $\tau$ are defined by:

\vskip -0.05 in
\begin{equation}
    \begin{aligned}
        & \mathtt{Hit}_{\tau} = \sum_{t,c,h,w} (x_{c,h,w}^{t} \geqslant \tau) \land (\hat{x}_{c,h,w}^t \geqslant \tau), \\
        & \mathtt{Mis}_{\tau} = \sum_{t,c,h,w} (x_{c,h,w}^{t} \geqslant \tau) \land (\hat{x}_{c,h,w}^t < \tau), \\
        & \mathtt{Fas}_{\tau} = \sum_{t,c,h,w} (x_{c,h,w}^{t} < \tau) \land (\hat{x}_{c,h,w}^t \geqslant \tau), \\
        & \mathtt{Cr}_{\tau} = \sum_{t,c,h,w} (x_{c,h,w}^{t} < \tau) \land (\hat{x}_{c,h,w}^t < \tau), 
    \end{aligned}
\end{equation}
where $\land$ represents the logical AND operation, as well as $t$, $c$, $h$, and $w$ denote the index for time, channel, height, and width.

\begin{table}
  \centering
  \caption{Schematic contingency table for the CSI metric. The prediction and ground-truth are binarized for calculation.}
  \label{tab:sc_t}
  \vspace{-0.3cm}
  \begin{tabular}{|c|c|c|}
    \hline
     & Observed & Not Observed \\ \hline
    Predicted & $\mathtt{Hit}$ (Hits) & $\mathtt{Fas}$ (False Alarms) \\ \hline
    Not Predicted & $\mathtt{Mis}$ (Misses) & $\mathtt{Cr}$ (Correct Rejections) \\ \hline
  \end{tabular}
  
\end{table}

We report CSI as the mean of CSI at the aforementioned six thresholds $\mathcal{T} = \{ 16,74,133,160,181,219 \}$, the formulation is as follows:

\vskip -0.05 in
\begin{equation}
    CSI = \frac{1}{6} \sum_{\tau \in \mathcal{T}} \frac{\mathtt{Hit}_{\tau}}{\mathtt{Hit}_{\tau} + \mathtt{Fas}_{\tau} + \mathtt{Mis}_{\tau}}.
\end{equation}

\noindent{\textbf{$C^{Nino3.4}$:}} The Nino3.4 index is computed by averaging the sea surface temperature anomalies over the area bounded by $(170^{\circ}W-120^{\circ}W, 5^{\circ}S-5^{\circ}N)$, serving as an indicator of the ENSO (El Ni\~{n}o–Southern Oscillation) conditions. Specifically, the Nino3.4 index is calculated through the three-month average:


\vskip -0.05 in
\begin{equation}
    y^t = \frac{1}{3} \sum_{i\in \{0, 1,2 \}} \frac{1}{C\cdot H\cdot W} \sum_{c,h,w} x_{c,h,w}^{t+i},
\end{equation}
where $C=1, H=3, W=11$ in \textit{ICAR-ENSO} dataset, as the data is represented as a grid with a spatial resolution of $5^{\circ}$ and a temporal interval of one month.

$C^{Nino3.4}$ is the correlation coefficient of Nino3.4 index. Given the L-length prediction results from the T timestamp, $\hat{\mathcal{X}}^{T+1, T+L} \in \mathbb{R}^{L\times C \times H \times W}$ and the target $\mathcal{X}^{T+1, T+L}$, they are firstly cropped to the aforementioned region, yielding $\hat{\mathcal{X}}^{T+1, T+L}, \mathcal{X}^{T+1, T+L} \in \mathbb{R}^{L\times 1 \times 3 \times 11}$. Through the three-month average, we get the Nino3.4 index for the prediction and the target, denoted respectively as $\hat{\mathcal{Y}}^{T+1, T+L-2}, \mathcal{Y}^{T+1, T+L-2} \in \mathbb{R}^{L-2}$. The $C^{Nino3.4}$ is defined as:

\vskip -0.05 in
\begin{equation}
    \begin{aligned}
        u^{t} = y^{t} - \frac{1}{L-2} \sum\limits_{i=T+1}^{T+L-2}{y^{t}}, \\
        \hat{u}^{t} = \hat{y}^{t} - \frac{1}{L-2} \sum\limits_{i=T+1}^{T+L-2}{\hat{y}^{t+i}}, \\
        C^{Nino3.4} = \frac{\sum\limits_{i=T+1}^{T+L-2} u^{t} \cdot \hat{u}^{t}}{\sqrt{\sum\limits_{i=T+1}^{T+L-2} (u^{t})^2 \cdot (\hat{u}^{t})^2}}
    \end{aligned}
\end{equation}

\section{Codebase Analysis}
\label{sec:suppl_codebase}

\subsection{Unified Codebase}

\begin{figure}[tb]
  \centering
  \includegraphics[width=1.0\textwidth]{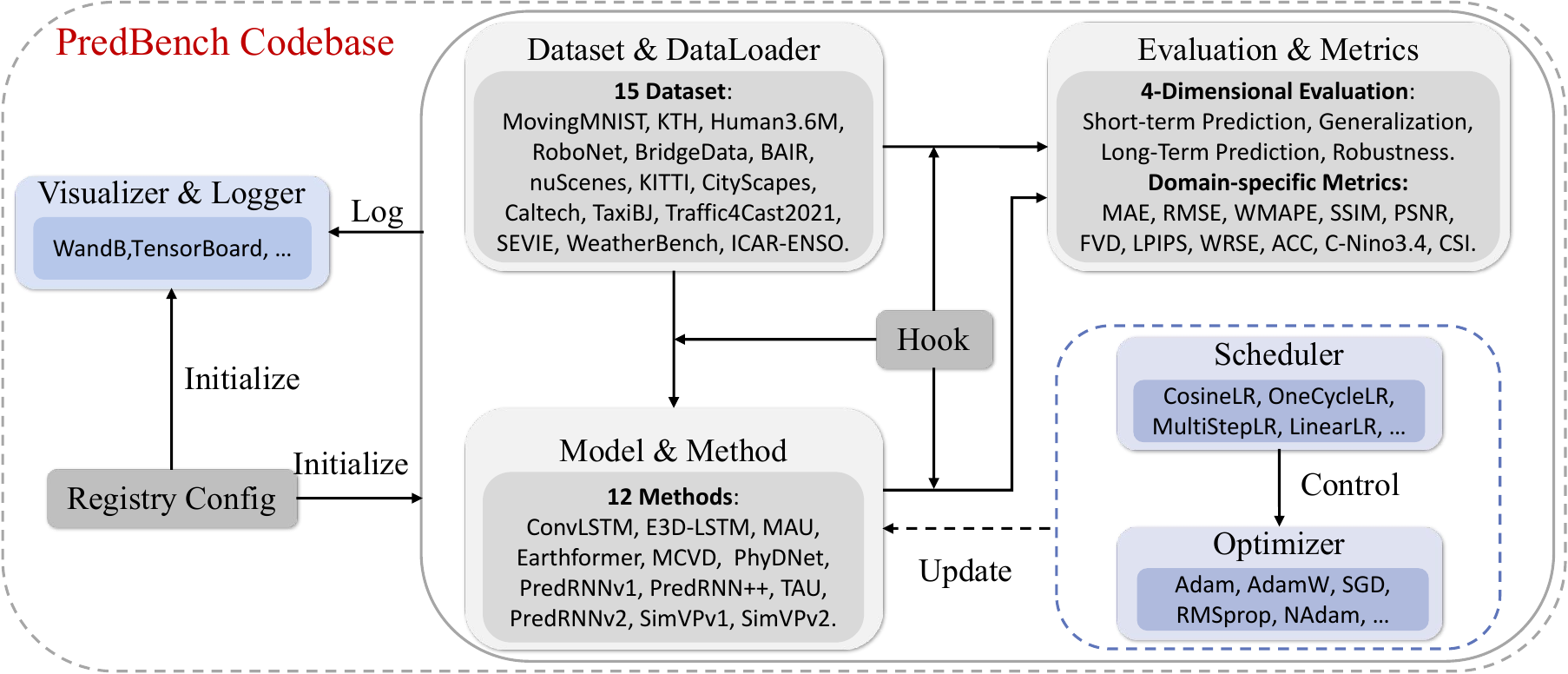}
  \vspace{-0.5cm}
  \caption{
      Overview of our \textbf{unified codebase}. 
  }
  \label{fig:codebase}
\end{figure}

We build a uniform codebase using MMEngine~\cite{mmengine2022}.
To ensure reproducibility and coherence, we utilize the codes of each model available on GitHub and make minimal modifications to fit our codebase. 
As shown in ~\cref{fig:codebase}, our codebase supports modular datasets and models, flexible configuration systems (Config and Hook), and rich analysis tools, resulting in a user-friendly system. It allows easy incoroporation of user-defined modules into any system component.

\begin{table*}[t]
\renewcommand\arraystretch{1.4}
\setlength\tabcolsep{1pt}
\center
\vskip 0.15in
\caption{\textbf{Reproduction results} on \textit{Moving-MNIST}~\cite{srivastava2015movingmnist}. The training data is generated dynamically. The MSE and MAE metrics are calculated in the normalized space (within the range of $[0,1]$).}
\label{tab:suppl_repr_mnist}
\vspace{-0.3cm}
\begin{adjustbox}{max width=.7\width}
\begin{tabular}{l|cc|cc|cc|cc}
\toprule[1.3pt]
  \multirow{2}*{Method}
  &\multicolumn{2}{c|}{\textbf{MSE$\downarrow$} }
  &\multicolumn{2}{c|}{\textbf{MAE$\downarrow$} }
  &\multicolumn{2}{c|}{\textbf{SSIM$\uparrow$}}
  &\multicolumn{2}{c}{\textbf{PSNR$\uparrow$}}
  \\
  & \textbf{original} & \textbf{ours} 
  & \textbf{original} & \textbf{ours} 
  & \textbf{original} & \textbf{ours} 
  & \textbf{original} & \textbf{ours} 
  \\
  
\Xhline{0.5pt}
    ConvLSTM
    & 29.80 & 29.63
    & 90.64 & 90.90
    & 0.9288 & 0.9290 
    & 22.10 & 22.12
    \\
    
    E3D-LSTM
    & 35.97 & 28.46
    & 78.28 & 68.54
    & 0.9320 & 0.9458
    & 21.11 & 22.60
    \\

    MAU 
    & 26.86 & 26.80
    & 78.22 & 78.20
    & 0.9398 & 0.9397
    & 22.57 & 22.76
    \\
    
    PhyDNet
    & 28.19 & 28.17
    & 78.64 & 69.17
    & 0.9374 & 0.9444
    & 22.62 & 23.18
    \\

    PredRNNv1 
    & 23.97 & 24.39
    & 72.82 & 73.61
    & 0.9462 & 0.9452
    & 23.28 & 23.18
    \\

    PredRNN++ 
    & 22.06 & 22.21
    & 69.58 & 69.93
    & 0.9509 & 0.9504
    & 23.65 & 23.62
    \\

    PredRNNv2 
    & 24.13 & 24.77
    & 73.73 & 75.48
    & 0.9453 & 0.9425
    & 23.21 & 23.19
    \\

    SimVPv1
    & 32.15 & 32.23 
    & 89.05 & 89.37
    & 0.9268 & 0.9268
    & 21.84 & 21.83
    \\

    SimVPv2 
    & 26.69 & 26.65
    & 77.19 & 76.97
    & 0.9402 & 0.9404
    & 22.78 & 22.78
    \\

    TAU 
    & 24.60 & 25.00
    & 71.93 & 73.73
    & 0.9454 & 0.9443
    & 23.19 & 23.11
    \\

    Earthformer 
    & 82.87 & 73.92
    & 23.99 & 23.93
    & 0.9445 & 0.9429
    & 23.09 & 23.24
    \\

    MCVD
    & 164.89 & 164.60
    & 64.12 & 64.21
    & 0.6290 & 0.6312
    & 19.12 & 19.12
    \\
    
\bottomrule[1.3pt]
\end{tabular}
\end{adjustbox}
\vskip -0.2in
\end{table*}

\begin{table*}[t]
\renewcommand\arraystretch{1.4}
\setlength\tabcolsep{1pt}
\center
\vskip 0.15in
\caption{\textbf{Reproduction results} on \textit{Human3.6M}~\cite{ionescu2013human}. It is worth noting that the validation dataset is not adopted in the reproduction experiments. The MSE and MAE metrics are calculated in the normalized space (within the range of $[0,1]$).}
\label{tab:suppl_repr_human}
\vspace{-0.3cm}
\begin{adjustbox}{max width=.8\columnwidth}
\begin{tabular}{l|cc|cc|cc|cc|cc}
\toprule[1.3pt]
  \multirow{2}*{Method}
  &\multicolumn{2}{c|}{\textbf{MSE$\downarrow$} }
  &\multicolumn{2}{c|}{\textbf{MAE$\downarrow$} }
  &\multicolumn{2}{c|}{\textbf{SSIM$\uparrow$}}
  &\multicolumn{2}{c|}{\textbf{PSNR$\uparrow$}}
  &\multicolumn{2}{c}{\textbf{LPIPS$\downarrow$}}
  \\
  & \textbf{original} & \textbf{ours} 
  & \textbf{original} & \textbf{ours} 
  & \textbf{original} & \textbf{ours} 
  & \textbf{original} & \textbf{ours} 
  & \textbf{original} & \textbf{ours} 
  
  \\
  
\Xhline{0.5pt}
    ConvLSTM 
    & 125.5 & 125.2
    & 1566.7 & 1541.7
    & 0.9813 & 0.9814
    & 33.40 & 33.43
    & 0.0356 & 0.0404
    \\
    
    E3D-LSTM 
    & 143.3 & 137.0
    & 1442.5 & 1589.7
    & 0.9803 & 0.9791 
    & 32.52 & 32.65
    & 0.0413 & 0.0310
    \\

    MAU 
    & 127.3 & 123.7
    & 1577.0 & 1548.4
    & 0.9812 & 0.9819 
    & 33.33 & 33.49
    & 0.0356 & 0.0385
    \\
    
    PhyDNet 
    & 125.7 & 142.8
    & 1614.7 & 1616.0
    & 0.9804 & 0.9807 
    & 33.05 & 33.09
    & 0.0371 & 0.03065
    \\

    PredRNNv1 
    & 113.2 & 113.9
    & 1458.3 & 1497.0
    & 0.9831 & 0.9825 
    & 33.94 & 33.84
    & 0.0325 & 0.0405
    \\

    PredRNN++ 
    & 110.0 & 109.15
    & 1452.2 & 1428.7
    & 0.9832 & 0.9835 
    & 34.02 & 34.06
    & 0.0320 & 0.0354
    \\

    PredRNNv2 
    & 114.9 & 117.7
    & 1484.7 & 1524.5
    & 0.9827 & 0.9818
    & 33.84 & 33.68
    & 0.0333 & 0.0268
    \\

    SimVPv1 
    & 115.8 & 122.9
    & 1511.5 & 1469.0
    & 0.9822 & 0.9826 
    & 33.73 & 33.64
    & 0.0347 & 0.0224
    \\

    SimVPv2 
    & 108.4 & 109.4
    & 1441.0 & 1430.9
    & 0.9834 & 0.9835
    & 34.08 & 34.08
    & 0.0322 & 0.0223
    \\

    TAU 
    & 113.3 & 113.3
    & 1390.7 & 1400.0
    & 0.9839 & 0.9839
    & 34.03 & 34.02
    & 0.0278 & 0.0198
    \\


    
\bottomrule[1.3pt]
\end{tabular}
\end{adjustbox}
\vskip -0.2in
\end{table*}

\begin{table*}[t]
\renewcommand\arraystretch{1.4}
\setlength\tabcolsep{1pt}
\center
\vskip 0.15in
\caption{\textbf{Reproduction results} of Earthformer~\cite{gao2022earthformer} on \textit{ICAR-ENSO}~\cite{ICARENSO}. $C^{Nino3.4}-W$ is the weighted $C^{Nino3.4}$ that evaluate the correlation skill of the Nino3.4 index.}
\vspace{-0.3cm}
\begin{adjustbox}{max width=.8\columnwidth}
\begin{tabular}{l|cc|cc|cc|cc|cc}
\toprule[1.3pt]
  \multirow{2}*{Method}
  &\multicolumn{2}{c|}{\textbf{MSE$\downarrow$} }
  &\multicolumn{2}{c|}{\textbf{MAE$\downarrow$} }
  &\multicolumn{2}{c|}{\textbf{$C^{Nino3.4}\uparrow$}}
  &\multicolumn{2}{c|}{\textbf{$C^{Nino3.4}-W\uparrow$}}
  &\multicolumn{2}{c}{\textbf{RMSE(Nino)$\downarrow$}}
  \\
  & \textbf{original} & \textbf{ours} 
  & \textbf{original} & \textbf{ours} 
  & \textbf{original} & \textbf{ours} 
  & \textbf{original} & \textbf{ours} 
  & \textbf{original} & \textbf{ours} 
  
  \\
  
\Xhline{0.5pt}
    
    Earthformer 
    & 0.2984 & 0.3140
    & 12.77 & 13.48
    & 0.6930 & 0.7020
    & 2.0750 & 2.1370
    & 0.6013 & 0.5384
    \\
\bottomrule[1.3pt]
\end{tabular}
\end{adjustbox}
\label{tab:suppl_repr_enso}
\vskip -0.2in
\end{table*}

\begin{table*}[t]
\renewcommand\arraystretch{1.4}
\setlength\tabcolsep{1pt}
\center
\vskip 0.15in
\caption{\textbf{Reproduction results} of Earthformer~\cite{gao2022earthformer} on \textit{SEVIR}~\cite{veillette2020sevir}. We demonstrate the results of the CSI metric at each threshold and their average.}
\label{tab:suppl_repr_sevir}
\vspace{-0.3cm}
\begin{adjustbox}{max width=1.\columnwidth}
\begin{tabular}{l|cc|cc|cc|cc|cc|cc|cc|cc|cc}
\toprule[1.3pt]
  \multirow{2}*{Method}
  &\multicolumn{2}{c|}{\textbf{MSE$\downarrow$} }
  &\multicolumn{2}{c|}{\textbf{MAE$\downarrow$} }
  &\multicolumn{2}{c|}{\textbf{CSI-16$\uparrow$}}
  &\multicolumn{2}{c|}{\textbf{CSI-74$\uparrow$}}
  &\multicolumn{2}{c|}{\textbf{CSI-133$\uparrow$}}
  &\multicolumn{2}{c|}{\textbf{CSI-160$\uparrow$}}
  &\multicolumn{2}{c|}{\textbf{CSI-181$\uparrow$}}
  &\multicolumn{2}{c|}{\textbf{CSI-219$\uparrow$}}
  &\multicolumn{2}{c}{\textbf{CSI-M$\uparrow$}}
  \\
  & \textbf{original} & \textbf{ours} 
  & \textbf{original} & \textbf{ours} 
  & \textbf{original} & \textbf{ours} 
  & \textbf{original} & \textbf{ours} 
  & \textbf{original} & \textbf{ours} 
  & \textbf{original} & \textbf{ours} 
  & \textbf{original} & \textbf{ours} 
  & \textbf{original} & \textbf{ours} 
  & \textbf{original} & \textbf{ours} 

  \\
  
\Xhline{0.5pt}
    
    Earthformer 
    & 234.09 & 229.57
    & 1671.2 & 1711.6
    & 0.7634 & 0.7528 
    & 0.6836 & 0.6891
    & 0.4177 & 0.4287
    & 0.3098 & 0.3209 
    & 0.2697 & 0.2791 
    & 0.1638 & 0.1640 
    & 0.4346 & 0.4391
    \\
\bottomrule[1.3pt]
\end{tabular}
\end{adjustbox}
\vskip -0.2in
\end{table*}

\begin{table*}[!htbp]
\renewcommand\arraystretch{1.4}
\setlength\tabcolsep{1pt}
\center
\vskip 0.15in
\caption{Ten rounds of experiments of PredRNN++ on \textit{TaxiBJ}.} 
\label{tab:std_taxibj}
\vspace{-0.3cm}
\begin{adjustbox}{max width=.8\columnwidth}
\begin{tabular}{l|cccccccccc|cc}
\toprule[1.3pt]
   & 1 & 2 & 3 & 4 & 5 & 6 & 7 & 8 & 9 & 10 & std$\downarrow$ & p$\_$value $\uparrow$ \\
  
  \Xhline{0.5pt}

  MAE & 10.0 & 9.99 & 9.93 & 10.07 & 9.74 & 9.93 & 9.99 & 9.95 & 9.95 & 9.94 & 0.0807 & 0.9188 \\
  RMSE & 15.29 & 15.26 & 15.32 & 15.46 & 15.01 & 15.23 & 15.27 & 15.23 & 15.23 & 15.31 & 0.1059 & 0.8561 \\
  WMAPE & 0.1247 & 0.1244 & 0.1236 & 0.1256 & 0.1211 & 0.1236 & 0.1243 & 0.1239 & 0.1239 & 0.1241 &  0.0011 & 0.9202 \\
  SSIM & 0.979 & 0.979 & 0.98 & 0.98 & 0.981 & 0.98 & 0.979 & 0.979 & 0.979 & 0.98 & 0.0007 & 1.0 \\
  PSNR & 39.08 & 39.11 & 39.08 & 38.97 & 39.18 & 39.11 & 39.09 & 39.11 & 39.12 & 38.99 & 0.059 & 1.0\\
  
\bottomrule[1.3pt]
\end{tabular}
\end{adjustbox}
\vspace{-0.5cm}
\end{table*}

\begin{table*}[!htbp]
\renewcommand\arraystretch{1.4}
\setlength\tabcolsep{1pt}
\center
\vskip 0.15in
\caption{Ten rounds of experiments of PredRNN++ on \textit{Moving-MNIST}. The MSE and MAE metrics are calculated in the normalized space (within the range of $[0,1]$).} 
\label{tab:std_mnist}
\vspace{-0.3cm}
\begin{adjustbox}{max width=.8\columnwidth}
\begin{tabular}{l|cccccccccc|cc}
\toprule[1.3pt]
   & 1 & 2 & 3 & 4 & 5 & 6 & 7 & 8 & 9 & 10 & std$\downarrow$ & p$\_$value $\uparrow$ \\
  
  \Xhline{0.5pt}
    
  MAE & 69.93 & 69.78 & 69.8 & 69.77 & 69.77 & 69.91 & 69.78 & 69.76 & 69.84 & 69.95 & 0.0699 & 0.8444 \\
  MSE & 22.21 & 22.2 & 22.16 & 22.11 & 22.18 & 22.22 & 22.2 & 22.15 & 22.17 & 22.28 & 0.0435 & 0.8007 \\
  SSIM & 0.95 & 0.951 & 0.951 & 0.951 & 0.951 & 0.95 & 0.951 & 0.951 & 0.95 & 0.95 & 0.0005 & 1.0 \\
  PSNR & 23.62 & 23.63 & 23.62 & 23.63 & 23.63 & 23.62 & 23.63 & 23.63 & 23.63 & 23.61 & 0.0067 & 0.6811 \\
  LPIPS & 0.0472 & 0.047 & 0.0472 & 0.0471 & 0.047 & 0.047 & 0.047 & 0.0471 & 0.0472 & 0.0471 & 8.3e-5 & 0.7404 \\
  FVD & 27.6 & 27.2 & 27.6 & 27.5 & 27.2 & 27.2 & 27.2 & 27.4 & 27.6 & 27.4 & 0.1700 & 0.6258 \\
  
\bottomrule[1.3pt]
\end{tabular}
\end{adjustbox}
\vspace{-0.3cm}
\end{table*}

\subsection{Reproduction results}

To ensure reproducibility, we conducted a comparative analysis between the performance of our model executed within our codebase and the model executed using the official code released by the authors. Both sets of experiments are executed under identical settings to ensure a fair and consistent evaluation.

The reproduction results are shown in Tabs. \ref{tab:suppl_repr_mnist}, \ref{tab:suppl_repr_human}, \ref{tab:suppl_repr_enso}, and \ref{tab:suppl_repr_sevir}. Comparing the results of the two implementations verifies the fidelity of our codebase and its ability to replicate the intended model faithfully. This meticulous comparison process helps guarantee the trustworthiness of our further findings.

\subsection{Codebase Reliability}

We found that PredRNN++~\cite{wang2018predrnn++} can serve as a good baseline (Finding 1 in section 4.1 of the main paper), so we use it to run 10 rounds of experiments on the \textit{TaxiBJ}~\cite{zhang2018taxibj}  and \textit{Moving-MNIST}~\cite{srivastava2015movingmnist} dataset and calculate the metrics separately. We calculate the standard deviations of the then metric values and divide them equally into two groups for calculating the p-values of the T-test.
The metrics, standard deviations, and the p-values of the T-test are shown in ~\cref{tab:std_taxibj,tab:std_mnist}. 

It is obvious that the standard deviations are close to $0$ and the p-values are close to $1$, underscoring the reliability of our codebase.

\section{Implementation Details}
\label{sec:suppl_implement}

We provide the detailed computational analysis for each model in \cref{tab:model_size}, where FLOPs is calculated with $H=W=64$, $C=1$, and $T_{\text{in}}=T_{\text{out}}=10$. CNN models (e.g., SimVP, TAU) have higher FPS, making them suitable for real-time applications. Despite low FLOPs, RNN models (e.g., PredRNN) are slower due to auto-regressive generation. Transformer models (e.g., Earthformer) are computationally intensive with $O(n^2)$ complexity. Diffusion models (e.g., MCVD) achieve high FPS but require iterative sampling (we use 250 steps), which must be considered for real-time applications.

Detailed information about the hyperparameters of the experiments for each method in PredBench is shown in Tabs. \ref{tab:suppl_imp_convlstm}, \ref{tab:suppl_imp_e3dlstm}, \ref{tab:suppl_imp_mau}, \ref{tab:suppl_imp_phydnet}, \ref{tab:suppl_imp_predrnnpp}, \ref{tab:suppl_imp_predrnnv1}, \ref{tab:suppl_imp_predrnnv2}, \ref{tab:suppl_imp_simvpv1}, \ref{tab:suppl_imp_simvpv2}, \ref{tab:suppl_imp_tau}, \ref{tab:suppl_imp_earthformer}, and \ref{tab:suppl_impm_mcvd}.

\section{Qualitative Results}

\label{sec:suppl_qua_res}
\begin{figure*}[t]
    \centering
    \includegraphics[width=0.9\linewidth]{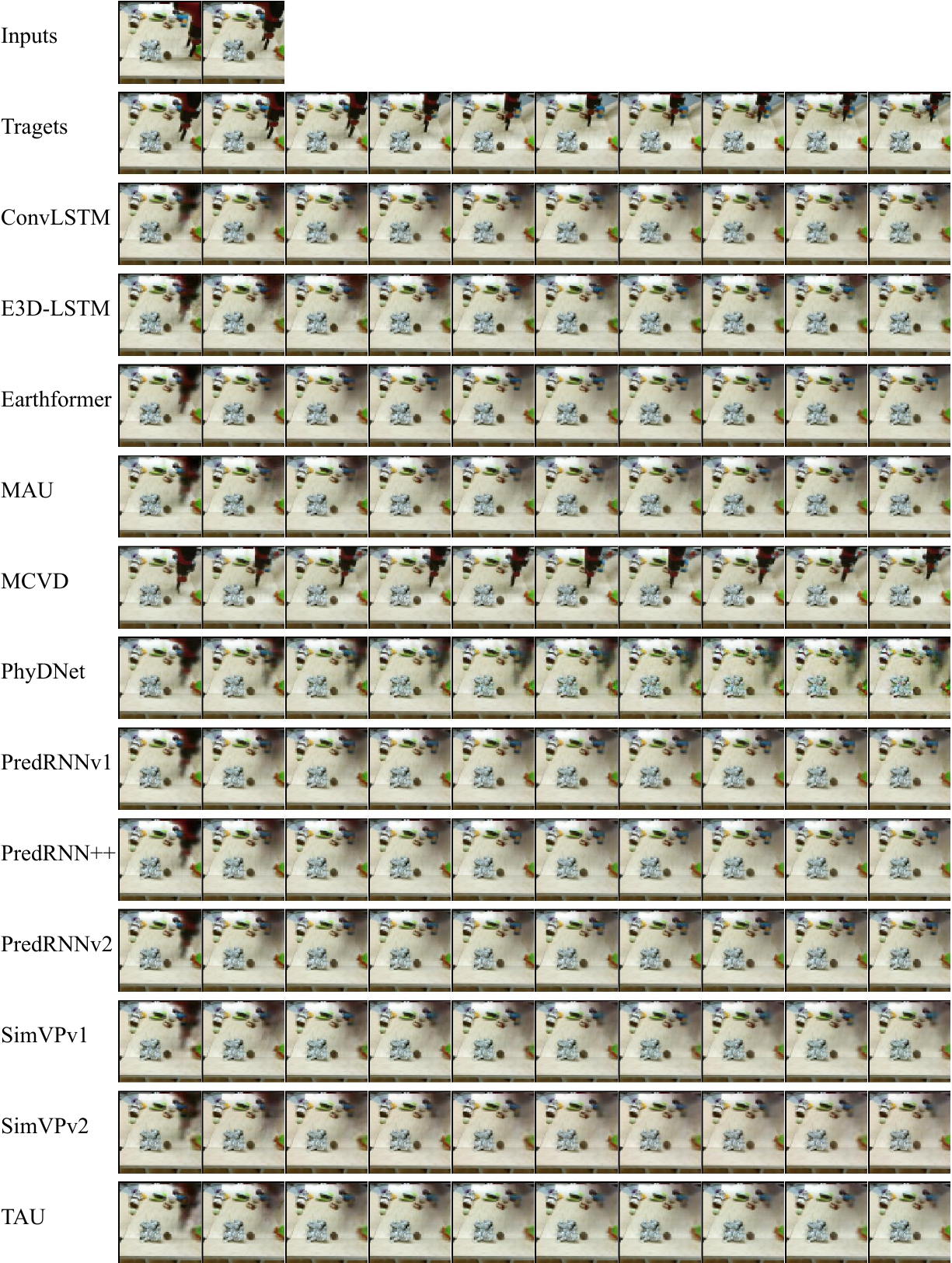}
    \caption{
        \textbf{Qualitative results} on \textit{BAIR}~\cite{ebert2017bair} (2 frames $\longrightarrow$ 10 frames).
    }
    \vspace{-0.3cm}
    \label{fig:qualitative_bair}
\end{figure*}

\begin{figure*}[t]
    \centering
    \includegraphics[width=1.0\linewidth]{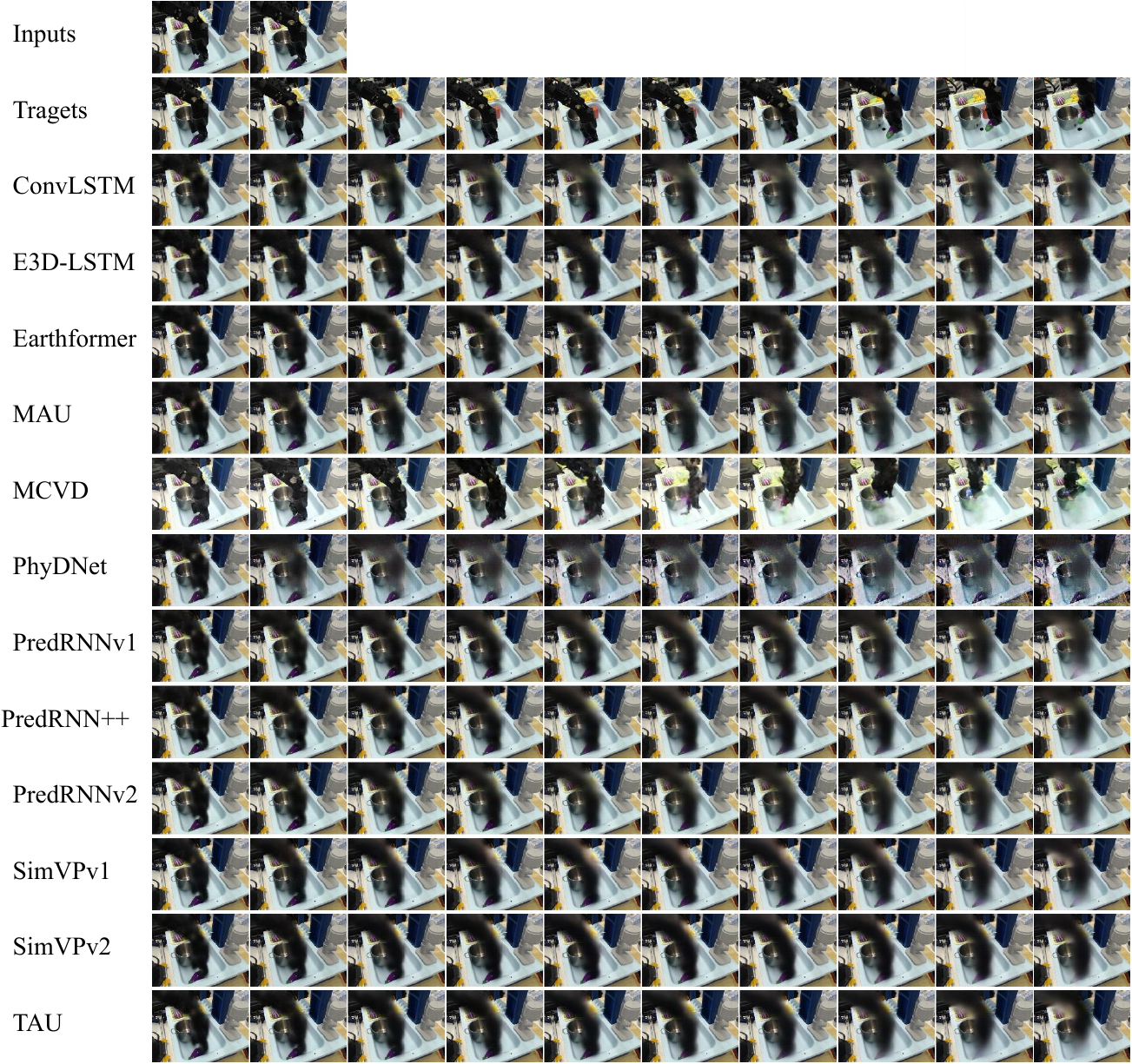}
    \caption{
        \textbf{Qualitative results} on \textit{BridgeData}~\cite{walke2023bridgedata} (2 frames $\longrightarrow$ 10 frames).
    }
    \vspace{-0.3cm}
    \label{fig:qualitative_bridgedata} 
\end{figure*}

\begin{figure*}[t]
    \centering
    \includegraphics[width=0.5\linewidth]{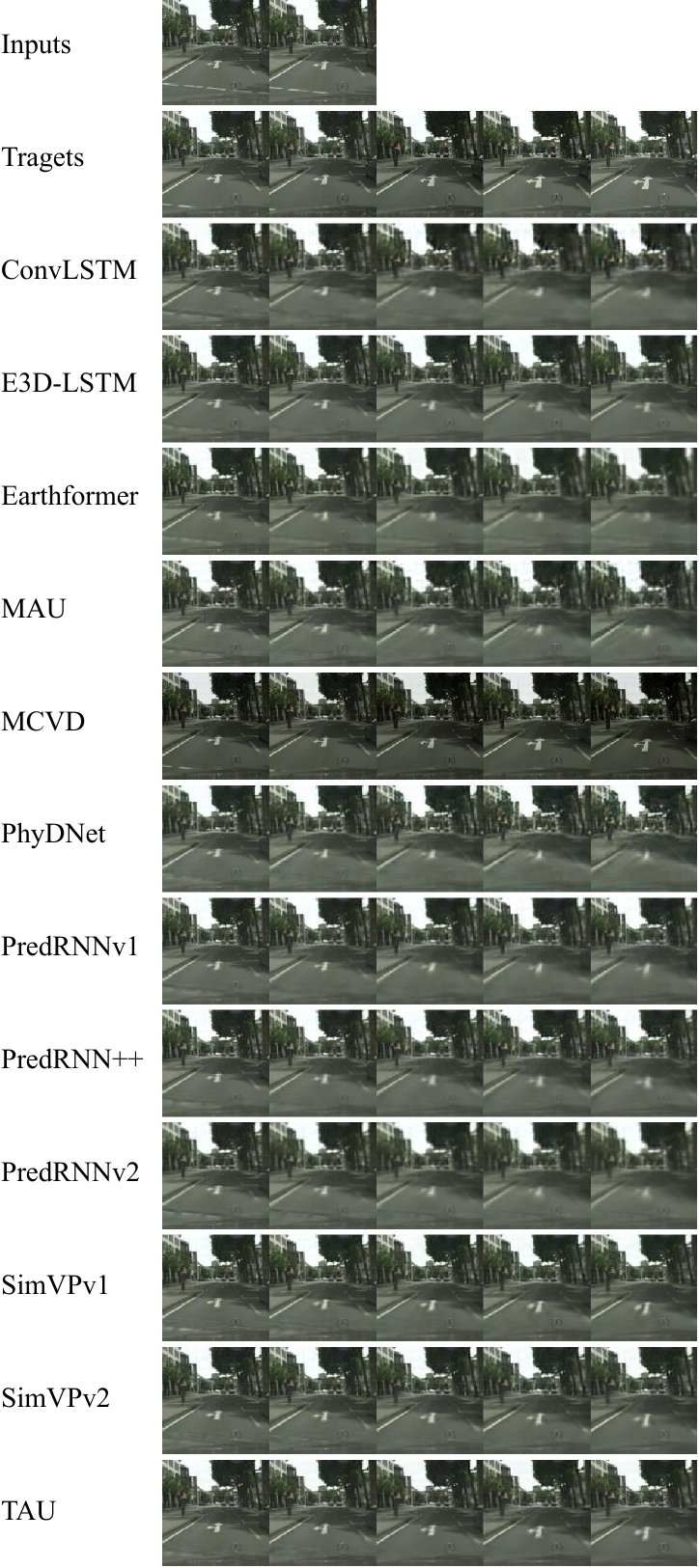}
    \caption{
        \textbf{Qualitative results} on \textit{CityScapes}~\cite{cordts2016cityscapes} (2 frames $\longrightarrow$ 5 frames).
    }
    \vspace{-0.3cm}
    \label{fig:qualitative_cityscapes}
\end{figure*}

\begin{figure*}[t]
    \centering
    \includegraphics[width=1.0\linewidth]{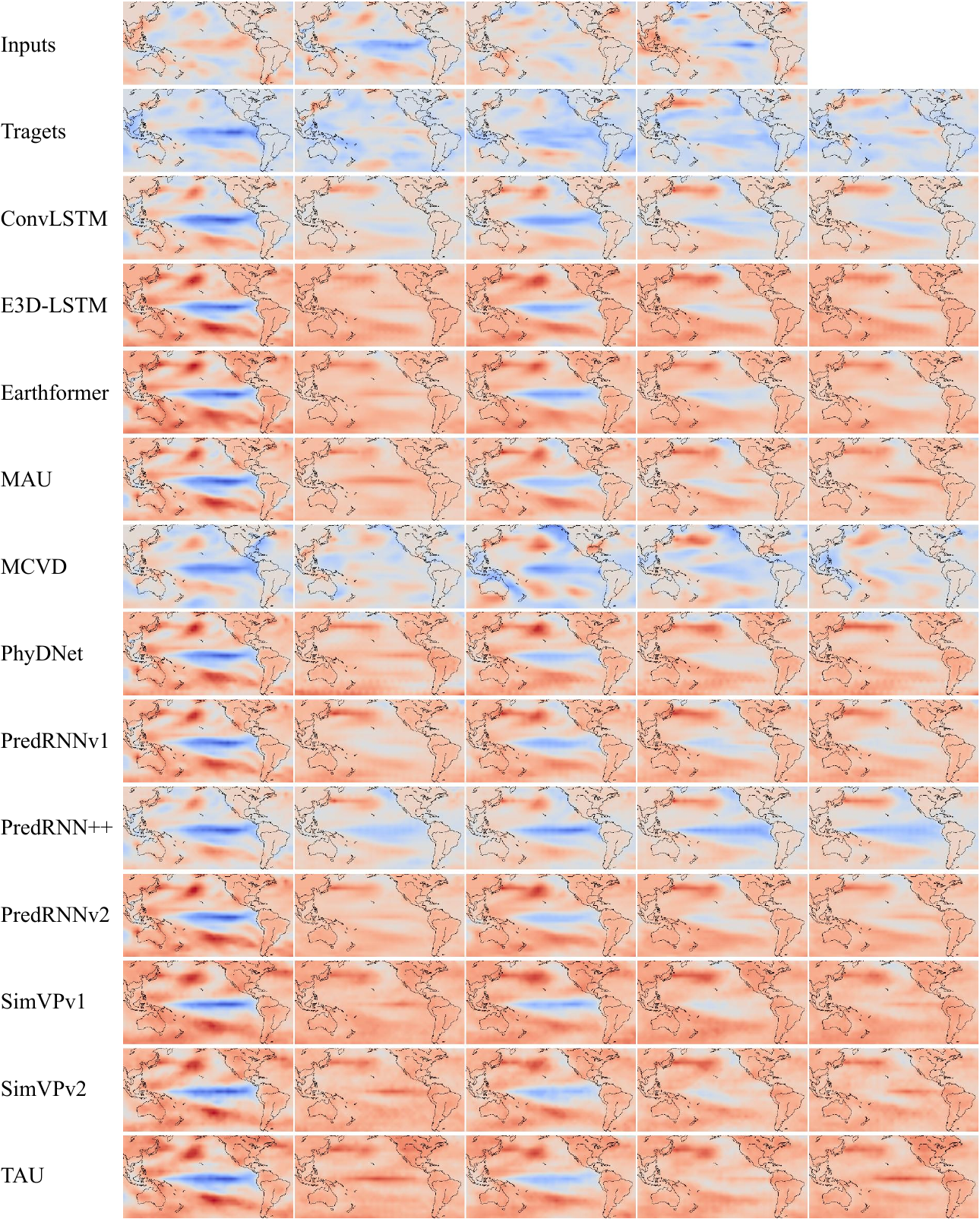}
    \caption{
        \textbf{Qualitative results} on \textit{ICAR-ENSO}~\cite{ICARENSO} (12 frames $\longrightarrow$ 14 frames). The sequences are visualized at the interval of 3 frames.
    }
    \vspace{-0.3cm}
    \label{fig:qualitative_enso}
\end{figure*}

\begin{figure*}[t]
    \centering
    \includegraphics[width=0.45\linewidth]{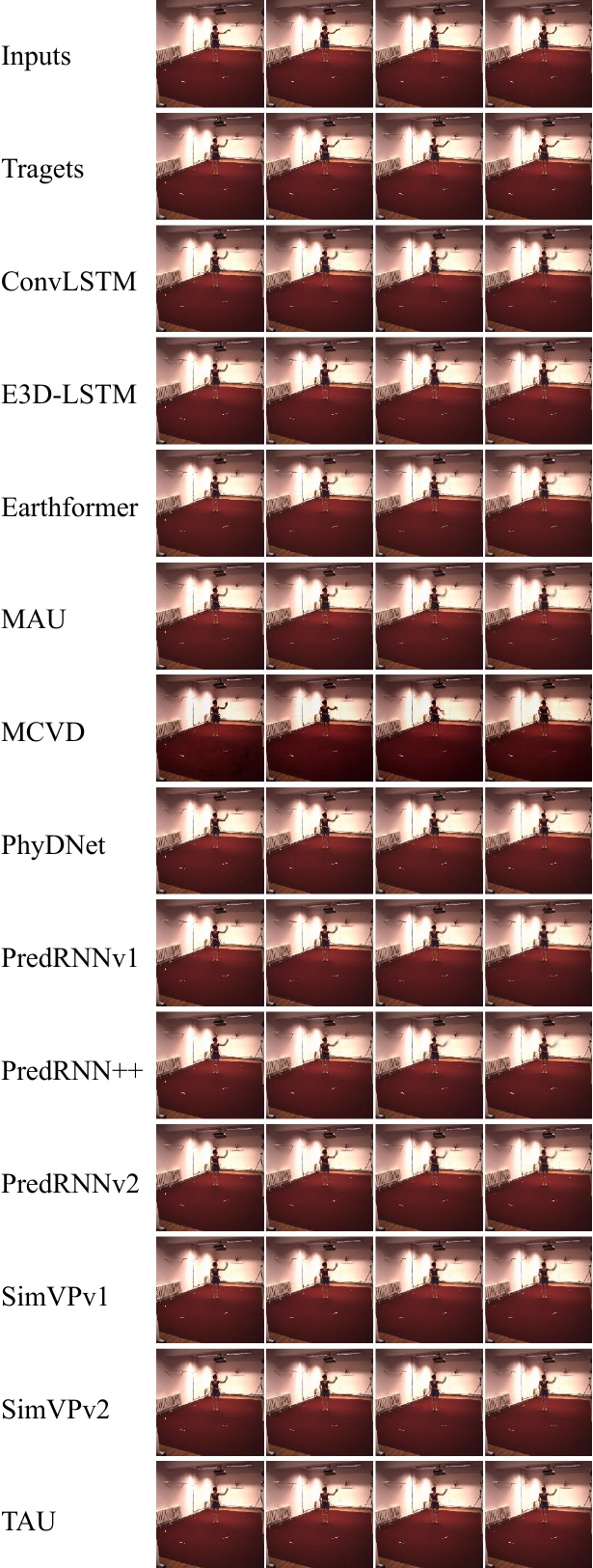}
    \caption{
        \textbf{Qualitative results} on \textit{Human3.6M}~\cite{ionescu2013human} (4 frames $\longrightarrow$ 4 frames).
    }
    \vspace{-0.3cm}
    \label{fig:qualitative_human}
\end{figure*}

\begin{figure*}[t]
    \centering
    \includegraphics[width=1.0\linewidth]{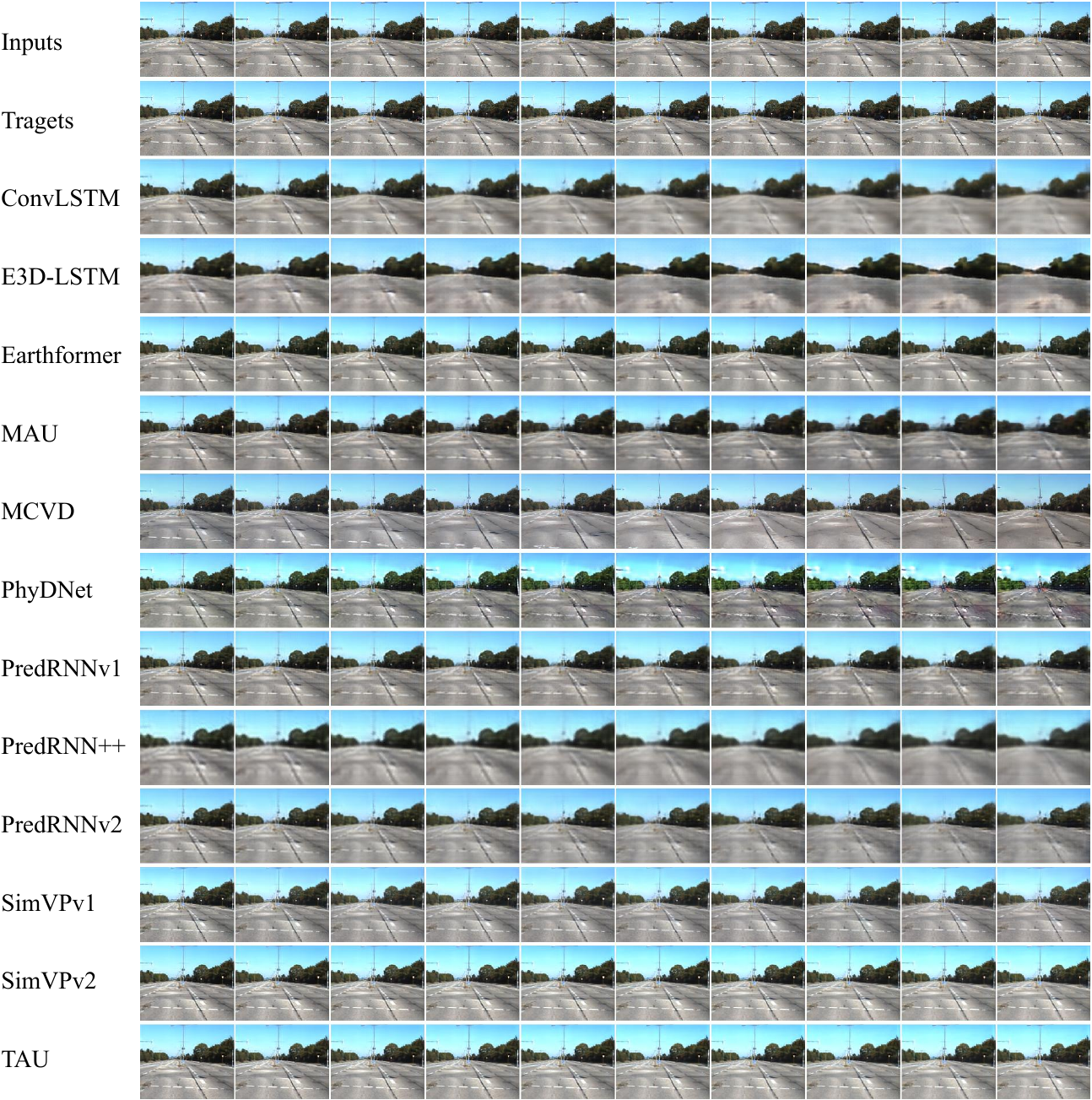}
    \caption{
        \textbf{Qualitative results} on \textit{KITTI}~\cite{geiger2013kitti} (10 frames $\longrightarrow$ 10 frames).
    }
    \vspace{-0.3cm}
    \label{fig:qualitative_kitti}
\end{figure*}

\begin{figure*}[t]
    \centering
    \includegraphics[width=1.0\linewidth]{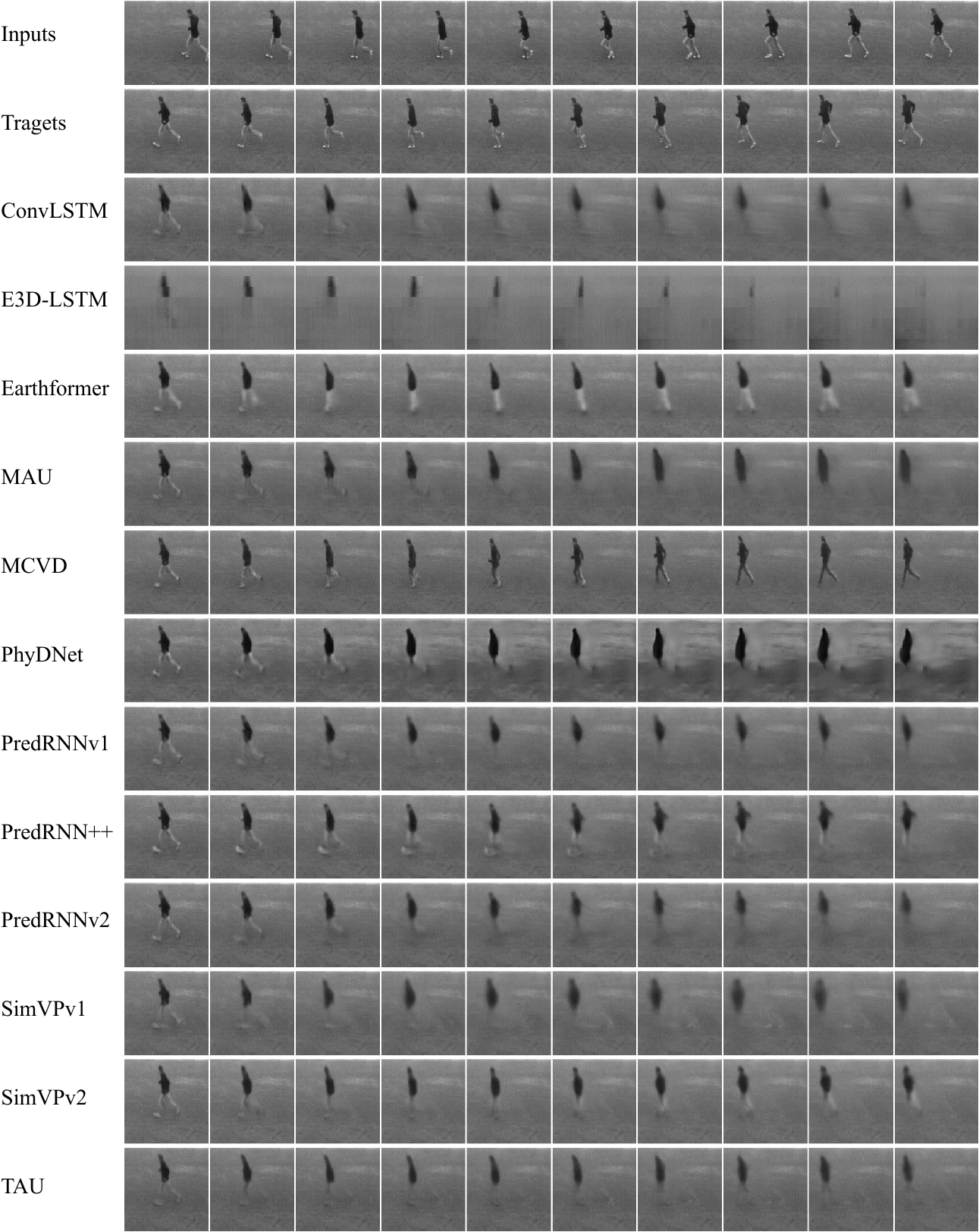}
    \caption{
        \textbf{Qualitative results} on \textit{KTH}~\cite{schuldt2004kth}  (10 frames $\longrightarrow$ 10 frames).
    }
    \vspace{-0.3cm}
    \label{fig:qualitative_kth}
\end{figure*}

\begin{figure*}[t]
    \centering
    \includegraphics[width=1.0\linewidth]{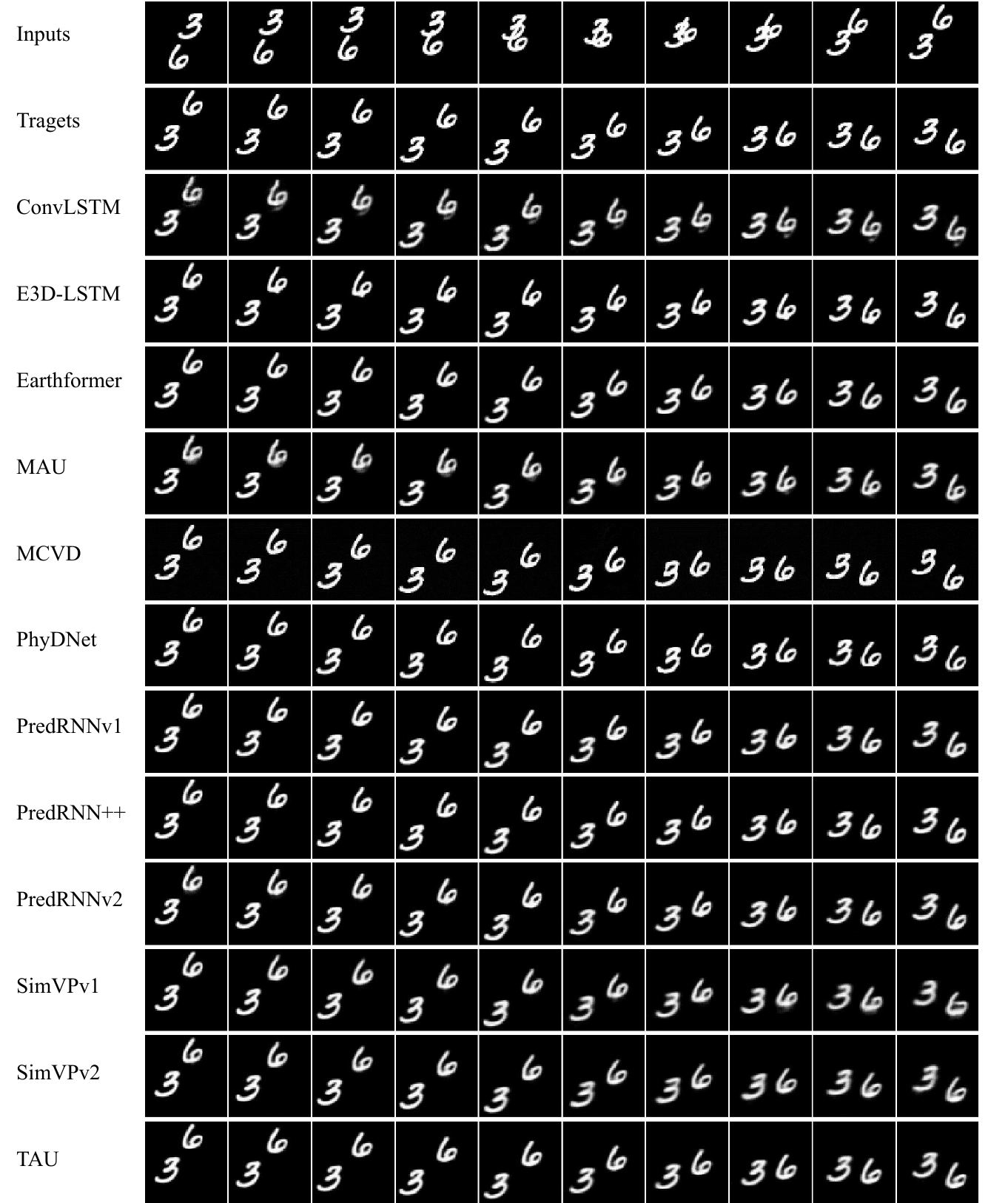}
    \caption{
        \textbf{Qualitative results} on \textit{Moving-MNIST}~\cite{srivastava2015movingmnist} (10 frames $\longrightarrow$ 10 frames).
    }
    \vspace{-0.3cm}
    \label{fig:qualitative_mnist}
\end{figure*}

\begin{figure*}[t]
    \centering
    \includegraphics[width=1.0\linewidth]{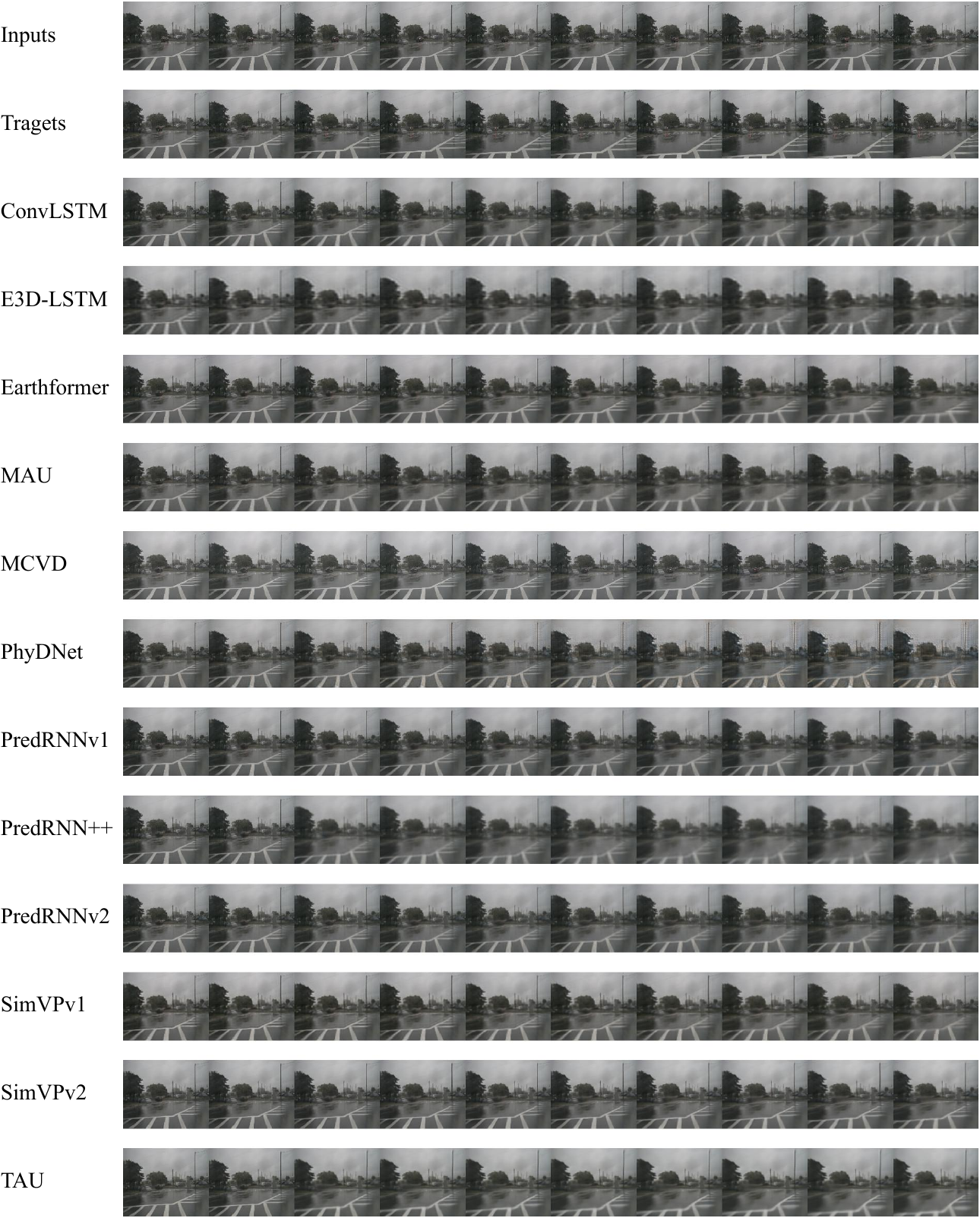}
    \caption{
        \textbf{Qualitative results} on \textit{nuScenes}~\cite{caesar2020nuscenes} (10 framse $\longrightarrow$ 10 frames).
    }
    \vspace{-0.3cm}
    \label{fig:qualitative_nuscenes}
\end{figure*}

\begin{figure*}[t]
    \centering
    \includegraphics[width=1.0\linewidth]{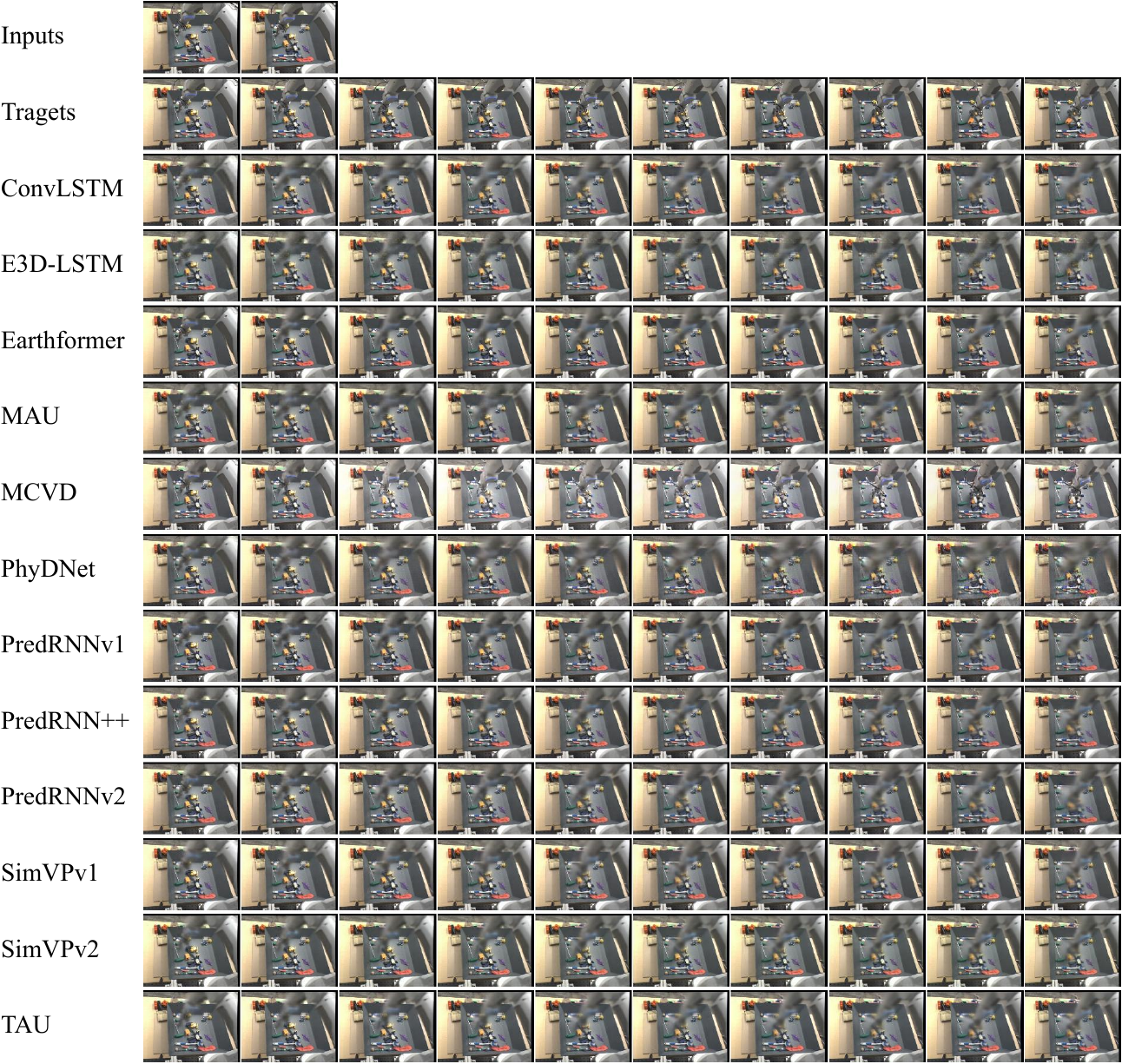}
    \caption{
        \textbf{Qualitative results} on \textit{RoboNet}~\cite{dasari2019robonet} (2 framse $\longrightarrow$ 10 frames).
    }
    \vspace{-0.3cm}
    \label{fig:qualitative_robonet}
\end{figure*}

\begin{figure*}[t]
    \centering
    \includegraphics[width=0.7\linewidth]{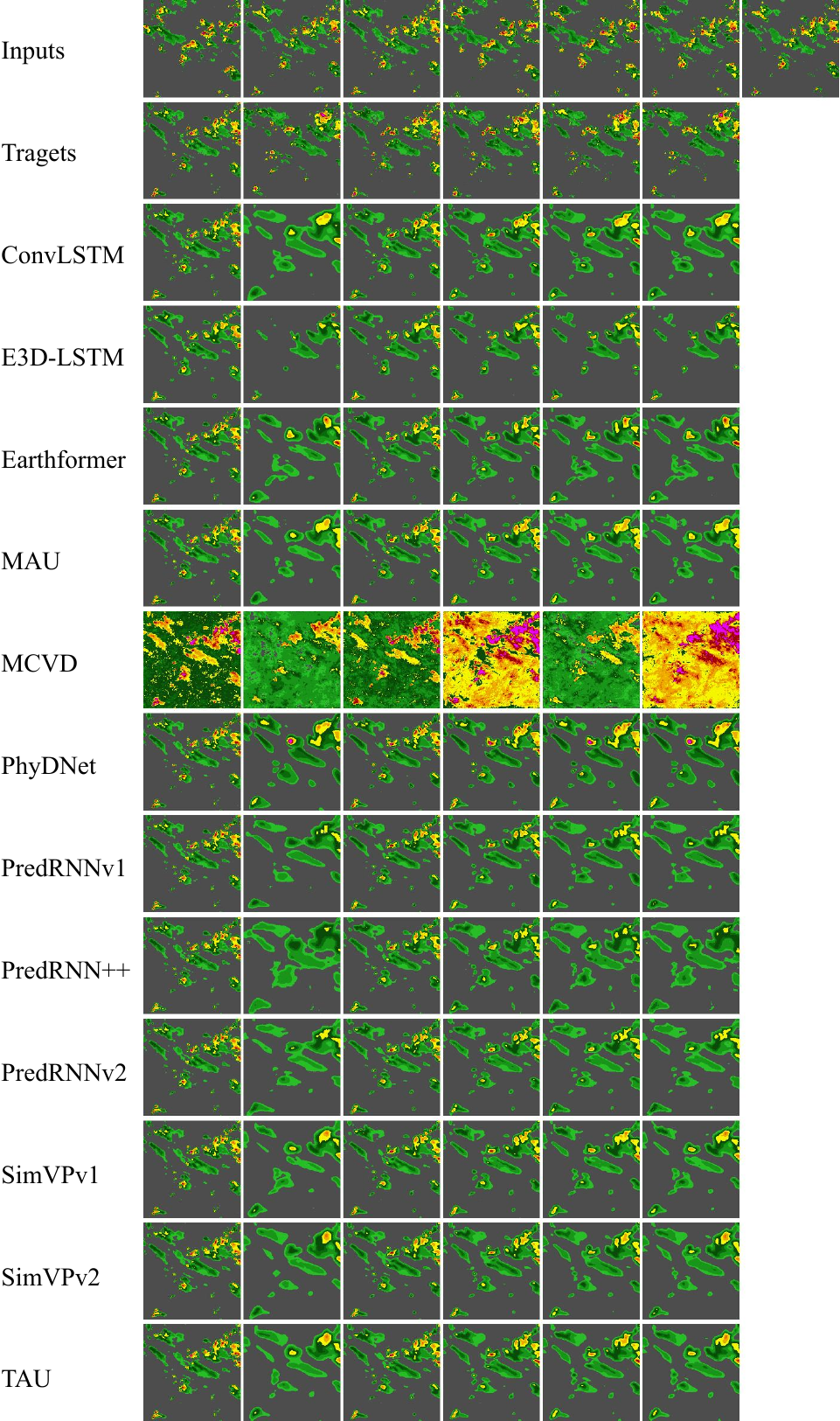}
    \caption{
        \textbf{Qualitative results} on \textit{SEVIR}~\cite{veillette2020sevir} (13 framse $\longrightarrow$ 12 frames). The sequences are visualized at the interval of 2 frames.
    }
    \vspace{-0.3cm}
    \label{fig:qualitative_sevir}
\end{figure*}

\begin{figure*}[t]
    \centering
    \includegraphics[width=0.45\linewidth]{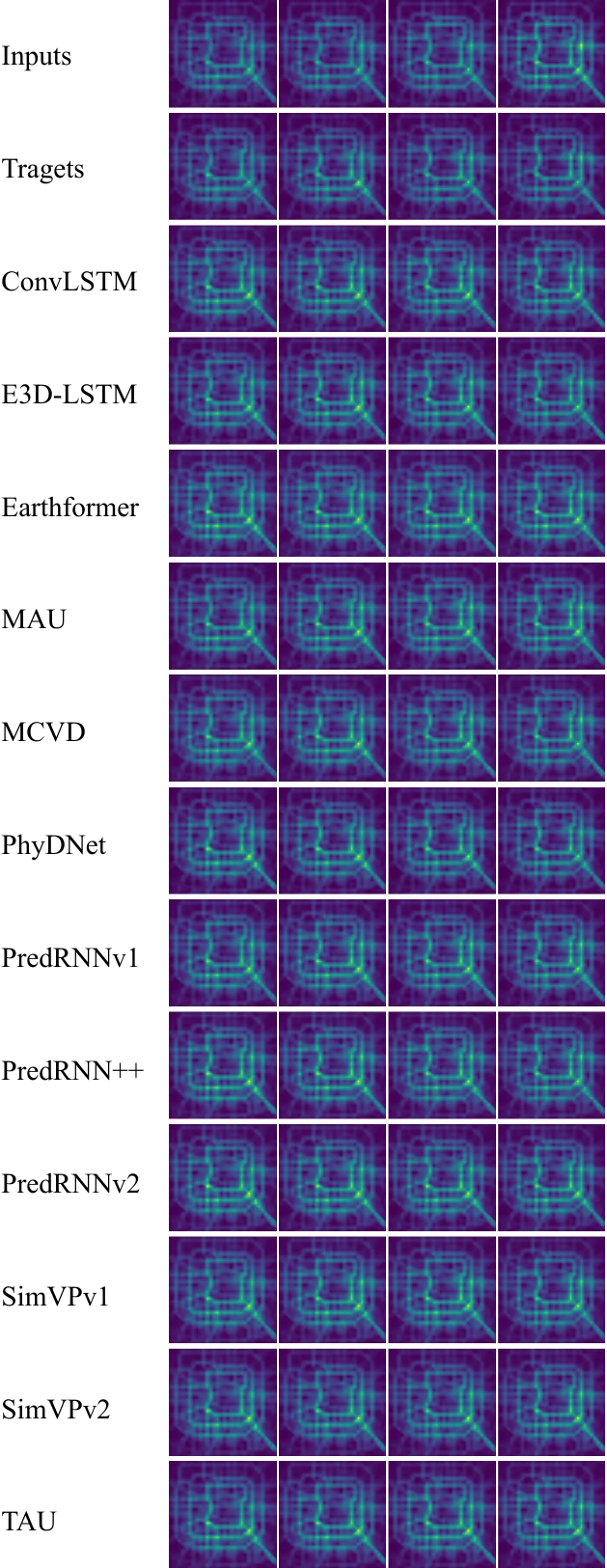}
    \caption{
        \textbf{Qualitative results} on \textit{TaxiBJ}~\cite{zhang2018taxibj} (4 framse $\longrightarrow$ 4 frames).
    }
    \vspace{-0.3cm}
    \label{fig:qualitative_taxibj}
\end{figure*}

\begin{figure*}[t]
    \centering
    \includegraphics[width=0.8\linewidth]{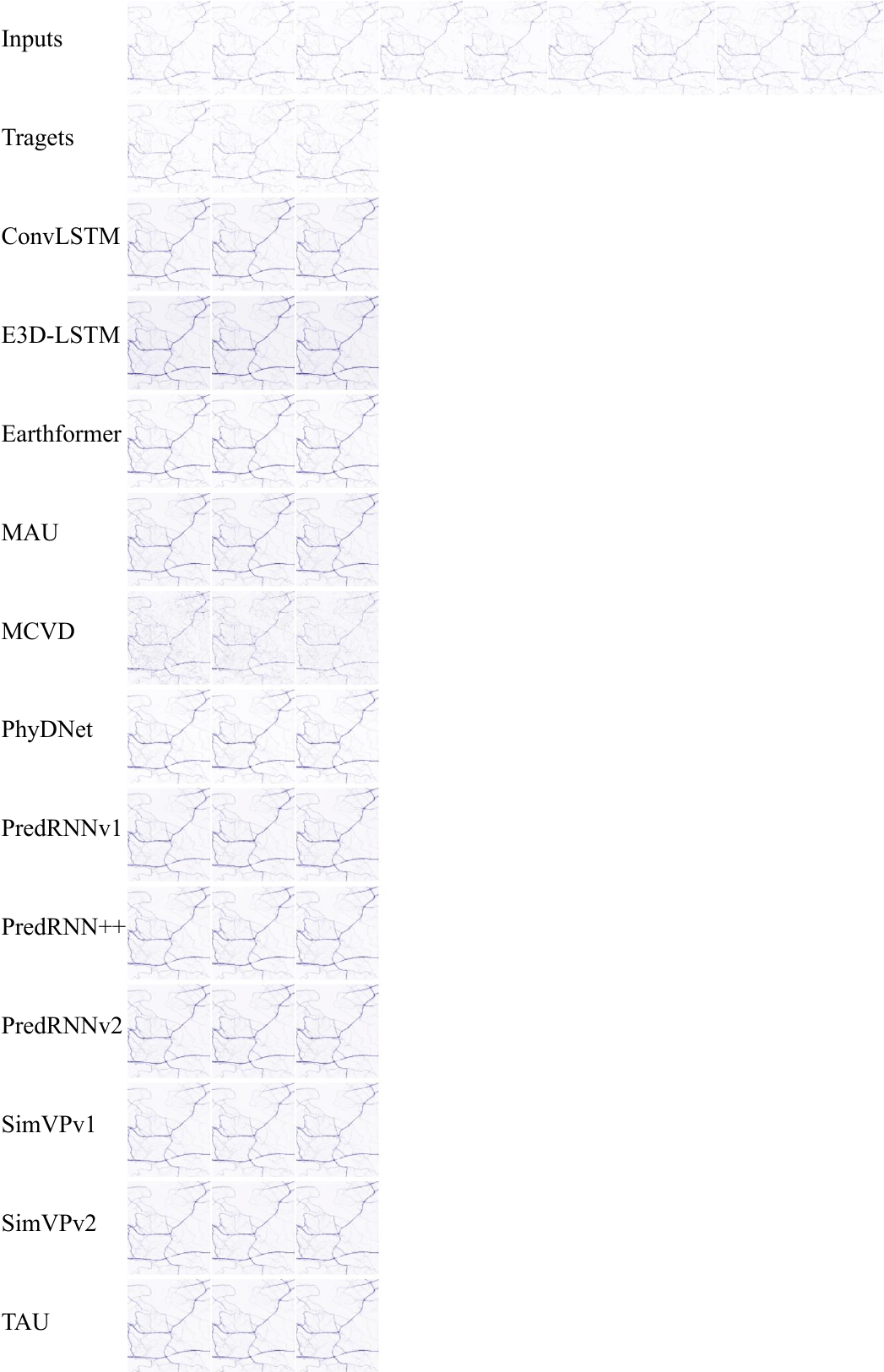}
    \caption{
        \textbf{Qualitative results} on \textit{Traffic4Cast2021}~\cite{Traffic4Cast2021} (9 framse $\longrightarrow$ 3 frames).
    }
    \vspace{-0.3cm}
    \label{fig:qualitative_traffic4cast2021}
\end{figure*}

\begin{figure*}[t]
    \centering
    \includegraphics[width=0.9\linewidth]{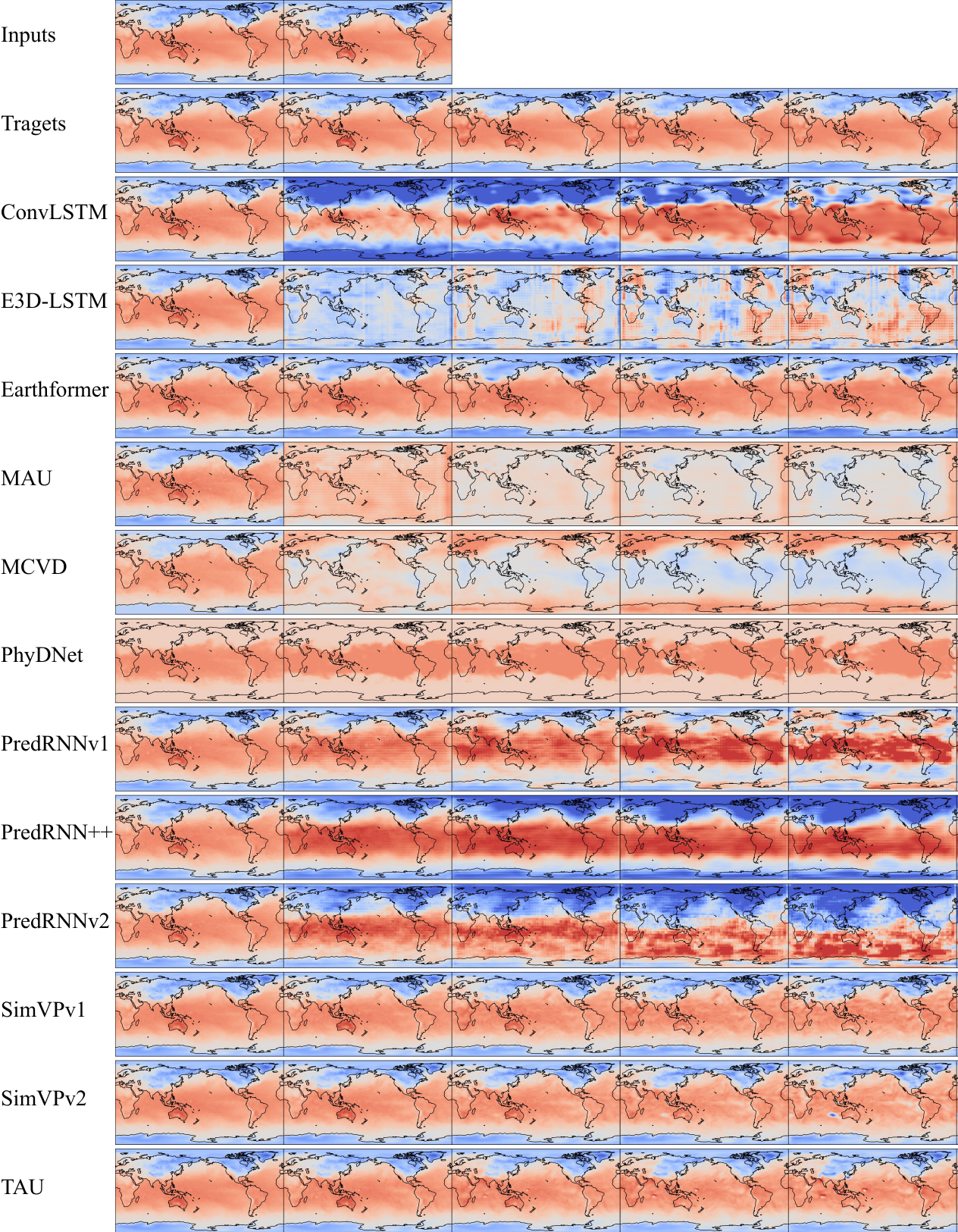}
    \caption{
        \textbf{Qualitative results} of t2m on \textit{WeatherBench}~\cite{garg2022weatherbench} (2 framse $\longrightarrow$ 20 frames). The target and predicted sequences are visualized at the interval of 4 frames. The models are learned to predict 1 frame based on 2 context frames, and the 2-20 frames in the predicted sequences are generated through extrapolation. 
    }
    \vspace{-0.3cm}
    \label{fig:qualitative_weatherbench_t2m}
\end{figure*}

\begin{figure*}[t]
    \centering
    \includegraphics[width=0.9\linewidth]{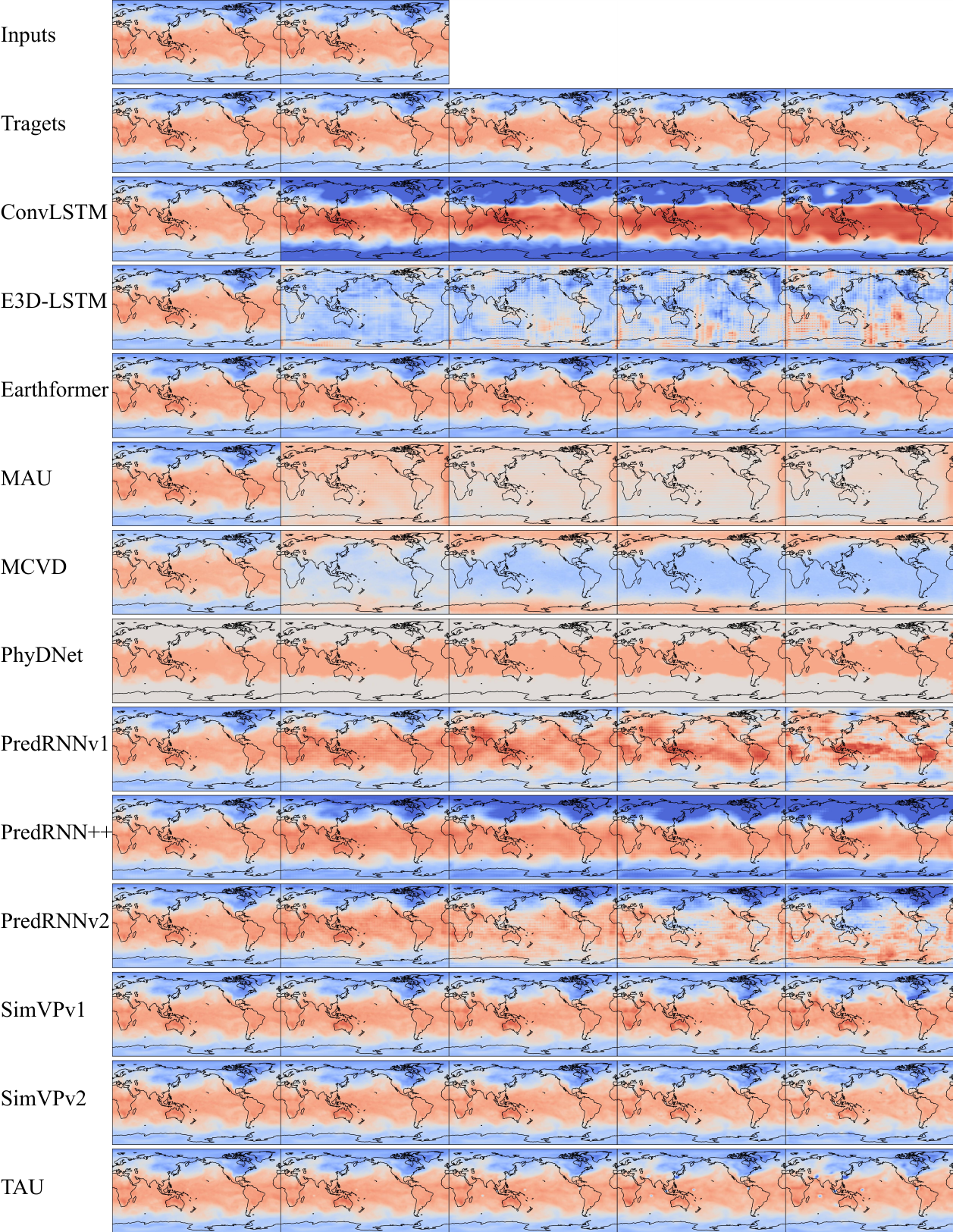}
    \caption{
        \textbf{Qualitative results} of t850 on \textit{WeatherBench}~\cite{garg2022weatherbench} (2 framse $\longrightarrow$ 20 frames). The target and predicted sequences are visualized at the interval of 4 frames. The models are learned to predict 1 frame based on 2 context frames, and the 2-20 frames in the predicted sequences are generated through extrapolation. 
    }
    \vspace{-0.3cm}
    \label{fig:qualitative_weatherbench_t850}
\end{figure*}

\begin{figure*}[t]
    \centering
    \includegraphics[width=0.9\linewidth]{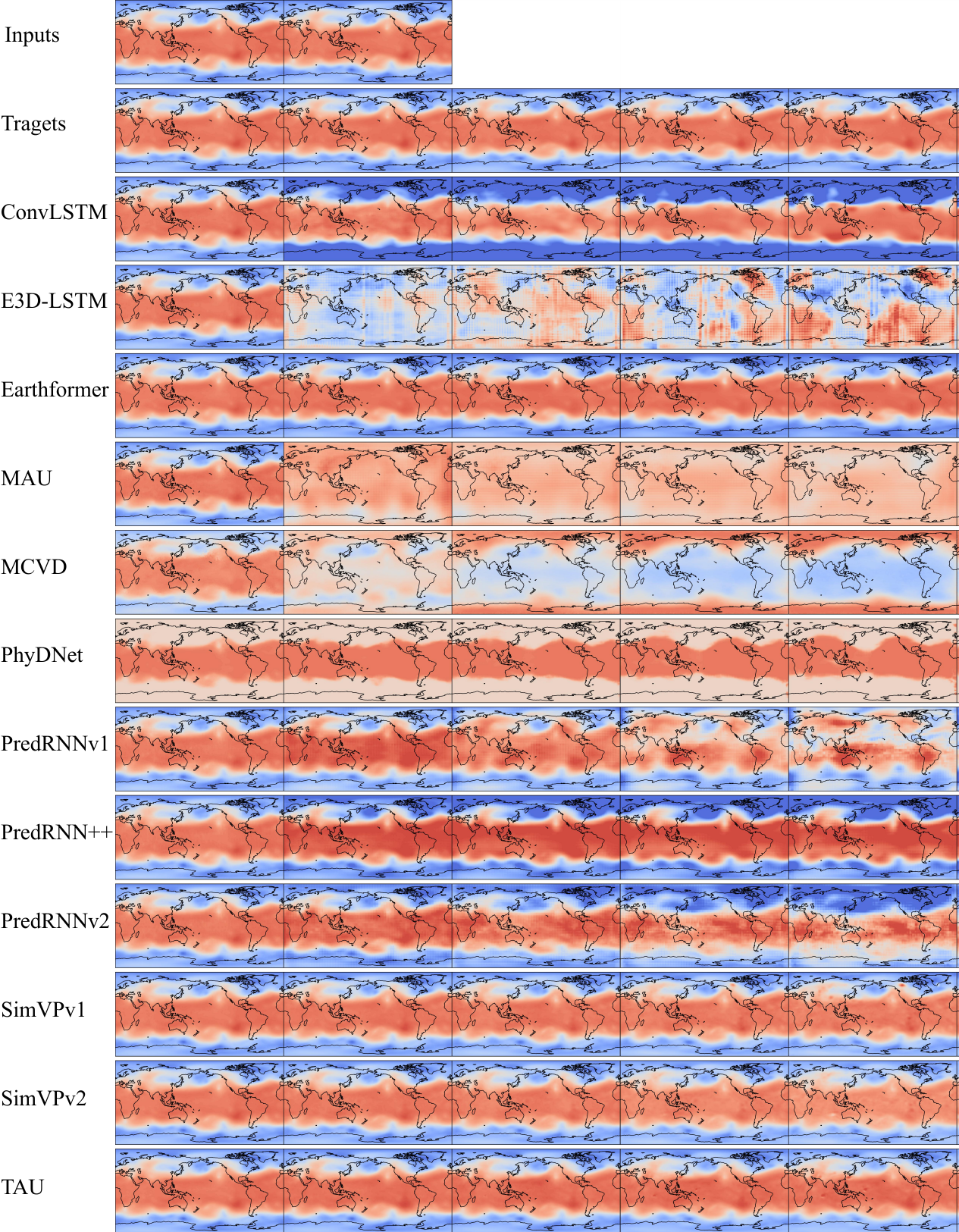}
    \caption{
        \textbf{Qualitative results} of z500 on \textit{WeatherBench}~\cite{garg2022weatherbench} (2 framse $\longrightarrow$ 20 frames). The target and predicted sequences are visualized at the interval of 4 frames. The models are learned to predict 1 frame based on 2 context frames, and the 2-20 frames in the predicted sequences are generated through extrapolation. 
    }
    \vspace{-0.3cm}
    \label{fig:qualitative_weatherbench_z500}
\end{figure*}

We provide the qualitative results of each model on these datasets, which are presented in Figs. \ref{fig:qualitative_bair}, \ref{fig:qualitative_bridgedata}, \ref{fig:qualitative_cityscapes}, \ref{fig:qualitative_enso}, \ref{fig:qualitative_human}, \ref{fig:qualitative_kitti}, \ref{fig:qualitative_kth}, \ref{fig:qualitative_mnist}, \ref{fig:qualitative_nuscenes}, \ref{fig:qualitative_robonet}, \ref{fig:qualitative_sevir}, \ref{fig:qualitative_taxibj}, \ref{fig:qualitative_traffic4cast2021}, \ref{fig:qualitative_weatherbench_t2m}, \ref{fig:qualitative_weatherbench_t850}, and \ref{fig:qualitative_weatherbench_z500}.

\section{Crowd Sourcing Human Assessment}
\label{sec:suppl_human}

As described in finding 2 in section 4.1 of the main paper, we find that LPIPS and FVD metrics are more aptly suited for tasks involving visual prediction, as they exhibit a stronger correlation with the human visual system. Furthermore, we have conducted a crowd-sourced human study to determine the most suitable metric for evaluating visual prediction models. 

\begin{figure*}[t]
    \centering
    \includegraphics[width=0.5\linewidth]{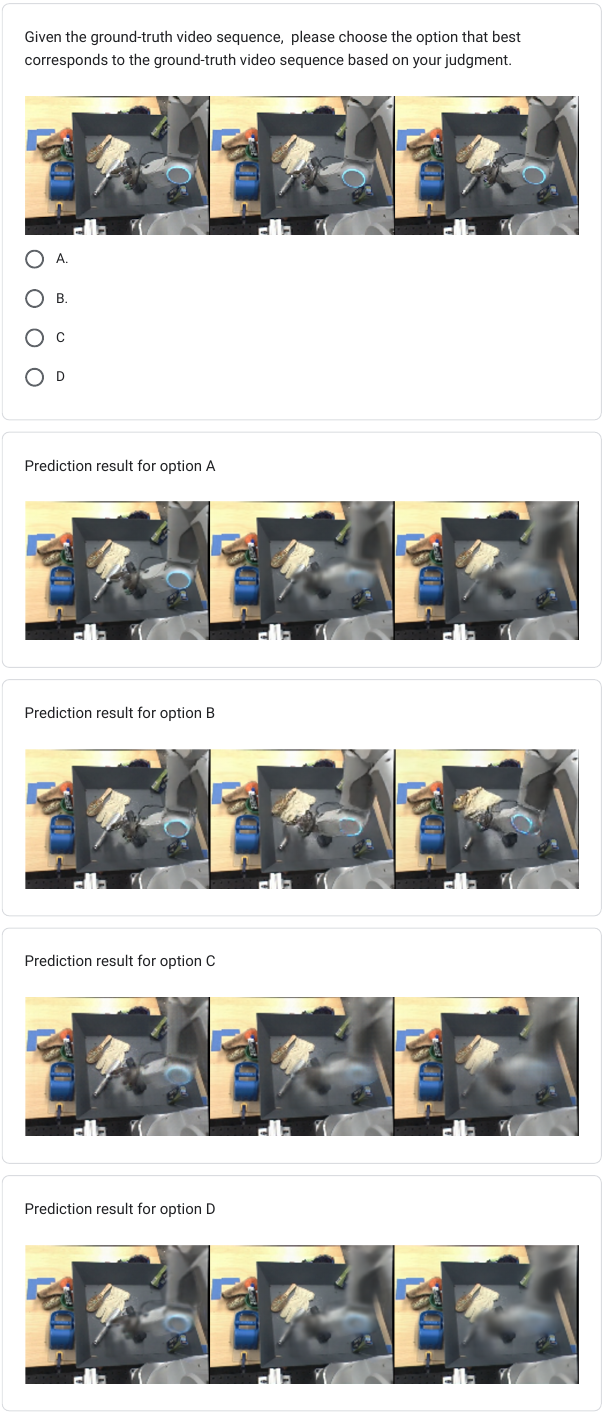}
    \caption{
         \textbf{An example of the human assessment questionnaire.} Given the ground-truth sequence, the user is required to select the predicted sequence that has the highest quality compared with the target. The predicted sequences for options A, B, C, and D are generated from Earthformer~\cite{gao2022earthformer}, MCVD~\cite{voleti2022mcvd}, PredRNN++~\cite{wang2018predrnn++}, and TAU~\cite{tan2023tau}. To ensure a fair and unprejudiced comparison, we have deliberately concealed the specific model information in the option descriptions.
    }
    \vspace{-0.3cm}
    \label{fig:human_assessment}
\end{figure*}

Notably, we find that the sequences predicted by MCVD~\cite{voleti2022mcvd} have the highest FVD and LPIPS, indicating a closer resemblance to human perception. However, these sequences performed poorly in terms of SSIM and PSNR. Conversely, methods such as Earthformer~\cite{gao2022earthformer}, PredRNN++~\cite{wang2018predrnn++} and TAU~\cite{tan2023tau} excel on SSIM and PSNR, while demonstrating inferior performance on FVD and LPIPS. To further validate our observations, we randomly sample the predicted results from Earthformer, MCVD, PredRNN++, and TAU on three representative datasets: \textit{BAIR} (15 sequences), \textit{RoboNet} (15 sequences), and \textit{Human3.6M} (5 sequences). We have designed a questionnaire, as illustrated in \cref{fig:human_assessment}, for the human assessment of these sampled results.

We have collected 100 crowd-sourced human evaluation questionnaires, and the feedback from these questionnaires solidified our observations. MCVD receives the highest rating as the best-predicted result in $72.83\%$ of the questions, followed by Earthformer with $4.35\%$, PredRNN++ with $19.57\%$, and TAU with $3.26\%$. These results further validate that LPIPS and FVD metrics are more effective in capturing the genuine visual effects of the predicted sequences.

\section{Discussion}
\label{sec:suppl_discuss}

\subsection{Broader Impact}

\noindent \textbf{Academic Impact}

In this work, we introduce PredBench, a comprehensive benchmark supporting diverse tasks and methods. PredBench integrates 12 established STP methods, covering CNN~\cite{lecun1989cnn,pu2023adaptive}, RNN~\cite{hochreiter1997lstm}, transformer\cite{dosovitskiy2020vit,wu2024llama}, and diffusion~\cite{song2020score,lu2024fit,guo2024smooth,pu2024efficientdit,lu2024hierarchical}.
Through standard experiments and multi-dimension evaluations on 15 diverse datasets, we thoroughly assess the performance of each model. We open-source our extensive codebase, serving as a valuable resource for researchers seeking to advance the state-of-the-art in spatio-temporal prediction.

\noindent \textbf{Social Impact}

Spatio-temporal prediction is a rapidly evolving field with significant implications across a wide range of domains and disciplines. The ability to accurately predict future states based on spatial and temporal data can drive advancements in numerous areas, including meteorology~\cite{ravuri2021skiful,bi2023pangu,lam2022graphcast,chen2023fengwu,han2024weather5k,gong2024cascast,xu2024extremecast,xu2024generalizing,han2024cra5,ling2024diffusion}, robotics~\cite{finn2016unsupervised,du2023learning,cao2024smart,zhang2022teleoperation}, generation\cite{li2024survey, lu2023seeing}, and autonomous vehicles~\cite{hu2023gaia,gao2023review}. Our PredBench conducts experiments and evaluations on diverse applications, aimed at providing meaningful results for social and industrial communities.

\subsection{Limitation}

While this work has provided prevalent methods, representative datasets, and several powerful benchmarks, this section explores the limitations expected to be addressed in future studies.

\noindent{\textbf{Training Limination.}} In pursuit of a fair comparison, we maintain the model architecture and model size with the original paper. However, specific architecture improvements or larger model size may yield enhanced results.

\noindent{\textbf{Benchmark Limination.}} Although we have implemented 12 methods in our benchmark, we will provide more methods in the future to provide a wider method spectrum. Besides, we have meticulously calibrated the dataset protocol, but there is still a lot of work to be done, such as the impact of the number of input frames.

\noindent{\textbf{Evaluation Limination.}} Due to resource limitations, our human evaluation only recruits 100 participants. Our human evaluation also lacks diversity in terms of participant background, as it only includes a few attributes such as age and gender. We hope that future work can improve the diversity and size of the participants. Furthermore, we hope explore more evaluation approaches and metrics to present a holistic assessment of models.

\begin{table}[t]
\renewcommand\arraystretch{1.4}
\setlength\tabcolsep{1pt}
\center
\vskip 0.15in
\caption{computational efficiency analysis for each model.} 
\label{tab:model_size}
\vspace{-0.3cm}
\begin{adjustbox}{max width=1.\columnwidth}
\begin{tabular}{lcccccccccccc}
\toprule[1.3pt]
  \textbf{Model} & ConvLSTM & E3D-LSTM & MAU & PhyDNet & PredRNNv1 & PredRNN++ & PredRNNv2 & SimVPv1 & SimVPv2 & TAU & Earthformer & MCVD \\
  \Xhline{0.5pt}

  params & 12.09M & 51.35M & 4.475M & 3.092M & 23.84M & 36.028M & 23.86M & 57.95M & 46.77M & 44.66M & 6.702M & 54.29M \\
  FLOPs & 58.80G & 299.0M & 17.79G & 15.33G & 116.0M & 175.0M & 117.0M & 19.43G & 16.53G & 15.95G & 33.65G & 29.15G \\
  FPS & 247.9 & 36.1 & 156.8 & 340.4 & 119.4 & 84.6 & 115.1 & 428.3 & 435.3 & 442.1 & 54.4 & 261.7 \\
  
\bottomrule[0.3pt]
\end{tabular}
\end{adjustbox}
\vspace{-0.3cm}
\end{table}

\begin{table*}[t]
\renewcommand\arraystretch{1.4}
\setlength\tabcolsep{1pt}
\center
\vskip 0.15in
\caption{\textbf{Hyper-parameters} of ConvLSTM~\cite{shi2015convlstm}. In the first column, \uline{BS}, \uline{LR}, \uline{Optim}, and \uline{Schd} represent the batch size, learning rate, optimizer, and learning rate scheduler, respectively. In the header row, \uline{M-MNIST} means \textit{Moving-MNIST}~\cite{srivastava2015movingmnist}, \uline{Traffic4Cast} denotes \textit{Traffic4Cast2021}~\cite{Traffic4Cast2021}, and \uline{ENSO} represents \textit{ICAR-ENSO}~\cite{ICARENSO}. The \uline{OneCy} means the \textit{OneCycleLR} scheduler, while the \uline{Cosine} denotes the \textit{CosineLR} scheduler. Unless otherwise specified, we directly utilize the default parameters of the optimizer and scheduler. In \textit{TaxiBJ}~\cite{zhang2018taxibj}, we adopt $pct\_start=0.1$ in the OneCycleLR scheduler rather than the default $pct\_start=0.3$.}
\label{tab:suppl_imp_convlstm}
\vspace{-0.3cm}
\begin{adjustbox}{max width=1.\columnwidth}
\begin{tabular}{l|cccccccccccccc}
\toprule[1.3pt]
  \textbf{Config} 
  & \textbf{M-MNIST} 
  & \textbf{KTH} 
  & \textbf{Human3.6M} 
  & \textbf{BAIR} 
  & \textbf{RoboNet} 
  & \textbf{BridgeData} 
  & \textbf{CityScapes} 
  & \textbf{KITTI} 
  & \textbf{nuScenes} 
  & \textbf{TaxiBJ} 
  & \textbf{Traffic4Cast}
  & \textbf{ENSO} 
  & \textbf{SEVIR} 
  & \textbf{WeatherBench}
  \\
    
\Xhline{0.5pt}
    BS & 16 & 16 & 16 & 64 & 64 & 64 & 64 & 16 & 64 & 16 & 64 & 64 & 32 & 64 \\
    LR & 5e-4 & 4e-5 & 1e-4 & 1e-4 & 1e-4 & 1e-4 & 1e-4 & 1e-3 & 1e-4 & 5e-4 & 1e-4 & 1e-4 & 1e-3 & 1e-4 \\
    Optim & Adam & Adam & Adam & Adam & Adam & Adam & Adam  & Adam & Adam & Adam & Adam & Adam & Adam & Adam  \\
    Schd & OneCy & None & Cosine & OneCy & OneCy & OneCy & OneCy & OneCy & OneCy & OneCy & OneCy & OneCy & OneCy & OneCy \\
    Epoch & 200 & 100 & 50 & 100 & 100 & 100 & 100 & 100 & 100 & 50 & 50 & 100 & 100 & 50 \\ 
    Loss & L2 & L2 & L2 & L2 & L2 & L2 & L2 & L2 & L2 & L2 & L2 & L2 & L2 & L2 \\
    dtype & BF16 & BF16 & BF16 & BF16 & BF16 & BF16 & BF16 & BF16 & BF16 & BF16 & BF16 & BF16 & BF16 & BF16 \\
\bottomrule[1.3pt]
\end{tabular}
\end{adjustbox}
\vskip -0.2in
\end{table*}

\begin{table*}[t]
\renewcommand\arraystretch{1.4}
\setlength\tabcolsep{1pt}
\center
\vskip 0.15in
\caption{\textbf{Hyper-parameters} of E3D-LSTM~\cite{wang2019e3dlstm}. In \textit{TaxiBJ}~\cite{zhang2018taxibj}, we adopt $pct\_start=0.1$ in the OneCycleLR scheduler rather than the default $pct\_start=0.3$.}
\label{tab:suppl_imp_e3dlstm}
\vspace{-0.3cm}
\begin{adjustbox}{max width=1.\columnwidth}
\begin{tabular}{l|cccccccccccccc}
\toprule[1.3pt]
  \textbf{Config} 
  & \textbf{M-MNIST} 
  & \textbf{KTH} 
  & \textbf{Human3.6M} 
  & \textbf{BAIR} 
  & \textbf{RoboNet} 
  & \textbf{BridgeData} 
  & \textbf{CityScapes} 
  & \textbf{KITTI} 
  & \textbf{nuScenes} 
  & \textbf{TaxiBJ} 
  & \textbf{Traffic4Cast}
  & \textbf{ENSO} 
  & \textbf{SEVIR} 
  & \textbf{WeatherBench}
  \\
    
\Xhline{0.5pt}
    BS & 16 & 8 & 16 & 64 & 64 & 64 & 64 & 16 & 64 & 16 & 64 & 32 & 64 & 64 \\ 
    LR & 1e-4 & 5e-4 & 1e-4 & 1e-4 & 1e-4 & 1e-4 & 1e-4 & 1e-3 & 1e-4 & 2e-4 & 1e-4 & 1e-3 & 1e-3 & 1e-4 \\ 
    Optim & Adam & Adam & Adam & Adam & Adam & Adam & Adam  & Adam & Adam & Adam & Adam & Adam & Adam & Adam  \\
    Sch & OneCy & OneCy & Cosine & OneCy & OneCy & OneCy & OneCy & OneCy & OneCy & None & OneCy & OneCy & OneCy & OneCy \\ 
    Epoch & 200 & 100 & 50 & 100 & 200 & 100 & 100 & 100 & 200 & 50 & 50 & 100 & 100 & 50 \\ 
    Loss & L2+L1 & L2+L1 & L2+L1 & L2+L1 & L2+L1 & L2+L1 & L2+L1 & L2+L1 & L2+L1 & L2+L1 & L2 & L2+L1 & L2+L1 & L2+L1 \\
    dtype & BF16 & BF16 & BF16 & BF16 & BF16 & BF16 & BF16  & BF16 & BF16 & BF16 & BF16 & BF16 & BF16 & BF16 \\
\bottomrule[1.3pt]
\end{tabular}
\end{adjustbox}
\vskip -0.2in
\end{table*}

\begin{table*}[t]
\renewcommand\arraystretch{1.4}
\setlength\tabcolsep{1pt}
\center
\vskip 0.15in
\caption{\textbf{Hyper-parameters} of MAU~\cite{chang2021mau}. In \textit{TaxiBJ}~\cite{zhang2018taxibj}, we adopt $pct\_start=0.1$ in the OneCycleLR scheduler rather than the default $pct\_start=0.3$.}
\label{tab:suppl_imp_mau}
\vspace{-0.3cm}
\begin{adjustbox}{max width=1.\columnwidth}
\begin{tabular}{l|cccccccccccccc}
\toprule[1.3pt]
  \textbf{Config} 
  & \textbf{M-MNIST} 
  & \textbf{KTH} 
  & \textbf{Human3.6M} 
  & \textbf{BAIR} 
  & \textbf{RoboNet} 
  & \textbf{BridgeData} 
  & \textbf{CityScapes} 
  & \textbf{KITTI} 
  & \textbf{nuScenes} 
  & \textbf{TaxiBJ} 
  & \textbf{Traffic4Cast}
  & \textbf{ENSO} 
  & \textbf{SEVIR} 
  & \textbf{WeatherBench}
  \\
    
\Xhline{0.5pt}
    BS & 16 & 16 & 16 & 64 & 64 & 64 & 64 & 16 & 64 & 16 & 64 & 64 & 32 & 64 \\ 
    LR & 1e-3 & 5e-4 & 1e-4 & 1e-4 & 1e-4 & 1e-4 & 1e-4 & 1e-3 & 1e-4 & 5e-4 & 1e-4 & 1e-4 & 1e-3 & 1e-4 \\ 
    Optim & Adam & Adam & Adam & Adam & Adam & Adam & Adam  & Adam & Adam & Adam & Adam & Adam & Adam & Adam  \\
    Sch & OneCy & OneCy & Cosine & OneCy & OneCy & OneCy & OneCy & OneCy & OneCy & OneCy & OneCy & OneCy & OneCy & OneCy \\ 
    Epoch & 200 & 100 & 50 & 100 & 200 & 100 & 100 & 100 & 200 & 50 & 50 & 100 & 100 & 50 \\ 
    Loss & L2 & L2 & L2 & L2 & L2 & L2 & L2 & L2 & L2 & L2 & L2 & L2 & L2 & L2 \\
    dtype & BF16 & BF16 & BF16 & BF16 & BF16 & BF16 & BF16  & BF16 & BF16 & BF16 & BF16 & BF16 & BF16 & BF16 \\
\bottomrule[1.3pt]
\end{tabular}
\end{adjustbox}
\end{table*}

\begin{table*}[t]
\renewcommand\arraystretch{1.4}
\setlength\tabcolsep{1pt}
\center
\vskip 0.15in
\caption{\textbf{Hyper-parameters} of PhyDNet~\cite{Guen2020phydnet}. In \textit{TaxiBJ}~\cite{zhang2018taxibj}, we adopt $pct\_start=0.1$ in the OneCycleLR scheduler rather than the default $pct\_start=0.3$. \uline{CM} represents its proposed kernel moment loss, and $\lambda_{CM}$ is its scaling factor.}
\label{tab:suppl_imp_phydnet}
\vspace{-0.3cm}
\begin{adjustbox}{max width=1.\columnwidth}
\begin{tabular}{l|cccccccccccccc}
\toprule[1.3pt]
  \textbf{Config} 
  & \textbf{M-MNIST} 
  & \textbf{KTH} 
  & \textbf{Human3.6M} 
  & \textbf{BAIR} 
  & \textbf{RoboNet} 
  & \textbf{BridgeData} 
  & \textbf{CityScapes} 
  & \textbf{KITTI} 
  & \textbf{nuScenes} 
  & \textbf{TaxiBJ} 
  & \textbf{Traffic4Cast}
  & \textbf{ENSO} 
  & \textbf{SEVIR} 
  & \textbf{WeatherBench}
  \\
    
\Xhline{0.5pt}
    BS & 16 & 16 & 16 & 64 & 64 & 64 & 64 & 16 & 64 & 16 & 64 & 64 & 32 & 64 \\ 
    LR & 1e-3 & 1e-3 & 1e-4 & 1e-4 & 1e-4 & 1e-4 & 1e-4 & 1e-3 & 1e-4 & 5e-4 & 1e-4 & 1e-4 & 1e-3 & 1e-4 \\ 
    Optim & Adam & Adam & Adam & Adam & Adam & Adam & Adam  & Adam & Adam & Adam & Adam & Adam & Adam & Adam  \\
    Sch & OneCy & OneCy & Cosine & OneCy & OneCy & OneCy & OneCy & OneCy & OneCy & OneCy & OneCy & OneCy & OneCy & OneCy \\ 
    Epoch & 200 & 100 & 50 & 100 & 200 & 100 & 100 & 100 & 200 & 50 & 50 & 100 & 100 & 50 \\ 
    Loss & L2+CM & L2+CM & L2+CM & L2+CM & L2+CM & L2+CM & L2+CM & L2+CM & L2+CM & L2+CM & L2+CM & L2+CM & L2+CM & L2+CM \\
    $\lambda_{CM}$ & 1.0 & 1.0 & 1.0 & 1.0 & 1.0 & 1.0 & 1.0 & 1.0 & 1.0 & 1.0 & 1.0 & 1.0 & 1.0 & 1.0 \\
    dtype & BF16 & BF16 & BF16 & BF16 & BF16 & BF16 & BF16  & BF16 & BF16 & BF16 & BF16 & BF16 & BF16 & BF16 \\
\bottomrule[1.3pt]
\end{tabular}
\end{adjustbox}
\vskip -0.2in
\end{table*}

\begin{table*}[t]
\renewcommand\arraystretch{1.4}
\setlength\tabcolsep{1pt}
\center
\vskip 0.15in
\caption{\textbf{Hyper-parameters} of PredRNNv1~\cite{wang2017predrnn}. In \textit{TaxiBJ}~\cite{zhang2018taxibj}, we adopt $pct\_start=0.1$ in the OneCycleLR scheduler rather than the default $pct\_start=0.3$.}
\label{tab:suppl_imp_predrnnv1}
\vspace{-0.3cm}
\begin{adjustbox}{max width=1.\columnwidth}
\begin{tabular}{l|cccccccccccccc}
\toprule[1.3pt]
  \textbf{Config} 
  & \textbf{M-MNIST} 
  & \textbf{KTH} 
  & \textbf{Human3.6M} 
  & \textbf{BAIR} 
  & \textbf{RoboNet} 
  & \textbf{BridgeData} 
  & \textbf{CityScapes} 
  & \textbf{KITTI} 
  & \textbf{nuScenes} 
  & \textbf{TaxiBJ} 
  & \textbf{Traffic4Cast}
  & \textbf{ENSO} 
  & \textbf{SEVIR} 
  & \textbf{WeatherBench}
  \\
    
\Xhline{0.5pt}
    BS & 16 & 16 & 16 & 64 & 64 & 64 & 64 & 16 & 64 & 16 & 64 & 64 & 32 & 64 \\ 
    LR & 5e-4 & 4e-5 & 1e-4 & 1e-4 & 1e-4 & 1e-4 & 1e-4 & 1e-3 & 1e-4 & 1e-4 & 1e-4 & 1e-4 & 1e-3 & 1e-4 \\ 
    Optim & Adam & Adam & Adam & Adam & Adam & Adam & Adam  & Adam & Adam & Adam & Adam & Adam & Adam & Adam  \\
    Sch & OneCy & OneCy & Cosine & OneCy & OneCy & OneCy & OneCy & OneCy & OneCy & OneCy & OneCy & OneCy & OneCy & OneCy \\ 
    Epoch & 200 & 100 & 50 & 100 & 200 & 100 & 100 & 100 & 200 & 50 & 50 & 100 & 100 & 50 \\ 
    Loss & L2 & L2 & L2 & L2 & L2 & L2 & L2 & L2 & L2 & L2 & L2 & L2 & L2 & L2 \\ 
    dtype & BF16 & BF16 & BF16 & BF16 & BF16 & BF16 & BF16  & BF16 & BF16 & BF16 & BF16 & BF16 & BF16 & BF16 \\
\bottomrule[1.3pt]
\end{tabular}
\end{adjustbox}
\vskip -0.2in
\end{table*}

\begin{table*}[t]
\renewcommand\arraystretch{1.4}
\setlength\tabcolsep{1pt}
\center
\vskip 0.15in
\caption{\textbf{Hyper-parameters} of PredRNN++~\cite{wang2018predrnn++}. In \textit{TaxiBJ}~\cite{zhang2018taxibj}, we adopt $pct\_start=0.1$ in the OneCycleLR scheduler rather than the default $pct\_start=0.3$.}
\label{tab:suppl_imp_predrnnpp}
\vspace{-0.3cm}
\begin{adjustbox}{max width=1.\columnwidth}
\begin{tabular}{l|cccccccccccccc}
\toprule[1.3pt]
  \textbf{Config} 
  & \textbf{M-MNIST} 
  & \textbf{KTH} 
  & \textbf{Human3.6M} 
  & \textbf{BAIR} 
  & \textbf{RoboNet} 
  & \textbf{BridgeData} 
  & \textbf{CityScapes} 
  & \textbf{KITTI} 
  & \textbf{nuScenes} 
  & \textbf{TaxiBJ} 
  & \textbf{Traffic4Cast}
  & \textbf{ENSO} 
  & \textbf{SEVIR} 
  & \textbf{WeatherBench}
  \\
    
\Xhline{0.5pt}
    BS & 16 & 16 & 16 & 64 & 64 & 64 & 64 & 16 & 64 & 16 & 64 & 64 & 32 & 64 \\ 
    LR & 1e-4 & 4e-5 & 1e-4 & 1e-4 & 1e-4 & 1e-4 & 1e-4 & 5e-4 & 1e-4 & 1e-4 & 1e-4 & 1e-4 & 1e-3 & 1e-4 \\ 
    Optim & Adam & Adam & Adam & Adam & Adam & Adam & Adam  & Adam & Adam & Adam & Adam & Adam & Adam & Adam  \\
    Sch & OneCy & OneCy & Cosine & OneCy & OneCy & OneCy & OneCy & OneCy & OneCy & OneCy & OneCy & OneCy & OneCy & OneCy \\ 
    Epoch & 200 & 100 & 50 & 100 & 200 & 100 & 100 & 100 & 200 & 50 & 50 & 100 & 100 & 50 \\ 
    Loss & L2 & L2 & L2 & L2 & L2 & L2 & L2 & L2 & L2 & L2 & L2 & L2 & L2 & L2 \\
    dtype & BF16 & BF16 & BF16 & BF16 & BF16 & BF16 & BF16  & BF16 & BF16 & BF16 & BF16 & BF16 & BF16 & BF16 \\
\bottomrule[1.3pt]
\end{tabular}
\end{adjustbox}
\vskip -0.2in
\end{table*}

\begin{table*}[t]
\renewcommand\arraystretch{1.4}
\setlength\tabcolsep{1pt}
\center
\vskip 0.15in
\caption{\textbf{Hyper-parameters} of PredRNNv2~\cite{wang2021predrnnv2}. In \textit{TaxiBJ}~\cite{zhang2018taxibj}, we adopt $pct\_start=0.1$ in the OneCycleLR scheduler rather than the default $pct\_start=0.3$. \uline{DC} means decouple loss proposed in PredRNNv2, and $\beta_{DC}$ is its scaling factor.}
\label{tab:suppl_imp_predrnnv2}
\vspace{-0.3cm}
\begin{adjustbox}{max width=1.\columnwidth}
\begin{tabular}{l|cccccccccccccc}
\toprule[1.3pt]
  \textbf{Config} 
  & \textbf{M-MNIST} 
  & \textbf{KTH} 
  & \textbf{Human3.6M} 
  & \textbf{BAIR} 
  & \textbf{RoboNet} 
  & \textbf{BridgeData} 
  & \textbf{CityScapes} 
  & \textbf{KITTI} 
  & \textbf{nuScenes} 
  & \textbf{TaxiBJ} 
  & \textbf{Traffic4Cast}
  & \textbf{ENSO} 
  & \textbf{SEVIR} 
  & \textbf{WeatherBench}
  \\
    
\Xhline{0.5pt}
    BS & 8 & 16 & 16 & 64 & 64 & 64 & 64 & 16 & 64 & 16 & 64 & 64 & 32 & 64 \\ 
    LR & 1e-4 & 1e-4 & 1e-4 & 1e-4 & 1e-4 & 1e-4 & 1e-4 & 1e-3 & 1e-4 & 5e-4 & 1e-4 & 1e-4 & 1e-3 & 1e-4 \\ 
    Optim & Adam & Adam & Adam & Adam & Adam & Adam & Adam  & Adam & Adam & Adam & Adam & Adam & Adam & Adam  \\
    Sch & OneCy & OneCy & Cosine & OneCy & OneCy & OneCy & OneCy & OneCy & OneCy & OneCy & OneCy & OneCy & OneCy & OneCy \\ 
    Epoch & 200 & 100 & 50 & 100 & 200 & 100 & 100 & 100 & 200 & 50 & 50 & 100 & 100 & 50 \\ 
    Loss & L2+DC & L2+DC & L2+DC & L2+DC & L2+DC & L2+DC & L2+DC & L2+DC & L2+DC & L2+DC & L2+DC & L2+DC & L2+DC & L2+DC \\ 
    $\beta_{DC}$ & 0.1 & 0.01 & 0.1 & 0.01 & 0.01 & 0.01 & 0.01 & 0.01 & 0.01 & 0.1 & 0.01 & 0.01 & 0.01 & 0.01 \\
    dtype & BF16 & BF16 & BF16 & BF16 & BF16 & BF16 & BF16  & BF16 & BF16 & BF16 & BF16 & BF16 & BF16 & BF16 \\
\bottomrule[1.3pt]
\end{tabular}
\end{adjustbox}
\vskip -0.2in
\end{table*}

\begin{table*}[t]
\renewcommand\arraystretch{1.4}
\setlength\tabcolsep{1pt}
\center
\vskip 0.15in
\caption{\textbf{Hyper-parameters} of SimVPv1~\cite{gao2022simvp}. In \textit{TaxiBJ}~\cite{zhang2018taxibj}, we adopt $pct\_start=0.1$ in the OneCycleLR scheduler rather than the default $pct\_start=0.3$.}
\label{tab:suppl_imp_simvpv1}
\vspace{-0.3cm}
\begin{adjustbox}{max width=1.\columnwidth}
\begin{tabular}{l|cccccccccccccc}
\toprule[1.3pt]
  \textbf{Config} 
  & \textbf{M-MNIST} 
  & \textbf{KTH} 
  & \textbf{Human3.6M} 
  & \textbf{BAIR} 
  & \textbf{RoboNet} 
  & \textbf{BridgeData} 
  & \textbf{CityScapes} 
  & \textbf{KITTI} 
  & \textbf{nuScenes} 
  & \textbf{TaxiBJ} 
  & \textbf{Traffic4Cast}
  & \textbf{ENSO} 
  & \textbf{SEVIR} 
  & \textbf{WeatherBench}
  \\
    
\Xhline{0.5pt}
    BS & 16 & 16 & 16 & 64 & 64 & 64 & 64 & 16 & 64 & 16 & 64 & 64 & 32 & 64 \\ 
    LR & 1e-3 & 1e-3 & 1e-4 & 1e-4 & 1e-3 & 1e-3 & 1e-4 & 5e-3 & 1e-3 & 1e-3 & 1e-4 & 1e-4 & 1e-3 & 1e-4 \\ 
    Optim & Adam & Adam & Adam & Adam & Adam & Adam & Adam  & Adam & Adam & Adam & Adam & Adam & Adam & Adam  \\
    Sch & OneCy & OneCy & Cosine & OneCy & OneCy & OneCy & OneCy & OneCy & OneCy & OneCy & OneCy & OneCy & OneCy & OneCy \\ 
    Epoch & 200 & 100 & 50 & 100 & 200 & 100 & 100 & 100 & 200 & 50 & 50 & 100 & 100 & 50 \\ 
    Loss & L2 & L2 & L2 & L2 & L2 & L2 & L2 & L2 & L2 & L2 & L2 & L2 & L2 & L2 \\
    dtype & BF16 & BF16 & BF16 & BF16 & BF16 & BF16 & BF16  & BF16 & BF16 & BF16 & BF16 & BF16 & BF16 & BF16 \\
\bottomrule[1.3pt]
\end{tabular}
\end{adjustbox}
\vskip -0.2in
\end{table*}

\begin{table*}[t]
\renewcommand\arraystretch{1.4}
\setlength\tabcolsep{1pt}
\center
\vskip 0.15in
\caption{\textbf{Hyper-parameters} of SimVPv2~\cite{tan2022simvpv2}. In \textit{TaxiBJ}~\cite{zhang2018taxibj}, we adopt $pct\_start=0.1$ in the OneCycleLR scheduler rather than the default $pct\_start=0.3$.}
\label{tab:suppl_imp_simvpv2}
\vspace{-0.3cm}
\begin{adjustbox}{max width=1.\columnwidth}
\begin{tabular}{l|cccccccccccccc}
\toprule[1.3pt]
  \textbf{Config} 
  & \textbf{M-MNIST} 
  & \textbf{KTH} 
  & \textbf{Human3.6M} 
  & \textbf{BAIR} 
  & \textbf{RoboNet} 
  & \textbf{BridgeData} 
  & \textbf{CityScapes} 
  & \textbf{KITTI} 
  & \textbf{nuScenes} 
  & \textbf{TaxiBJ} 
  & \textbf{Traffic4Cast}
  & \textbf{ENSO} 
  & \textbf{SEVIR} 
  & \textbf{WeatherBench}
  \\
    
\Xhline{0.5pt}
    BS & 16 & 16 & 16 & 64 & 64 & 64 & 64 & 16 & 64 & 16 & 64 & 64 & 32 & 64 \\ 
    LR & 1e-3 & 1e-3 & 1e-4 & 1e-4 & 1e-3 & 1e-3 & 1e-4 & 5e-3 & 1e-3 & 1e-3 & 1e-4 & 1e-4 & 1e-3 & 1e-4 \\ 
    Optim & Adam & Adam & Adam & Adam & Adam & Adam & Adam  & Adam & Adam & Adam & Adam & Adam & Adam & Adam  \\
    Sch & OneCy & OneCy & Cosine & OneCy & OneCy & OneCy & OneCy & OneCy & OneCy & OneCy & OneCy & OneCy & OneCy & OneCy \\ 
    Epoch & 200 & 100 & 50 & 100 & 200 & 100 & 100 & 100 & 200 & 50 & 50 & 100 & 100 & 50 \\ 
    Loss & L2 & L2 & L2 & L2 & L2 & L2 & L2 & L2 & L2 & L2 & L2 & L2 & L2 & L2 \\
    dtype & BF16 & BF16 & BF16 & BF16 & BF16 & BF16 & BF16  & BF16 & BF16 & BF16 & BF16 & BF16 & BF16 & BF16 \\
\bottomrule[1.3pt]
\end{tabular}
\end{adjustbox}
\vskip -0.2in
\end{table*}

\begin{table*}[t]
\renewcommand\arraystretch{1.4}
\setlength\tabcolsep{1pt}
\center
\vskip 0.15in
\caption{\textbf{Hyper-parameters} of TAU~\cite{tan2023tau}. In \textit{TaxiBJ}~\cite{zhang2018taxibj}, we adopt $pct\_start=0.1$ in the OneCycleLR scheduler rather than the default $pct\_start=0.3$. \uline{DDR} denotes the differential divergence regularization proposed in TAU, and $\alpha_{DDR}$ is its scaling factor.}
\label{tab:suppl_imp_tau}
\vspace{-0.3cm}
\begin{adjustbox}{max width=1.\columnwidth}
\begin{tabular}{l|cccccccccccccc}
\toprule[1.3pt]
  \textbf{Config} 
  & \textbf{M-MNIST} 
  & \textbf{KTH} 
  & \textbf{Human3.6M} 
  & \textbf{BAIR} 
  & \textbf{RoboNet} 
  & \textbf{BridgeData} 
  & \textbf{CityScapes} 
  & \textbf{KITTI} 
  & \textbf{nuScenes} 
  & \textbf{TaxiBJ} 
  & \textbf{Traffic4Cast}
  & \textbf{ENSO} 
  & \textbf{SEVIR} 
  & \textbf{WeatherBench}
  \\
    
\Xhline{0.5pt}
    BS & 64 & 64 & 64 & 64 & 16 & 16 & 16 & 16 & 64 & 64 & 32 & 16 & 64 & 64 \\ 
    LR & 1e-4 & 1e-3 & 1e-4 & 1e-4 & 1e-4 & 5e-3 & 1e-3 & 1e-3 & 1e-3 & 1e-3 & 1e-3 & 1e-3 & 1e-4 & 1e-4 \\ 
    Optim & Adam & Adam & Adam & Adam & Adam & Adam & Adam  & Adam & Adam & Adam & Adam & Adam & Adam & Adam  \\
    Sch & OneCy & OneCy & OneCy & OneCy & Cosine & OneCy & OneCy & OneCy & OneCy & OneCy & OneCy & OneCy & OneCy & OneCy \\ 
    Epoch & 100 & 100 & 100 & 100 & 50 & 100 & 100 & 200 & 200 & 200 & 100 & 50 & 50 & 50 \\
    Loss & L2+DDR & L2+DDR & L2+DDR & L2+DDR & L2+DDR & L2+DDR & L2+DDR & L2+DDR & L2+DDR & L2+DDR & L2+DDR & L2+DDR & L2+DDR & L2+DDR \\ 
    $\alpha_{DDR}$ & 0.1 & 0.1 & 0.1 & 0.1 & 0.1 & 0.1 & 0.1 & 0.1 & 0.1 & 0.1 & 0.1 & 0.1 & 0.1 & 0.1 \\ 
    dtype & BF16 & BF16 & BF16 & BF16 & BF16 & BF16 & BF16  & BF16 & BF16 & BF16 & BF16 & BF16 & BF16 & BF16 \\
\bottomrule[1.3pt]
\end{tabular}
\end{adjustbox}
\vskip -0.2in
\end{table*}

\begin{table*}[t]
\renewcommand\arraystretch{1.4}
\setlength\tabcolsep{1pt}
\center
\vskip 0.15in
\caption{\textbf{Hyper-parameters} of Earthformer~\cite{gao2022earthformer}. In the first column, \uline{WD} means weight decay of the optimizer, and \uline{Clip} represents that $clip\_grad$ is adopted with $max\_norm=1.0$.}
\label{tab:suppl_imp_earthformer}
\vspace{-0.3cm}
\begin{adjustbox}{max width=1.\columnwidth}
\begin{tabular}{l|cccccccccccccc}
\toprule[1.3pt]
  \textbf{Config} 
  & \textbf{M-MNIST} 
  & \textbf{KTH} 
  & \textbf{Human3.6M} 
  & \textbf{BAIR} 
  & \textbf{RoboNet} 
  & \textbf{BridgeData} 
  & \textbf{CityScapes} 
  & \textbf{KITTI} 
  & \textbf{nuScenes} 
  & \textbf{TaxiBJ} 
  & \textbf{Traffic4Cast}
  & \textbf{ENSO} 
  & \textbf{SEVIR} 
  & \textbf{WeatherBench}
  \\
    
\Xhline{0.5pt}
    BS & 32 & 32 & 32 & 64 & 64 & 64 & 64 & 32 & 64 & 32 & 64 & 64 & 32 & 64 \\ 
    Optim & AdamW & AdamW & AdamW & AdamW & AdamW & AdamW & AdamW  & AdamW & AdamW & AdamW & AdamW & AdamW & AdamW & AdamW  \\
    Sch & OneCy & OneCy & OneCy & OneCy & OneCy & OneCy & OneCy & OneCy & OneCy & OneCy & OneCy & OneCy & OneCy & OneCy \\ 
    WD & 1e-5 & 1e-5 & 1e-5 & 1e-5 & 1e-5 & 1e-5 & 1e-5 & 1e-5 & 1e-5 & 1e-5 & 1e-5 & 1e-5 & 1e-5 & 1e-5 \\
    Clip & 1.0 & 1.0 & 1.0 & 1.0 & 1.0 & 1.0 & 1.0 & 1.0 & 1.0 & 1.0 & 1.0 & 1.0 & 1.0 & 1.0 \\
    LR & 1e-3 & 1e-3 & 1e-3 & 1e-4 & 1e-3 & 1e-4 & 1e-4 & 1e-3 & 1e-3 & 1e-3 & 1e-4 & 1e-4 & 1e-3 & 1e-4 \\ 
    Epoch & 200 & 100 & 100 & 100 & 200 & 100 & 100 & 100 & 200 & 50 & 50 & 100 & 100 & 50 \\ 
    Loss & L2 & L2 & L2 & L2 & L2 & L2 & L2 & L2 & L2 & L2 & L2 & L2 & L2 & L2 \\
    dtype & BF16 & BF16 & BF16 & BF16 & BF16 & BF16 & BF16  & BF16 & BF16 & BF16 & BF16 & BF16 & BF16 & BF16 \\
\bottomrule[1.3pt]
\end{tabular}
\end{adjustbox}
\vskip -0.2in
\end{table*}

\begin{table*}[t]
\renewcommand\arraystretch{1.4}
\setlength\tabcolsep{1pt}
\center
\vskip 0.15in
\caption{\textbf{Hyper-parameters} of MCVD~\cite{voleti2022mcvd}. \uline{Linear} means the \textit{LinearLR} scheduler with 5000 iterations for warm-up. \uline{WD} means weight decay of the optimizer, and \uline{Clip} represents that $clip\_grad$ is adopted with $max\_norm=1.0$.}
\label{tab:suppl_impm_mcvd}
\vspace{-0.3cm}
\begin{adjustbox}{max width=1.\columnwidth}
\begin{tabular}{l|cccccccccccccc}
\toprule[1.3pt]
  \textbf{Config} 
  & \textbf{M-MNIST} 
  & \textbf{KTH} 
  & \textbf{Human3.6M} 
  & \textbf{BAIR} 
  & \textbf{RoboNet} 
  & \textbf{BridgeData} 
  & \textbf{CityScapes} 
  & \textbf{KITTI} 
  & \textbf{nuScenes} 
  & \textbf{TaxiBJ} 
  & \textbf{Traffic4Cast}
  & \textbf{ENSO} 
  & \textbf{SEVIR} 
  & \textbf{WeatherBench}
  \\
    
\Xhline{0.5pt}
    BS & 64 & 64 & 64 & 64 & 128 & 128 & 64 & 64 & 128 & 64 & 128 & 64 & 128 & 64 \\ 
    Optim & Adam & Adam & Adam & Adam & Adam & Adam & Adam  & Adam & Adam & Adam & Adam & Adam & Adam & Adam  \\
    Sch & Linear & Linear & Linear & Linear & Linear & Linear & Linear & Linear & Linear & Linear & Linear & Linear & Linear & Linear  \\ 
    WD & 0.0 & 0.0 & 0.0 & 0.0 & 0.0 & 0.0 & 0.0 & 0.0 & 0.0 & 0.0 & 0.0 & 0.0 & 0.0 & 0.0 \\
    Clip & 1.0 & 1.0 & 1.0 & 1.0 & 1.0 & 1.0 & 1.0 & 1.0 & 1.0 & 1.0 & 1.0 & 1.0 & 1.0 & 1.0 \\
    LR & 2e-4 & 2e-4 & 1e-4 & 1e-4 & 4e-4 & 1e-4 & 1e-4 & 2e-4 & 1e-4 & 1e-4 & 4e-4 & 1e-4 & 4e-4 & 1e-4 \\ 
    Iter & 5e5 & 5e5 & 1e6 & 5e5 & 1e6 & 1e6 & 5e5 & 5e5 & 1e6 & 5e5 & 2e6 & 5e5 & 1e6 & 1e6 \\ 
    Loss & L2 & L2 & L2 & L2 & L2 & L2 & L2 & L2 & L2 & L2 & L2 & L2 & L2 & L2 \\ 
    dtype & BF16 & BF16 & BF16 & BF16 & BF16 & BF16 & BF16  & BF16 & BF16 & BF16 & BF16 & BF16 & BF16 & BF16 \\
\bottomrule[1.3pt]
\end{tabular}
\end{adjustbox}
\vskip -0.2in
\end{table*}

\end{document}